\documentclass[twoside,11pt]{article}

\usepackage{jmlr2e}

\usepackage{hyperref}       % hyperlinks
\usepackage{url}            % simple URL typesetting
\usepackage{booktabs}       % professional-quality tables
\usepackage{amsfonts}       % blackboard math symbols
\usepackage{nicefrac}       % compact symbols for 1/2, etc.
\usepackage{microtype}      % microtypography

\usepackage{times}
\usepackage{calc}
\usepackage{mathtools}
\usepackage{color}
\usepackage{graphicx} % more modern
\usepackage{DejaVuSans}
\usepackage{parskip}
\usepackage{epsfig}
\usepackage{subfig}
\usepackage{verbatim}
\usepackage{xcolor}
\usepackage{colortbl}
\usepackage{listings}
\usepackage{algorithm}
\usepackage{algorithmic}
\usepackage{booktabs}
\usepackage{enumitem}
\usepackage{bbm}
\usepackage{multirow}
\usepackage{pifont}
\usepackage{wrapfig}
\usepackage{longtable}
\usepackage{eqparbox}
\usepackage{xspace}
\usepackage{kky}
\usepackage{herbertDefns2}
\usepackage{laiDefns}
\usepackage{mockusDefns}
\usepackage{gittinsDefns}
\usepackage{thesisDefns}
\usepackage{koopmansDefns}
\usepackage{bossDefns}

\jmlrheading{21}{2020}{1-25}{4/18; Revised 12/19}{03/20}{18-223}{
Kirthevasan Kandasamy,
Karun Raju Vysyraju,
Willie Neiswanger,
Biswajit Paria,
Christopher R. Collins,
Jeff Schneider,
Barnab\'as P\'oczos,
and
Eric P. Xing
}

\ShortHeadings{\dragonfly: Scalable \& Robust Bayesian Optimisation}
{Kandasamy, Vysyraju, Neiswanger, Paria, Collins, Schneider, P\'oczos, Xing}
\firstpageno{1}

\begin{document}

\title{Tuning Hyperparameters without Grad Students: \\
Scalable and Robust Bayesian Optimisation with \dragonfly}

\newcommand{\cmusymbol}{{\text{\incmtt{1}}}}
\newcommand{\ricesymbol}{{\text{\incmtt{2}}}}
% \newcommand{\ricesymbol}{\flat}

% \author{\name Kirthevasan Kandasamy$\,^\cmusymbol$  \email kandasamy@cs.cmu.edu \\
\author{\name Kirthevasan Kandasamy \email kandasamy@eecs.berkeley.edu \\
%        \addr 
%        University of California, Berkeley,
%        Berkeley, CA 94709, USA
%        \AND
       \name Karun Raju Vysyaraju  \email kvysyara@andrew.cmu.edu \\
%        \AND
       \name Willie Neiswanger \email willie@cs.cmu.edu \\
%        \AND
       \name Biswajit Paria \email bparia@cs.cmu.edu \\
%        \AND
       \name Christopher R. Collins \email crcollins@cmu.edu  \\
%        \AND
       \name Jeff Schneider \email schneide@cs.cmu.edu  \\
%        \AND
       \name Barnab\'as P\'oczos \email bapoczos@cs.cmu.edu \\
%        \AND
       \name Eric P. Xing \email epxing@cs.cmu.edu  \\
       \addr 
       Carnegie Mellon University, 
       Pittsburgh, PA 15213, USA \\
      }

\editor{Bayesian Optimization Special Issue}

\maketitle

\newcommand{\insertAlgoDragonfly}{ 
\begin{algorithm}%[H] 
\vspace{0.02in} 
\begin{algorithmic}[1] 
\REQUIRE $\ninit$, $\ncyc$, $\acqlist$, $\hplist$.
% \STATE $\Dcal_0 \leftarrow \{(\xj, \yj)\}_{j=1}^{\ninit}$
% where $\{\xj\}_{j}$
% Evaluate $\func$ at $\ninit$ points.
\STATE $\Dcal_0 \leftarrow$
Evaluate $\func$ at $\ninit$ points.
\STATE $\maxvalue\leftarrow$ maximum $y$ value in $\Dcal_0$.
\STATE $\wacqtt{} = \initacqwt\one_{|\acqlist|}$.
\STATE $\whptt{} = \inithpwt\one_{2}$.
% \STATE 
% \STATE $\filtrjj{1} \leftarrow \emptyset$, 
%   $\quad\GPtt{1} \leftarrow \GP(\zero, \kernel)$. 
\FOR{$j=0, 1, 2 \dots$}
\STATE Wait for a worker to finish. \label{line:wait} 
\STATE $\filtrj\leftarrow \filtrjj{j-1}\cup\{(q,y)\}$ where $(q,y)$ are the
worker's previous query and observation. \label{line:update_data} 
\IF[update weights if new max-value was found]{$y > \maxvalue$,}
\STATE $\wacqttjj{}{\acqlabel(q)} = \wacqttjj{}{\acqlabel(q)} + 1$.
\STATE $\whpttjj{}{\gphplabel(q)} = \whpttjj{}{\gphplabel(q)} + 1$.
\ENDIF
\IF[updates for GP hyperparameters]{$\mod(t, \ncyc) = 0$,\;}
  \STATE $\Thetasfp\leftarrow$ sample $\ncyc$ GP hyperparameter values.
  \STATE $\thetamml\leftarrow$ maximise GP marginal likelihood to find best
GP hyperparameter values.
  \label{line:hpsample} 
\ENDIF
\STATE $\theta\leftarrow\textrm{multinomail-sample}(
  [\textrm{pop}(\Theta), \thetamml], \whptt{})$.
  \COMMENT{choose GP hyperparameters}
\STATE $\alpha\leftarrow\textrm{multinomial-sample}(\acqlist, \wacqtt{})$.
  \COMMENT{choose acquisition}
\STATE
  $\mutmo\leftarrow$ Compute posterior GP mean given $\filtrj$ using $\theta$.
  \label{line:hppop}
%   i.e. $\GPtmo=\GP(\mutmo,\kerneltmo; \filtrj  \cup
% \{(x, \mutmo(x))\}_{x\in\hallucfiltrj}, \theta)$.
\STATE Compute hallucinated posterior GP
  $\GPtmo\leftarrow \GP(\mutmo,\kerneltmo; \filtrj  \cup
\{(x, \mutmo(x))\}_{x\in\hallucfiltrt}, \theta)$.
\label{line:update_posterior} 
\STATE $q'\leftarrow$ Determine next query for evaluation using
acquisition $\alpha$ and GP $\GPtmo$.
\label{line:selcriterion} 
\STATE Re-deploy worker with an evaluation at $q'$. \label{line:redeploy} 
\ENDFOR
\end{algorithmic} 
\caption{$\;$Bayesian Optimisation in \dragonflys with $M$
asynchronous workers \label{alg:dragonfly}}
\end{algorithm}
}

\newcommand{\insertTableDatasetInfo}{
\begin{table}
\begin{center}
\begin{tabular}{c|c|c|c|c|c|c}
\toprule
& Energy & Protein & Naval & Blog & Indoor & Slice \\
\toprule
$d$ & 27 & 11 & 7 & - & - & - \\
\toprule
$D$ & 25 & 9 & 17 & 281 & 529 & 385 \\
\midrule
$N$ &  20,000 & 20,000 &9,000 & 48,000 & 16,750 & 40,500  \\
\bottomrule
\end{tabular}
\end{center}
\caption{
The domain dimensionality $d=\dim(\Xcal)$ for optimisation, 
data dimensionality $D$ and the size of the dataset $N$ for all datasets
used in our experiments.
For the first three datasets, the fidelity space $\Zcal$
was the amount of data used for training
while for last three $\Zcal$ was the number of iterations the model was trained for.
In all cases, we used $75\%$ of the data for training and $20\%$ for validation.
\label{tb:datasetinf}
}
\end{table}
}

\newcommand{\insertTableMislabelMat}{
\newcommand{\lmmsmalltablehspace}{\hspace{-0.05in}}
\begin{table}
\centering
\begin{minipage}[c]{3.3in}
\vspace{-0.1in}
\small
\begin{tabular}{l|cccccc}
& \lmmsmalltablehspace{\small\convthree} \lmmsmalltablehspace
& \lmmsmalltablehspace{\small\convfive} \lmmsmalltablehspace
& \lmmsmalltablehspace{\small\maxpool} \lmmsmalltablehspace
& \lmmsmalltablehspace{\small\avgpool} \lmmsmalltablehspace
& \lmmsmalltablehspace{\small\fc} \lmmsmalltablehspace\\
\hline
{\small \convthree} & $0$ & $0.2$  & $\infty$ & $\infty$ & $\infty$  \\
{\small \convfive} & $0.2$ & $0$ & $\infty$ & $\infty$ & $\infty$  \\
{\small \maxpool}\hspace{-0.05in} & $\infty$ & $\infty$  & $0$ & $0.25$ & $\infty$  \\
{\small \avgpool}\hspace{-0.05in} & $\infty$ & $\infty$  & $0.25$ & $0$ & $\infty$  \\
{\small \fc} & $\infty$ & $\infty$  & $\infty$ & $\infty$ & $0$  \\
\end{tabular}
\end{minipage}
\hspace{0.2in}
\begin{minipage}[c]{2.4in}
\caption{\small
An example label mismatch cost matrix $\mislabmat$.
$\mislabmat(\textrm{\inlabelfont{x}},\textrm{\inlabelfont{y}})$ is the penalty
for matching unit mass in a layer with label \inlabelfont{x} to one with label
\inlabelfont{y}.  There is zero cost for matching identical layers, small cost
for similar layers, and infinite cost for disparate layers.
\label{tb:mislabmatsmall}
}
\end{minipage}
\end{table}
}

\newcommand{\skipmultirowheight}{*}
\definecolor{lightblue}{rgb}{0.68, 0.85, 0.9}
\definecolor{lightskyblue}{rgb}{0.53, 0.81, 0.98}
\newcommand{\coloursmall}{\cellcolor{white}}
\newcommand{\colourbig}{\cellcolor{white}}

\newcommand{\insertTableEAModifiersSmall}{
\begin{table*}
\centering
\small
\begin{tabular}{|c|p{4.7in}|}
\hline
\textbf{Operation} & \multicolumn{1}{c|}{\textbf{Description}} \\
\hline
{\small \decsingle} \coloursmall &
Pick a layer at random and decrease the number of units by
                          $1/8$. \\
\hline
% \parbox[t]{1cm}{\decenmasse}
% \gaspecialcell{\newline \decenmasse}
% \pbox{2in}{$\quad$\\ $\quad$\\\decenmasse}
\decenmasse
\colourbig & 
Pick several layers at random in topological order and decrease the number of units by $1/8$ for all of them.
% First topologically order the networks, randomly pick $1/8$ of the layers (in order) and
% decrease the number of units by $1/8$.
% For networks with eight layers or fewer pick a $1/4$ of the layers (instead of 1/8)
% and for those with four layers or fewer pick $1/2$.
\\
\hline
\incsingle \coloursmall & Pick a layer at random and increase the number of units by
                          $1/8$. \\
\hline
\incenmasse \colourbig &
Pick several layers  at random in topological order and increase the number of units by $1/8$ for all of them.
% Choose a large sub set of layers, as for \decenmasse{}, and
%   increase the number of units by $1/8$.
\\
\hline
\hline
\branch \colourbig &
Pick a random path $u_1,u_2,\dots,u_{k-1}, u_k$, duplicate layers
$u_2,\dots,u_{k-1}$ and connect them to $u_1$ and $u_k$.
% This modifier duplicates a random path in the network.
% Randomly pick a node $u_1$ and then pick one of its children $u_2$ randomly.
% Keep repeating to generate
% a path $u_1,u_2,\dots,u_{k-1}, u_k$ until you decide to stop randomly.
% Create duplicate layers $\tilde{u}_2,\dots,\tilde{u}_{k-1}$ where
% $\tilde{u}_i = u_i$ for $i=2,\dots,k-1$.
% Add these layers along with new edges $(u_1, \tilde{u}_2)$, $(\tilde{u}_{k-1}, u_k)$,
% and $(\tilde{u}_{j}, \tilde{u}_{j+1})$ for $j=2,\dots,k-2$.
\\
\hline
\removelayer \coloursmall &
Pick a layer at random and remove it.
Connect the layer's parents to its children if necessary.
% If this layer was the only child (parent) of  any of its parents (children) $u$,
% then adds an edge from $u$ (one of its parents) to one of its children ($u$).
 \\
\hline
\colourbig \skiplayer &
Randomly pick layers $u,v$ where $u$ is topologically
before $v$. Add $(u,v)$ to $\edges$.
\\
\hline
\swaplayer \coloursmall & Randomly pick a layer and change its label. \\
\hline
\wedgelayer \coloursmall & Randomly remove an edge $(u,v)$. 
Create a new layer $w$ and add  $(u,w), (w,v)$ to $\edges$.
% $\laylabel(w)$.
% Remove $(u,v)$ from $\edges$ and add $(u,w), (w,v)$.
% If applicable, set the number of units $\layunits(w)$ to be $(\layunits(u) +
% \layunits(v))/2$.
\\
\hline
% % \coloursmall & \multicolumn{-2}{c|c}{Test2} \colourbig \\
% \multicolumn{2}{c|c|}{Test2} \colourbig \\
\end{tabular}
\vspace{-0.05in}
\caption{\small
Descriptions of modifiers to transform one network to another.
The first four change the number of units in the layers but do not change the
architecture, while the last five change the architecture.
% Table~\ref{tb:nnmodifiers} in Appendix~\ref{app:implementation} describes each operation
% in detail.
% The operations shaded in blue make generally ``small'' changes to the network, i.e. the
% distance between the original and modified network is small.
% Those shaded in yellow make generally ``large'' changes.
% \skiplayer{} might make a small or large change depending on the chosen $u,v$.
% \toworkon{Should we justify \incenmasse{}, and \decenmasse{}.}
\label{tb:nnmodifierssmall}
}
\vspace{-0.10in}
\end{table*}
}

\newcommand{\scinot}[2]{$#1\mathrm{e}{#2}$}
\newcommand{\dflmscolwidth}{12.0mm}
\newcommand{\insertModSelTable}{
\begin{table}
{\scriptsize
\setlength{\tabcolsep}{6pt}
\begin{tabular}{l|m{\dflmscolwidth}|m{\dflmscolwidth}|m{\dflmscolwidth}|m{\dflmscolwidth}|
m{\dflmscolwidth}|m{\dflmscolwidth}|m{\dflmscolwidth}|m{12mm}}
\toprule
\vphantom{T}\hspace{-0.05in}Dataset
  & \centering\dragonfly
  & \centering\dragonfly \newline (lowest)
  & \centering\dragonfly \newline \plusmf
  & \centering\gpyopt
  & \centering\smac
  & \centering \rand
  & \centering\rand\newline(lowest)
  &           \hyperband
  \\
\midrule
\vphantom{T}\hspace{-0.05in}RFR (News)
  & \centering{$0.6456$}\newline{$\pm0.08044$}
  & \centering{$.6631$}\newline{$\pm0.0110$}
  & \centering{$\bf 0.5203$}\newline{$\bf \pm0.0805$}
  & \centering{$0.5904$}\newline{$\pm0.0635$}
  & \centering{$0.9802$}\newline{$\pm0.0291$}
  & \centering{$0.9314$}\newline{$\pm0.0371$}
  & \centering{$0.8501$}\newline{$\pm0.0563$}
  &           {$0.6812$}\newline{$\pm0.0412$}
  \\
\midrule
\vphantom{T}\hspace{-0.05in}GBR (Naval)
  & \centering{\scinot{5.82}{-5}}\newline\centering{$\pm$\scinot{1.52}{-5}}
%   & \centering{$2.34$}\newline{$\pm0.01$}
  & \centering{\scinot{\bf 1.14}{-5}}\newline{\centering $\bf \pm$\scinot{\bf 7.3}{-7}}
  & \centering{\scinot{3.00}{-5}}\newline{\centering $\pm$\scinot{1.00}{-5}}
  & \centering{\scinot{3.26}{-4}}\newline{$\pm$\scinot{1.46}{-5}}
  & \centering{\scinot{1.97}{-5}}\newline{$\pm$\scinot{5.27}{-6}}
  & \centering{\scinot{1.04}{-3}}\newline{$\pm$\scinot{1.10}{-5}}
  & \centering{\scinot{1.31}{-5}}\newline{$\pm$\scinot{2.34}{-6}}
  &           {\scinot{\bf 1.13}{-5}}\newline{$\bf \pm$\scinot{\bf 6.4}{-7}}
  \\
\midrule
\vphantom{T}\hspace{-0.05in}SALSA (Energy) \hspace{-0.08in}
  & \centering{$0.0097$}\newline{$\pm0.0015$}
  & \centering{$0.9950$}\newline{$\pm0.0033$}
  & \centering{$0.0095$}\newline{$\pm0.0031$}
  & \centering{$0.0105$}\newline{$\pm0.0022$}
  & \centering{$\bf 0.0084$}\newline{$\bf \pm0.0003$}
  & \centering{$0.0634$}\newline{$\pm0.0019$}
  & \centering{$0.9974$}\newline{$\pm0.0032$}
  &           {$0.3217$}\newline{$\pm0.1032$}
  \\
\bottomrule
\end{tabular}
\vspace{-0.08in}
}
\caption{\small
Final least squared errors in the regression problems of Section~\ref{sec:dflexpmodsel}.
In addition to the methods in Figure~\ref{fig:dflmodsel}, we also compare to
\dragonflys and random search at the lowest fidelity, as well as \hyperband.
\label{tb:modselresults}
\vspace{-0.1in}
}
\end{table}
}

\newcommand{\insertTableRankSummary}{
\begin{table}
{\scriptsize
\setlength{\tabcolsep}{6pt}
\begin{tabular}{l|m{\dflmscolwidth}|m{\dflmscolwidth}|m{\dflmscolwidth}|m{\dflmscolwidth}|
m{\dflmscolwidth}|m{\dflmscolwidth}|m{\dflmscolwidth}|m{10mm}}
\toprule
\vphantom{T}\hspace{-0.05in}Method
  & \centering\dragonfly
  & \centering\gpyopt
  & \centering\spearmint
  & \centering\smac
  & \centering\hyperopt
  & \centering\pdoo
  & \centering\rand
  &           \evoalg
  \\
\midrule
\vphantom{T}\hspace{-0.05in}Euclidean
%   & \centering{$8-5-3-0$}
  & \centering{8-5-2-0}
  & \centering{3-3-4-2}
  & \centering{3-3-3-0}
  & \centering{1-0-3-2}
  & \centering{0-3-0-2}
  & \centering{0-0-3-4}
  & \centering{0-0-1-5}
  &           {N/A}
  \\
\midrule
\vphantom{T}\hspace{-0.05in}Non-Euclidean
  & \centering{5-2-0-1}
  & \centering{2-0-1-1}
  & \centering{1-3-0-2}
  & \centering{0-2-4-1}
  & \centering{0-0-2-1}
  & \centering{N/A}
  & \centering{0-0-1-0}
  &           {0-1-0-1}
  \\
\bottomrule
\end{tabular}
\vspace{-0.08in}
}
\caption{\small
A summary of the ranks of each method on the Euclidean and non-Euclidean synthetic
problems at the end of $200$ evaluations.
We show the number of times each method was ranked within the first four in the order
$n1-n2-n3-n4$ where $<nk>$ is the number of times the method was ranked $k$\ssth in
the problems.
There are a total of 15 Euclidean problems (12 from Figure~\ref{fig:dfleuc}, 3 from
Figure~\ref{fig:dflnoisy}) and 8 non-Euclidean problems (6 from Figure~\ref{fig:dflcpresults},
2 from Figure~\ref{fig:dflnoisy}).
We exclude the multi-fidelity methods from this table.
\label{tb:ranks}
\vspace{-0.1in}
}
\end{table}
}

\newcommand{\imtextspace}{-0.05in}
\newcommand{\imfourwidth}{1.3in}
\newcommand{\imfourhspace}{0.1in}
\newcommand{\imthreewidth}{1.7in}
\newcommand{\imthreehspace}{0.1in}

\newcommand{\imarrwtwo}{2.88in}
\newcommand{\imhsptwo}{-0.00in}
\newcommand{\imarrwthree}{1.99in}
\newcommand{\imhspthree}{-0.03in}
\newcommand{\imleftspace}{-0.10in}
\newcommand{\imsinglecol}{2.495in}
\newcommand{\imcaptionspace}{-0.05in}

\newcommand{\insertHorFigFidelSpace}{ 
\begin{figure}
\centering
  \begin{minipage}[c]{2.3in}
    \includegraphics[width=2.2in]{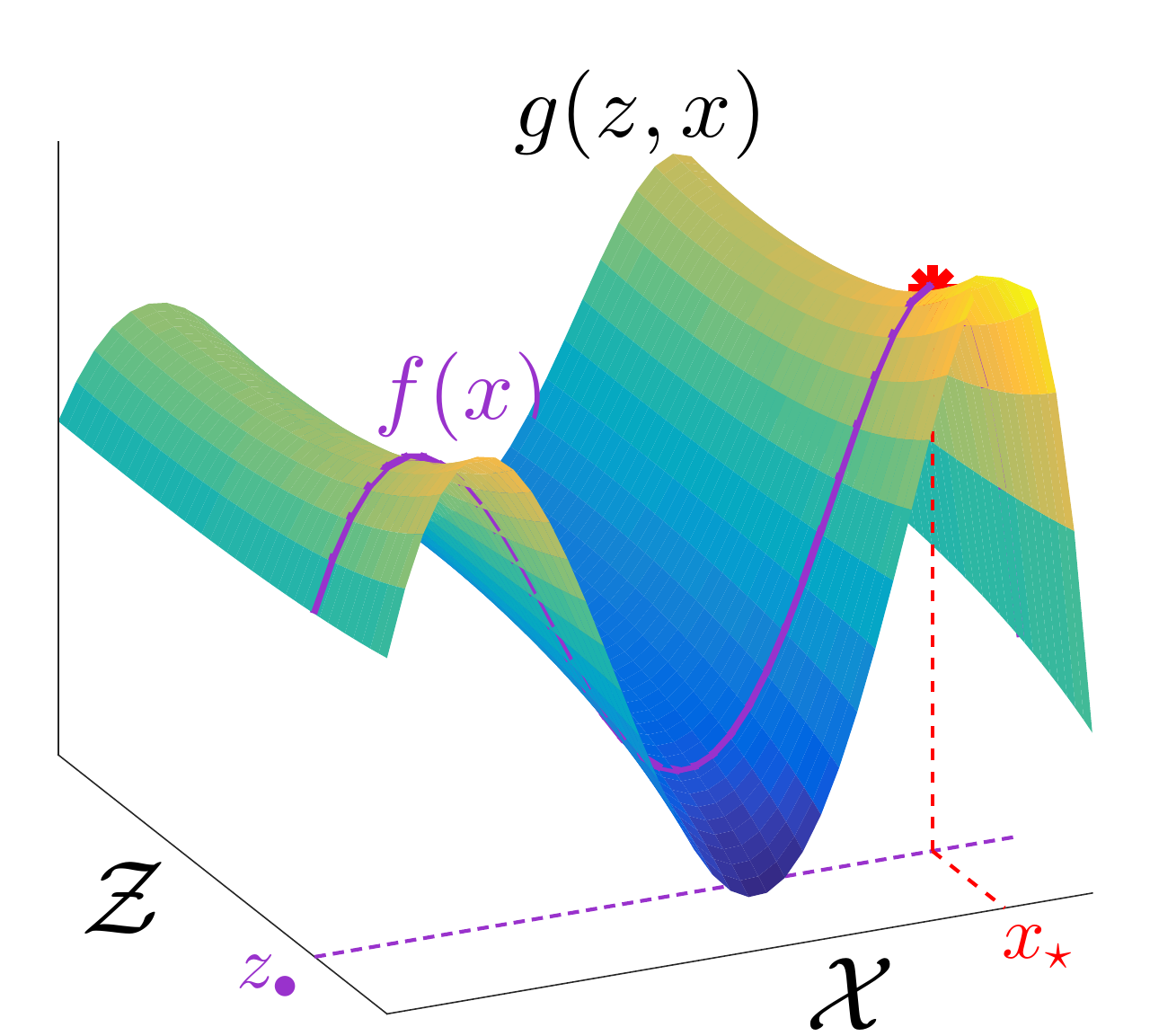}
  \end{minipage} \hspace{0.3in}
  \begin{minipage}[l]{3.0in}
  \vspace{-0.0in}
    \caption{
$g:\Zcal\times\Xcal\rightarrow\RR$ is a function defined on the product space of the
fidelity space $\Zcal$ and domain $\Xcal$.
The purple line is $f(x) = g(\zhf, x)$.
We wish to find the maximiser $\xopt \in \argmax_{x\in\Xcal} f(x)$.
% The multi-fidelity framework is attractive when $g$ is smooth across $\Zcal$ as
% illustrated in the figure.
  \label{fig:fidelSpace}
    }
  \end{minipage}
%   \vspace{\imtextspace}
\end{figure}
}

\newcommand{\dflablationfigwidth}{2.15in}
\newcommand{\dflablationfigspace}{-0.25in}
\newcommand{\insertFigdflhps}{
\begin{figure}
\centering
\hspace{\dflablationfigspace}
  \includegraphics[width=\dflablationfigwidth]{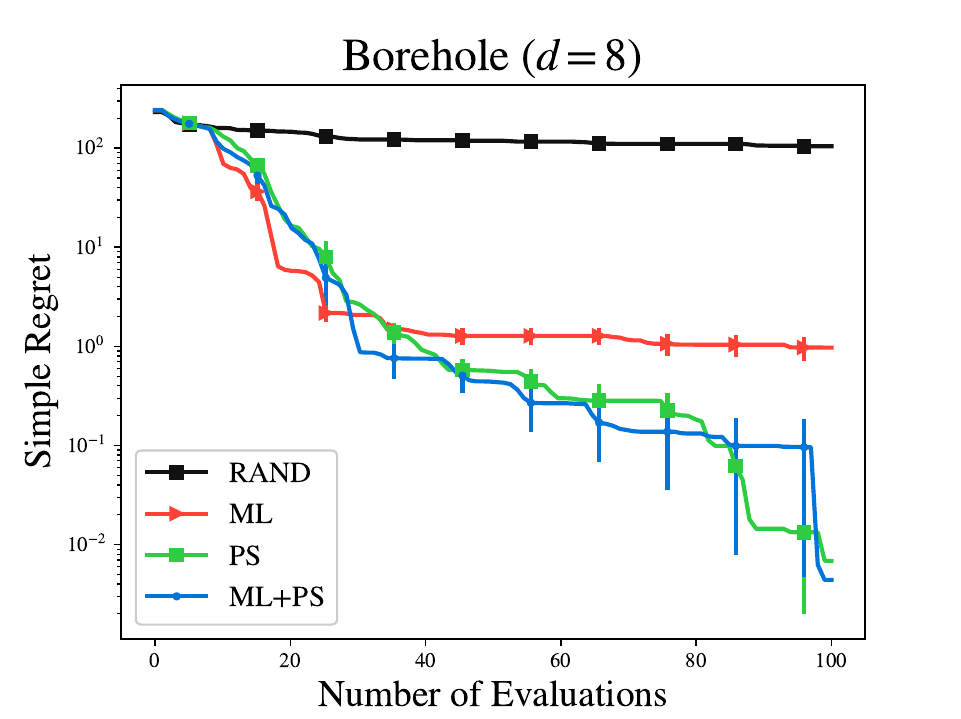}
\hspace{\dflablationfigspace}
  \includegraphics[width=\dflablationfigwidth]{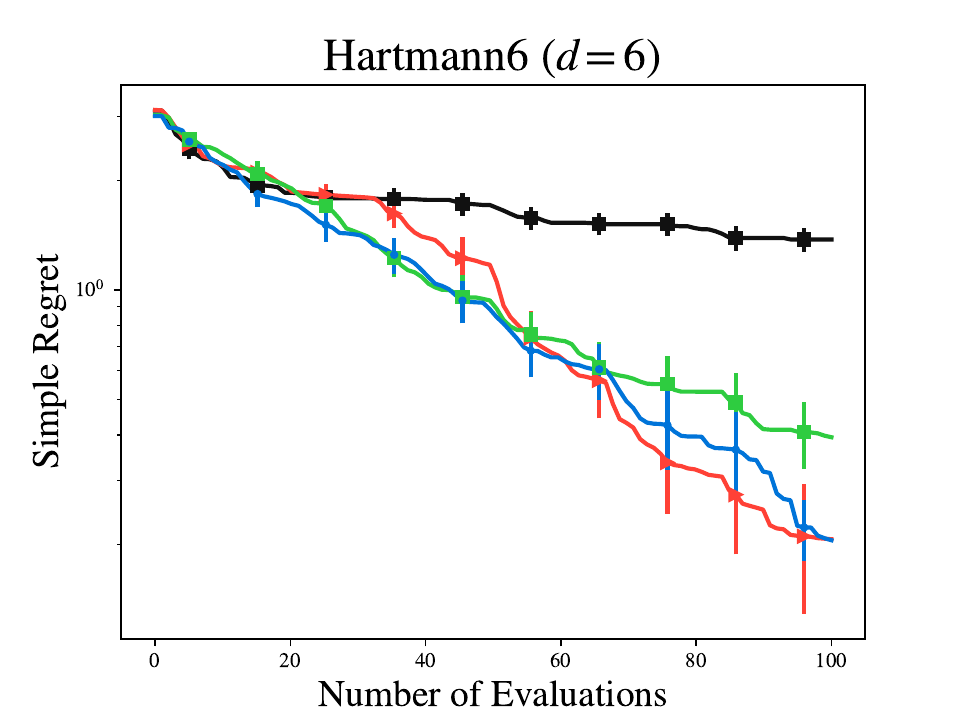}
\hspace{\dflablationfigspace}
  \includegraphics[width=\dflablationfigwidth]{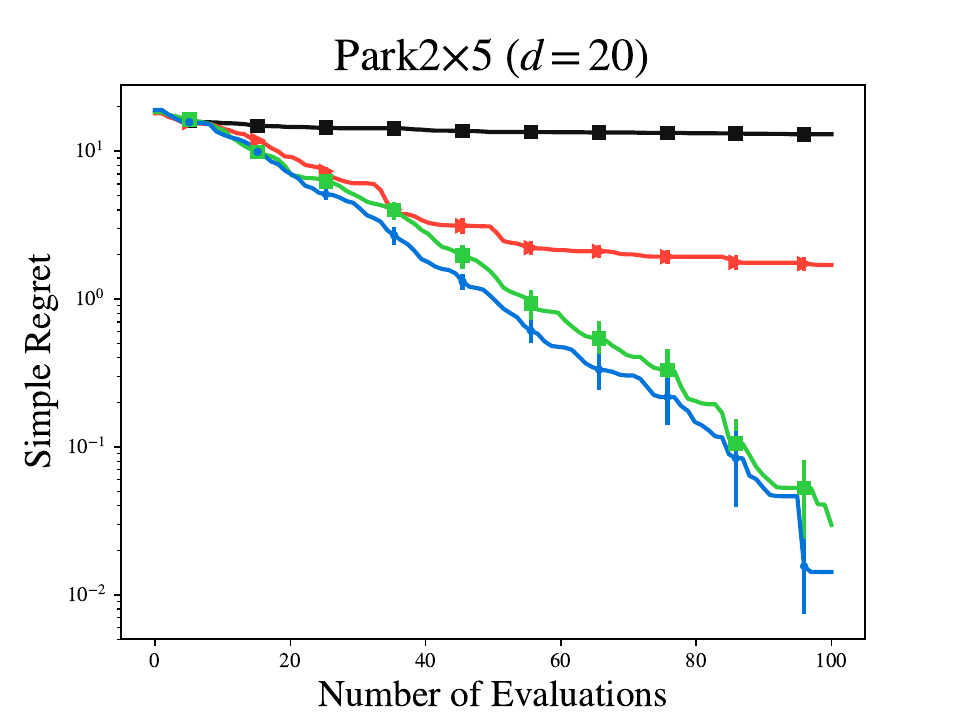}
\hspace{\dflablationfigspace}
% \vspace{\imcaptionspace}
\caption{\small
Comparison using only maximum likelihood (\mml), only posterior sampling (\sfp) and
the combined sampling approach (\mml+\sfp) as described in Section~\ref{sec:dflgphps}.
We have also shown random sampling (\rand) for comparison.
% The combined approach is able to perform as well as or better than the best
% choice for the given problem.
We plot the simple regret~\eqref{eqn:regretDefn}, so lower is better.
All curves were produced by averaging over $10$ independent runs.
Error bars indicate one standard error.
\label{fig:dflgphps}
\vspace{\imtextspace}
}
\end{figure}
}

\newcommand{\insertFigdflacqs}{
\begin{figure}
\centering
\hspace{\dflablationfigspace}
  \includegraphics[width=\dflablationfigwidth]{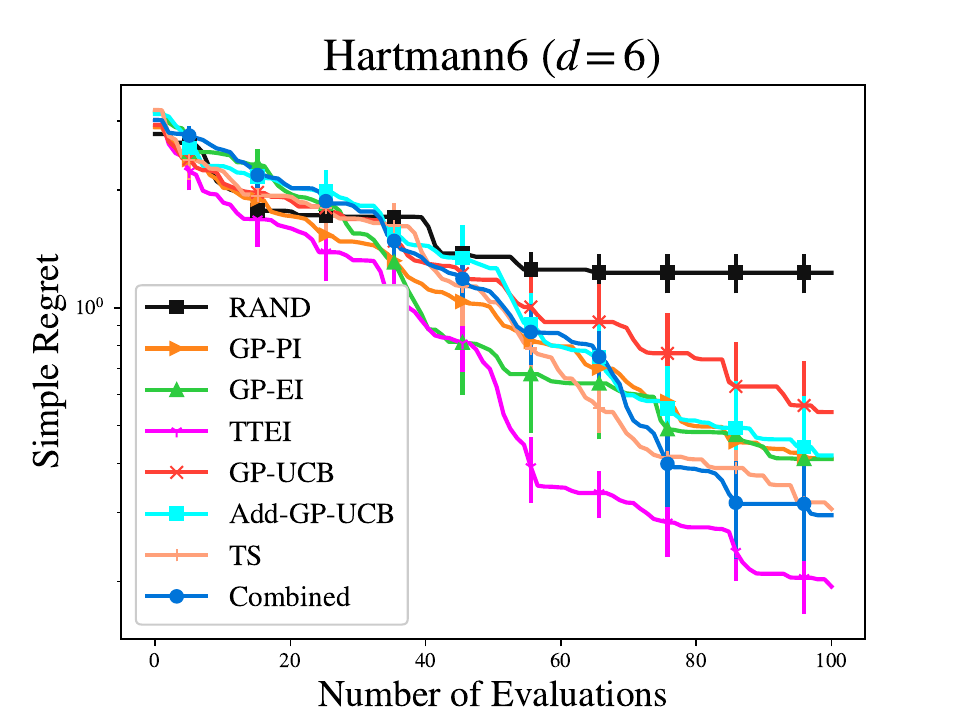}
\hspace{\dflablationfigspace}
  \includegraphics[width=\dflablationfigwidth]{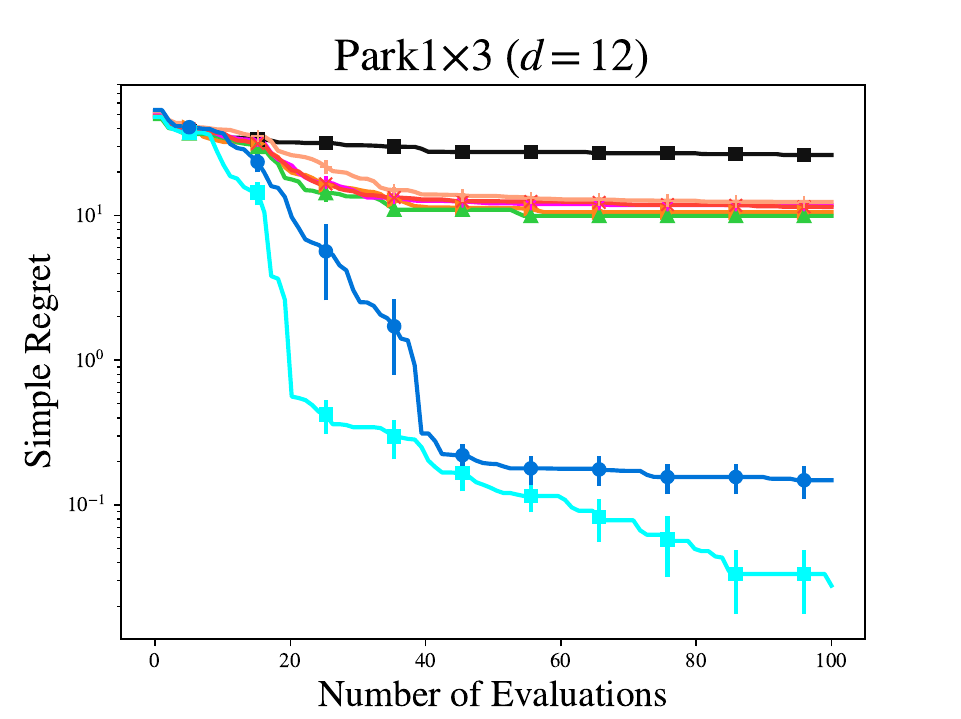}
\hspace{\dflablationfigspace}
  \includegraphics[width=\dflablationfigwidth]{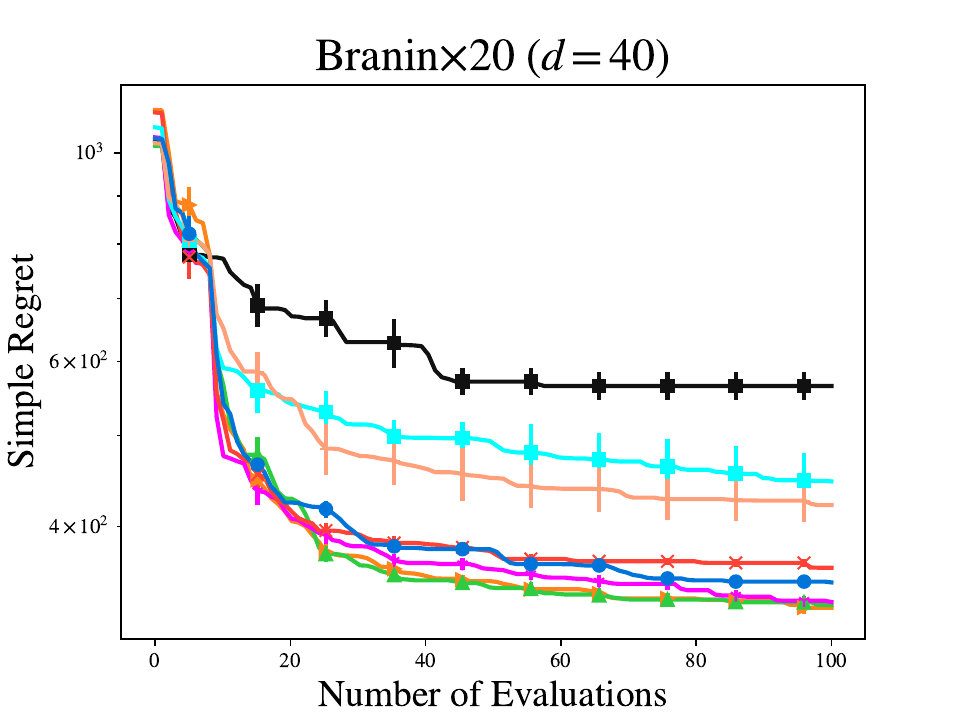}
\hspace{\dflablationfigspace}
\vspace{\imcaptionspace}
\caption{\small
\label{fig:dflacqs}
Comparison of using  individual acquisitions such as \gpucb, \gpei, \ttei,
\ts, \probi, and \addgpucbs versus
the combined sampling method  as described in Section~\ref{sec:dflacqs}.
We have also shown random sampling (\rand) for comparison.
% The combined approach is typically able to perform almost as well as the
% single best acquisition on each individual problem.
We plot the simple regret~\eqref{eqn:regretDefn}, so lower is better.
Error bars indicate one standard error.
All curves were produced by averaging over $10$ independent runs.
\vspace{\imtextspace}
}
\end{figure}
}

\newcommand{\insertKDEHorFigure}{
\begin{figure}
\centering
%   \vspace{-0.1in}
  \hspace{-0.1in}
  \begin{minipage}[c]{2.0in}
    \includegraphics[width=2.2in]{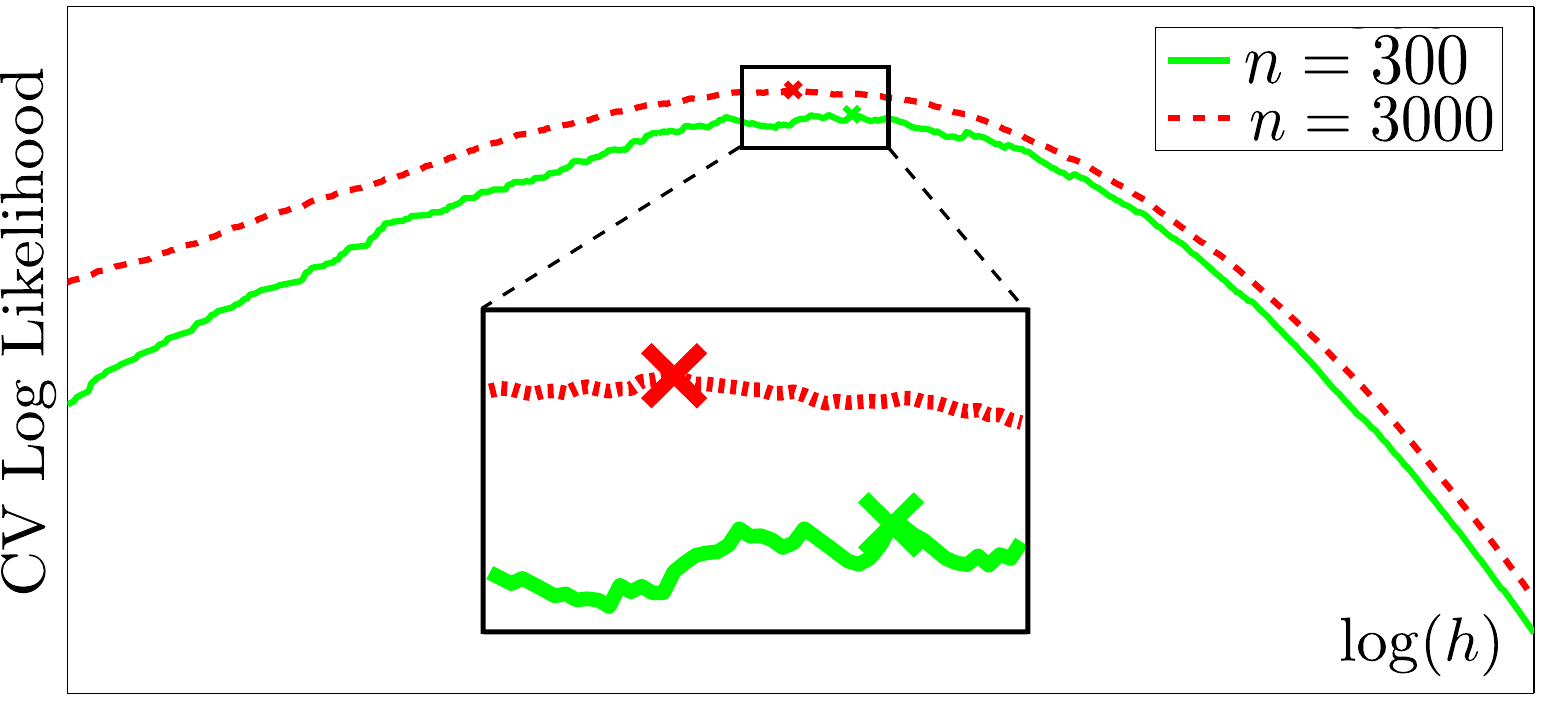}
  \end{minipage} \hspace{0.3in}
  \begin{minipage}[l]{3.6in}
%   \vspace{-0.1in}
    \caption{ \small
Average cross validation log likelihood on datasets of size $300$
and $3000$ on a synthetic kernel density estimation task.
The crosses are the maxima.
The maximisers are different
since optimal hyper-parameters depend on the training set size.
That said, the curve
for $n=300$ approximates the $n=3000$ curve quite well.
  \label{fig:kdeEg}
    }
  \end{minipage}
  \vspace{\imtextspace}
\end{figure}
}

\newcommand{\insertHorFigExpDecayKernel}{ 
\begin{figure}
\centering
  \begin{minipage}[c]{1.6in}
  \vspace{-0.1in}
    \includegraphics[width=1.5in]{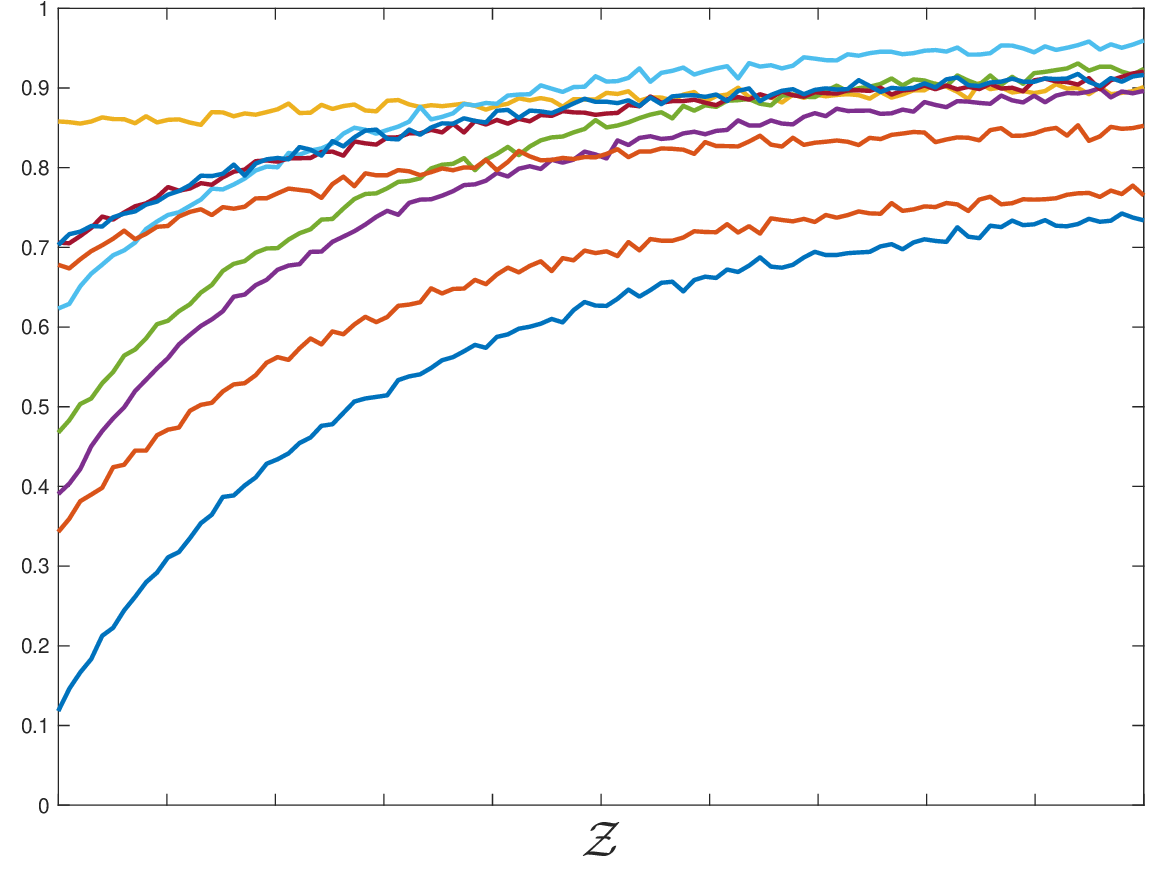}
  \end{minipage} \hspace{0.06in}
  \begin{minipage}[l]{4.2in}
    \caption{
An illustration of GP sample paths drawn from the
    exponential decay kernel~\citep{swersky2014freeze} conditioned on being positive.
They are suitable for representing the validation accuracy along a fidelity dimension
in model selection where, for e.g. validation accuracy tends to increase as
we use more data and/or train for more iterations.
  \label{fig:expdecay}
    }
  \end{minipage}
  \vspace{\imtextspace}
  \vspace{\imtextspace}
  \vspace{\imtextspace}
\end{figure}
}

\newcommand{\insertFigEuchHP}{
\begin{figure}
\centering
    \includegraphics[width=\imthreewidth]{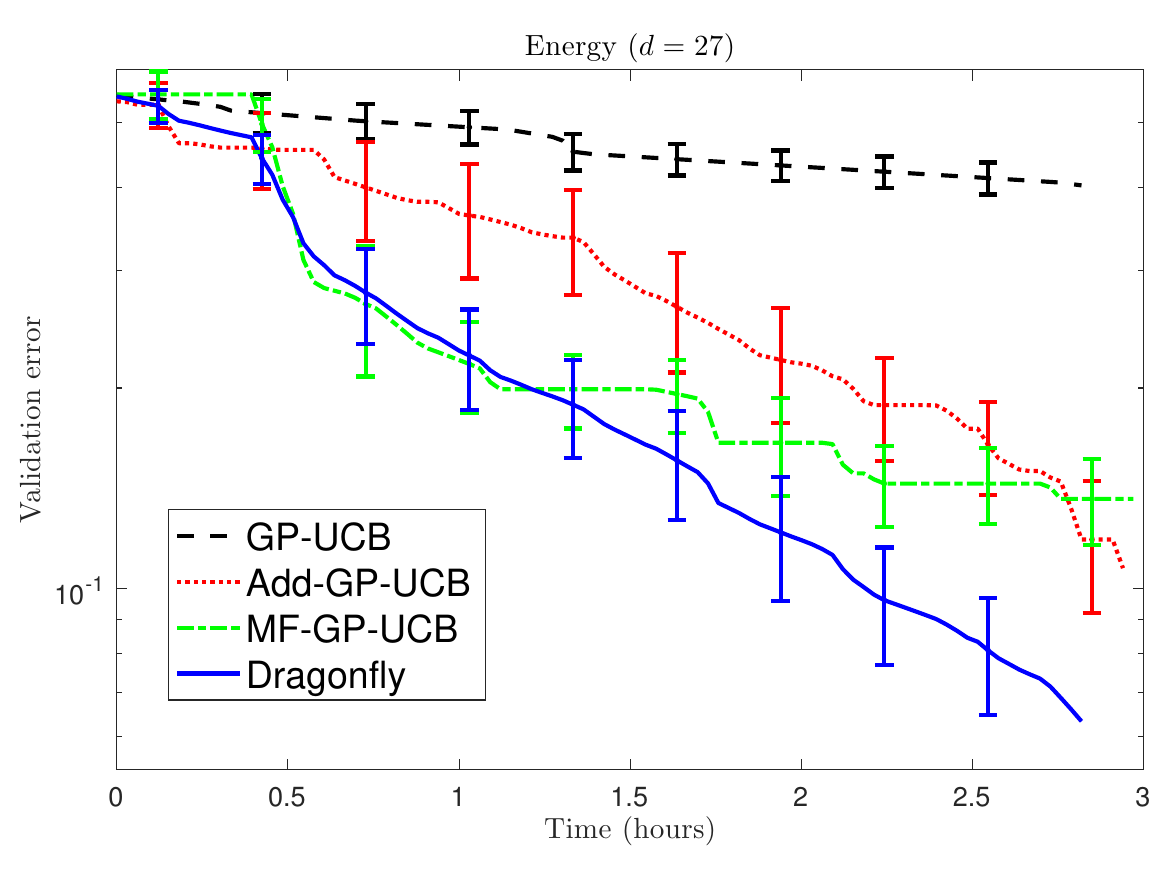}
      \hspace{\imthreehspace}
    \includegraphics[width=\imthreewidth]{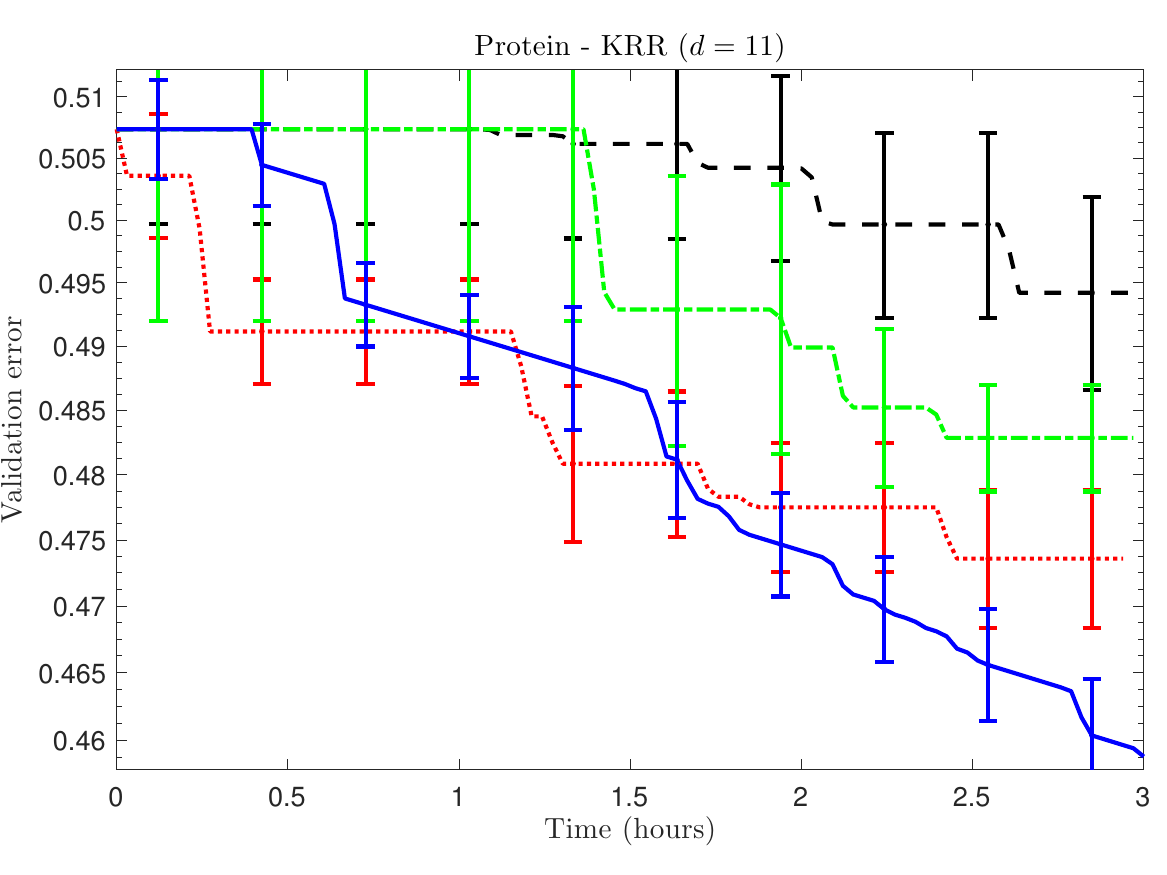}
      \hspace{\imthreehspace}
    \includegraphics[width=\imthreewidth]{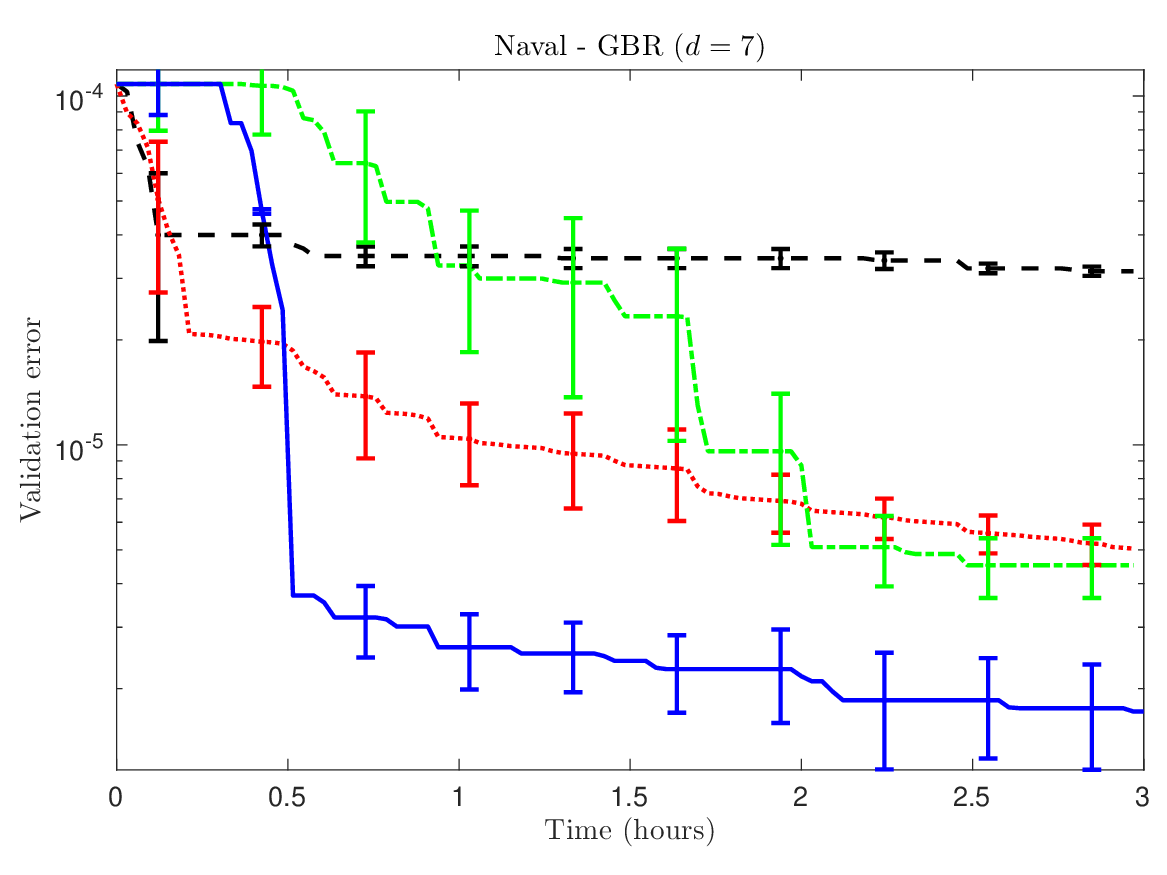}
    \caption{
\emph{Results on tuning scalar hyper-parameters:}
In all figures, the $x$-axis is time.
The $y$ axis is the mean squared validation error (lower is better).
The title of each figure states the dataset and the algorithm.
$d$ denotes the number of hyperparameters (dimensionality of the domain).
In all cases, we used a one dimensional fidelity space where we chose between
20\% to 100\% ($\zhf = 100\% \text{of the data}$).
All figures were averaged over at least 6 independent runs of each method.
Error bars indicate one standard error.
  \label{fig:euchp}
    }
\end{figure}
}

\newcommand{\insertFigNNAs}{
\begin{figure}
\centering
    \includegraphics[width=\imthreewidth]{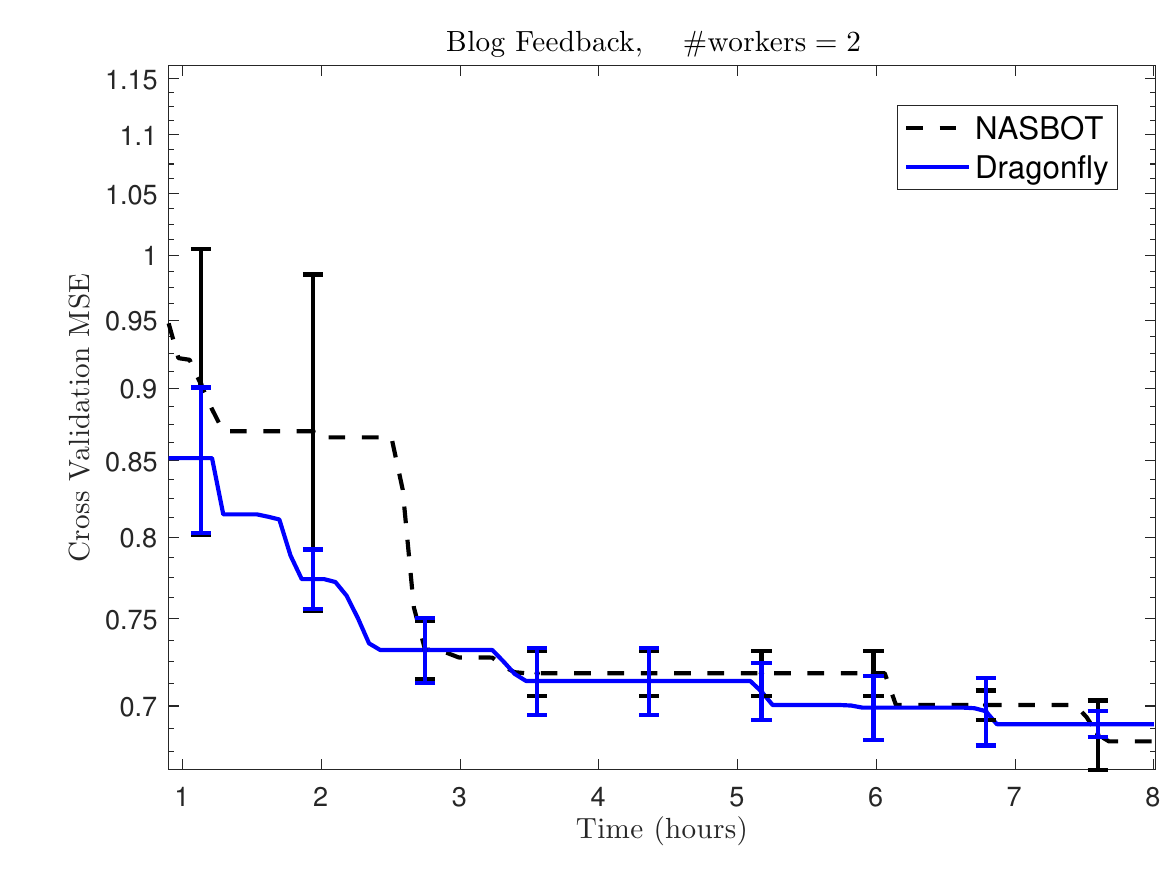}
      \hspace{\imthreehspace}
    \includegraphics[width=\imthreewidth]{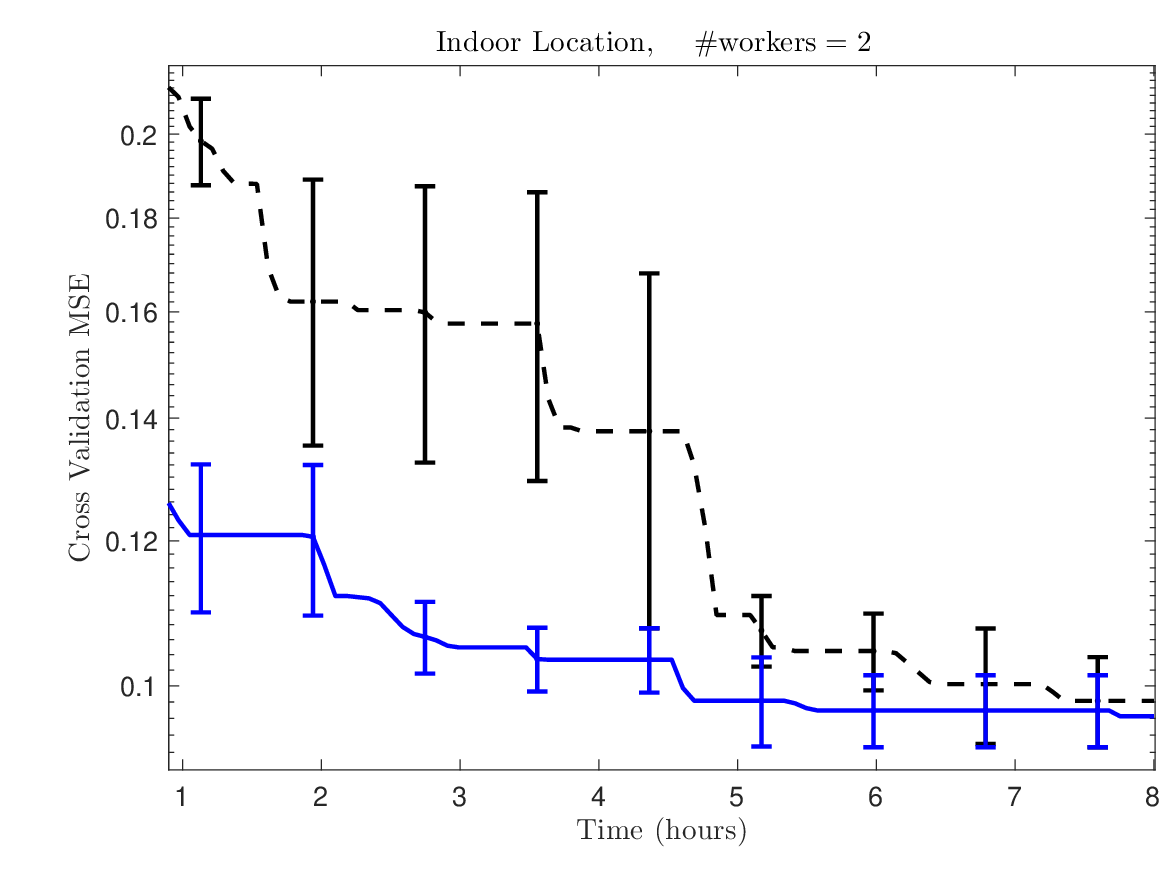}
      \hspace{\imthreehspace}
    \includegraphics[width=\imthreewidth]{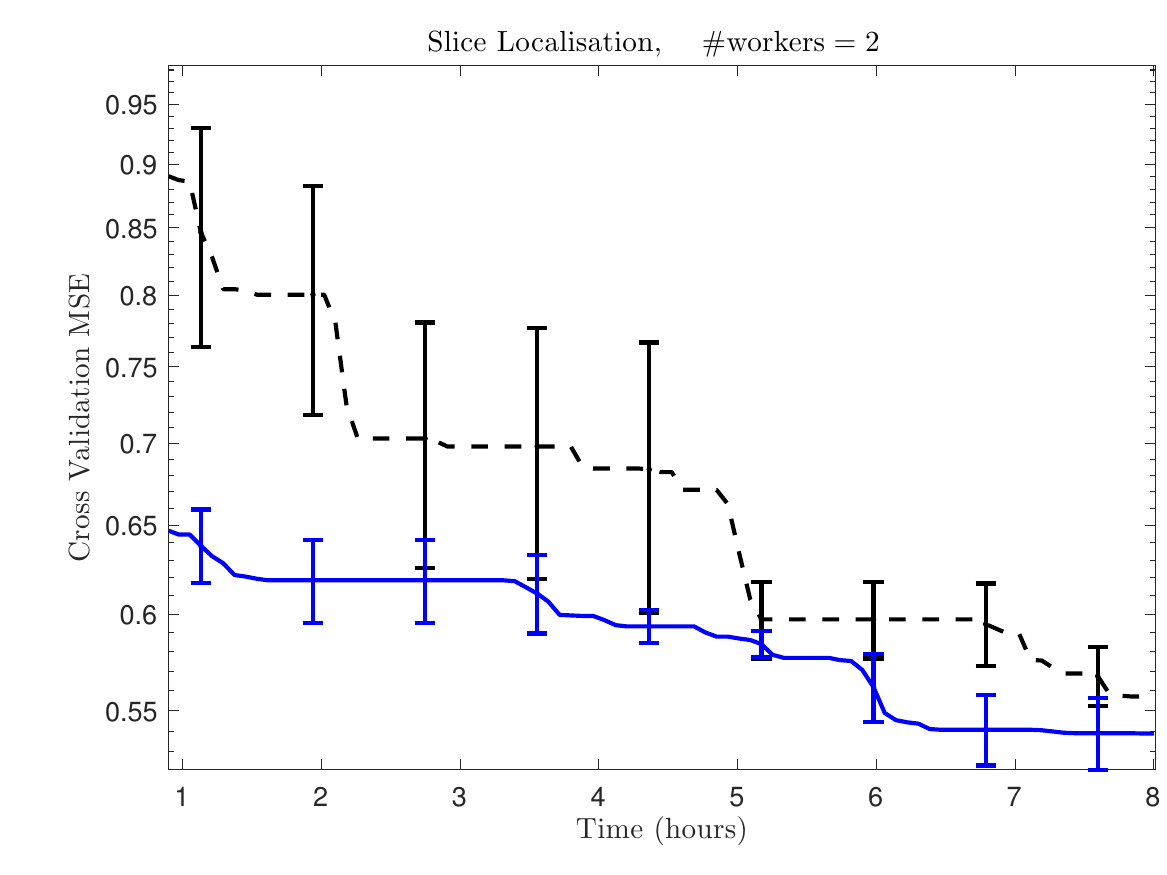}
    \caption{
\emph{Results on the neural architecture search experiments:}
In all figures, the $x$-axis is time.
The $y$ axis is the mean squared validation error (lower is better).
The title of each figure states the dataset.
In all cases, we used a one dimensional fidelity space where we chose the number of
batch iterations from 4000 to 20,000 ($\zhf=20,000$).
All figures were averaged over at least 5 independent runs.
Error bars indicate one standard error.
  \label{fig:nna}
    }
\end{figure}
}

\newcommand{\insertFigNNEgsMain}{
\newcommand{\mainnnfigwidth}{1.1in}
\newcommand{\mainnnfigheight}{2.232in}
\newcommand{\mainnnfighsp}{\hspace{0.025in}}
\begin{figure}
\begin{center}
\centering
\begin{minipage}[c]{2.80in}
\subfloat[]{
\includegraphics[height=\mainnnfigheight]{figs/cnn_egs_1/1}
\label{fig:mainnneg1}}\mainnnfighsp
\subfloat[]{
\includegraphics[height=\mainnnfigheight]{figs/cnn_egs_1/0}
\label{fig:mainnneg2}}\mainnnfighsp
\subfloat[]{
\includegraphics[height=\mainnnfigheight]{figs/cnn_egs_1/2} 
\label{fig:mainnneg3}}
\vspace{-0.05in}
\end{minipage}
\hspace{0.1in}
\begin{minipage}[c]{3.0in}
\caption{
An illustration of some CNN architectures.
In each layer, $i$: indexes the layer, followed by the label (e.g \convthree),
and then the number of units
(e.g. number of filters). \\[0.05in]
% The input and output layers are pink while the decision (\softmax) layers
% are green.\\
% \emph{From Sec.~\ref{sec:nndistmain}:}
The layer mass, used in \otmann, is denoted in parentheses.
The following are the \otmanns distances $d$ for the networks.
All self distances are $0$, i.e.
$d(\Gcal,\Gcal) = 0$.
Moreover,
$d(\textrm{a},\textrm{b}) = 175.1$,\;
$d(\textrm{a},\textrm{c}) = 1479.3$,\;
$d(\textrm{b},\textrm{c}) = 1621.4$,\;
% $d(\textrm{\subref*{fig:mainnneg1}},\textrm{\subref*{fig:mainnneg2}}) = 175.1$,\;
% $d(\textrm{\subref*{fig:mainnneg1}},\textrm{\subref*{fig:mainnneg3}}) = 1479.3$,\;
% $d(\textrm{\subref*{fig:mainnneg2}},\textrm{\subref*{fig:mainnneg3}}) = 1621.4$.
% Normalised:
% $\dbar(\textrm{(a)},\textrm{(b)}) = 0.0286$,\;
% $\dbar(\textrm{(a)},\textrm{(c)}) = 0.2395$,\;
% $\dbar(\textrm{(b)},\textrm{(c)}) = 0.2625$.%
% $\dbar(\textrm{\subref*{fig:mainnneg1}},\textrm{\subref*{fig:mainnneg2}}) = 0.0286$,\;
% $\dbar(\textrm{\subref*{fig:mainnneg1}},\textrm{\subref*{fig:mainnneg3}}) = 0.2395$,\;
% $\dbar(\textrm{\subref*{fig:mainnneg2}},\textrm{\subref*{fig:mainnneg3}}) = 0.2625$.%
% \hspace{-0.1in}
% % 
% % while the decision layers are shown in ...
\label{fig:mainnnegs}
% \vspace{-0.20in}
}
\end{minipage}
\end{center}
\end{figure}
}

\newcommand{\insertFigGPUCB}{
\begin{figure}
\centering
\hspace{\imleftspace}
  \includegraphics[width=\imarrwthree]{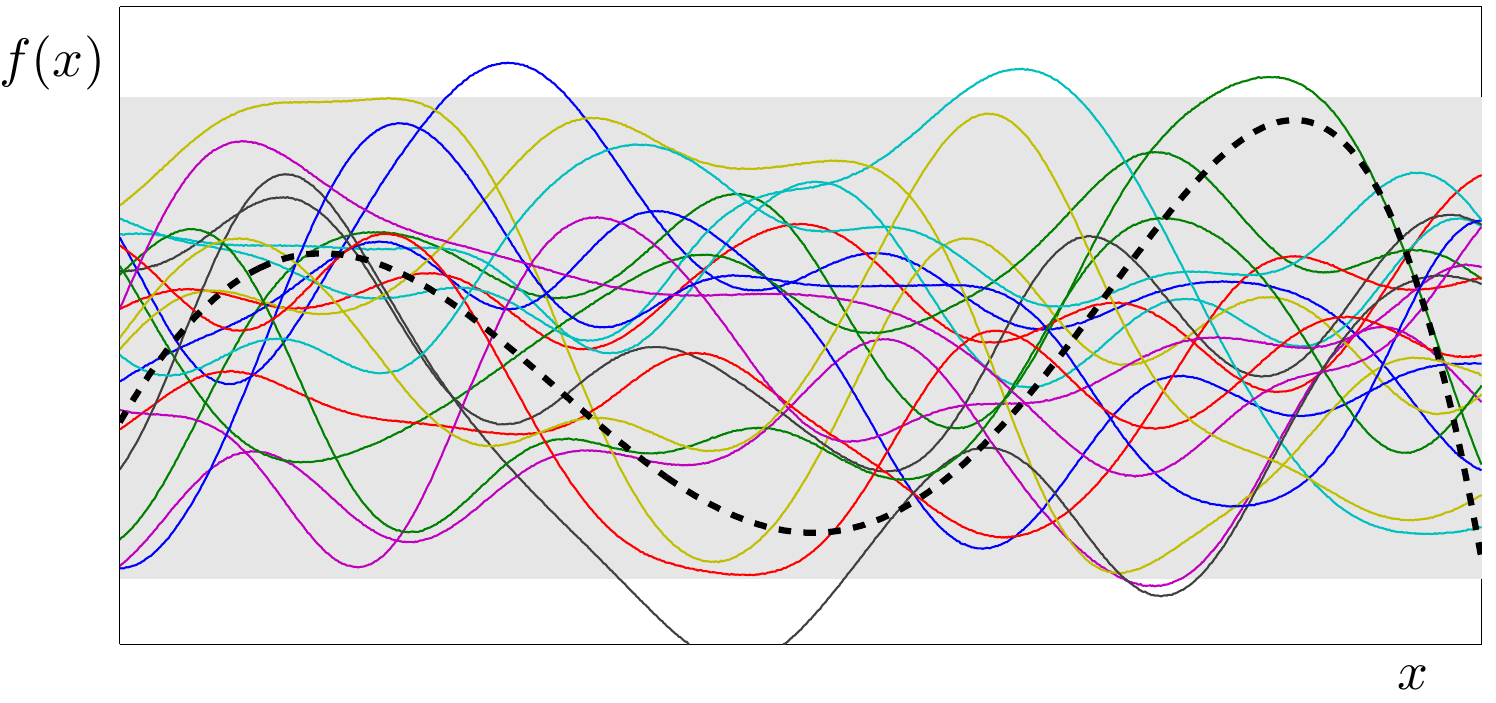} \hspace{\imhspthree}
  \includegraphics[width=\imarrwthree]{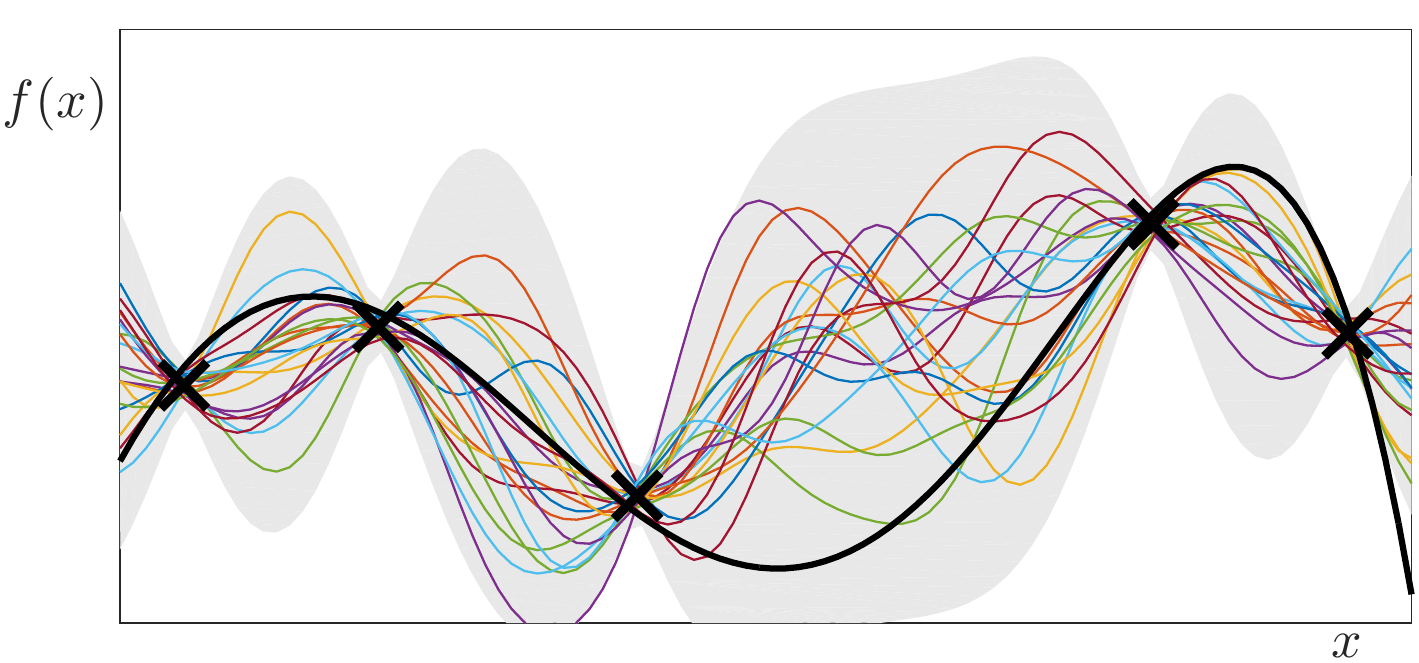} \hspace{\imhspthree}
  \includegraphics[width=\imarrwthree]{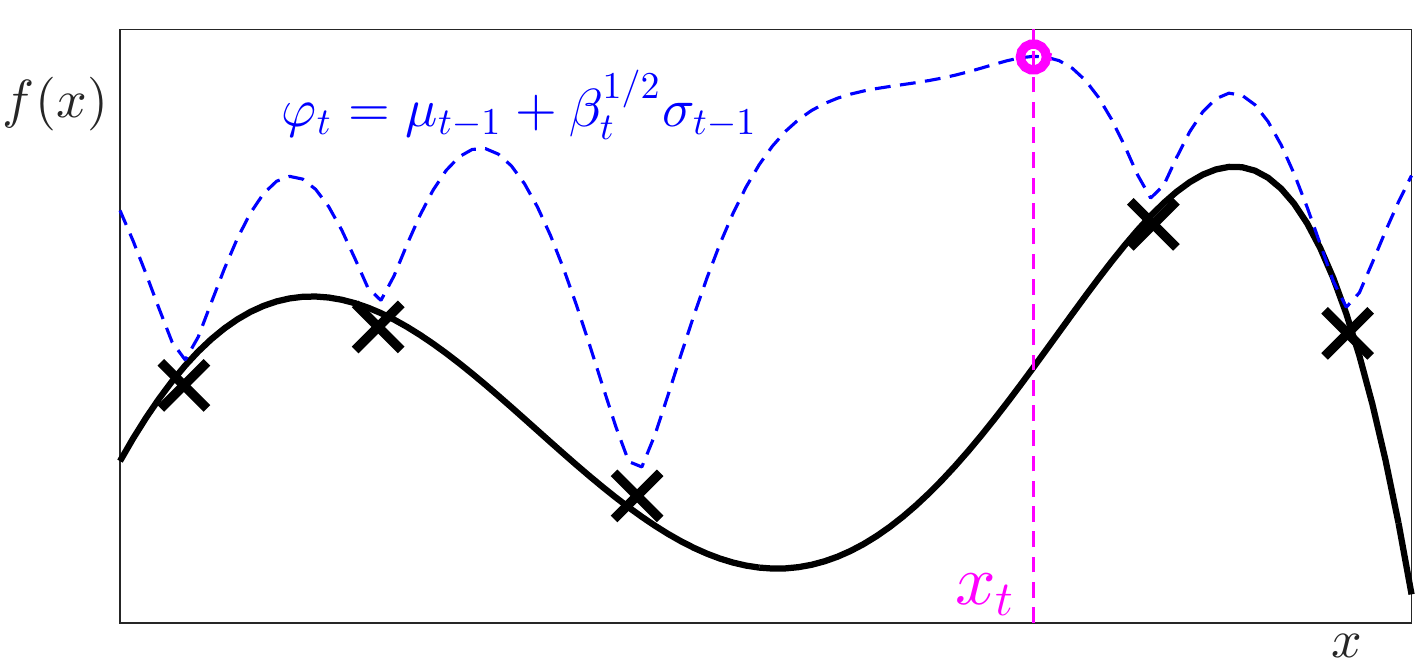}
\vspace{-0.23in}
\caption{\small
An illustration of GPs and \bayo.  The first figure shows the function of
interest $\func$ (black line) before any observations and illustrates a GP that
represents the prior uncertainty.  The shaded region represents a $99\%$
confidence region for $\func$ and the coloured lines are samples from the GP.
The second figure shows some noisy observations (black $\times$'s) of $\func$
and the posterior GP conditioned on the observations.  The confidence region
has shrunk around the observations.  In the third figure we illustrate \gpucbs
when we have to pick the next point $\xt$ given observations as shown in the
second figure.  The \gpucbs acquisition $\utilt$ upper bounds $\func$. At time
$t$, we choose the maximiser of $\utilt$ for evaluation, i.e $\xt \in \argmax_x
\utilt(x)$.
\label{fig:gpucb}
  \vspace{\imtextspace}
  \vspace{\imtextspace}
}
\end{figure}
}

\newcommand{\insertFigdfleucresults}{
\newcommand{\imarrwdfleuc}{2.15in}
\newcommand{\imhspdfleuc}{-0.25in}
\begin{figure}
\centering
\hspace{\imhspdfleuc}
  \includegraphics[width=\imarrwdfleuc]{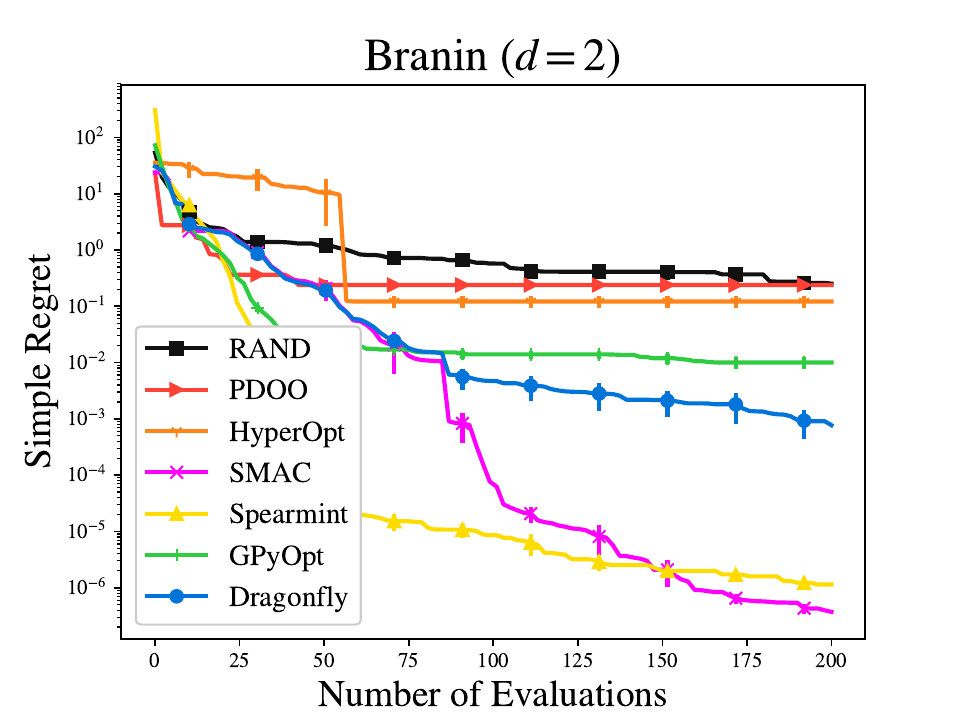}
\hspace{\imhspdfleuc}
  \includegraphics[width=\imarrwdfleuc]{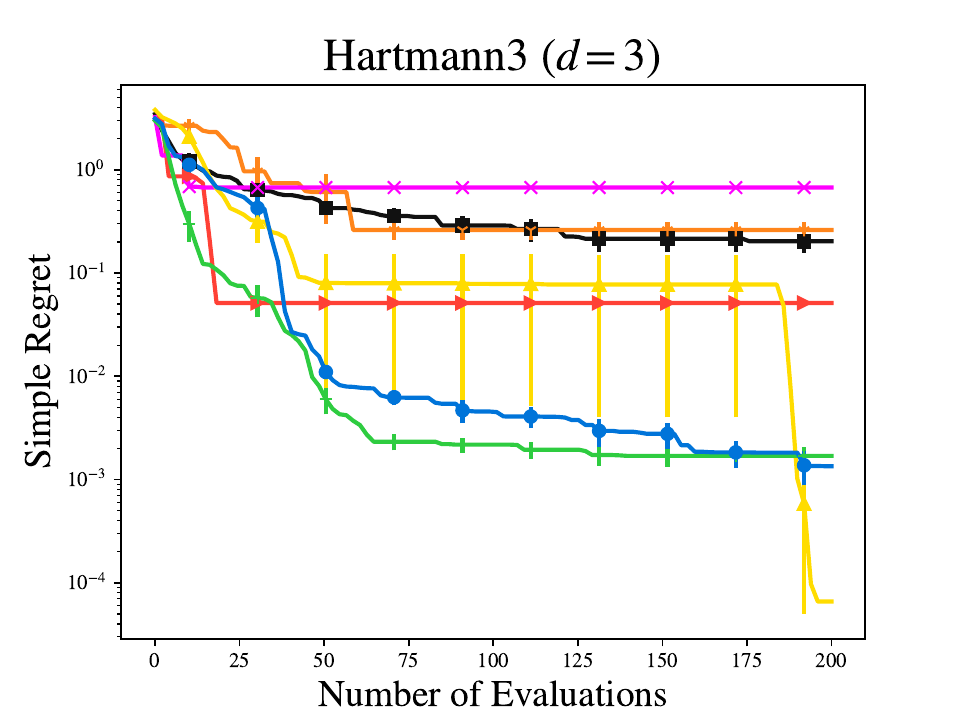}
\hspace{\imhspdfleuc}
  \includegraphics[width=\imarrwdfleuc]{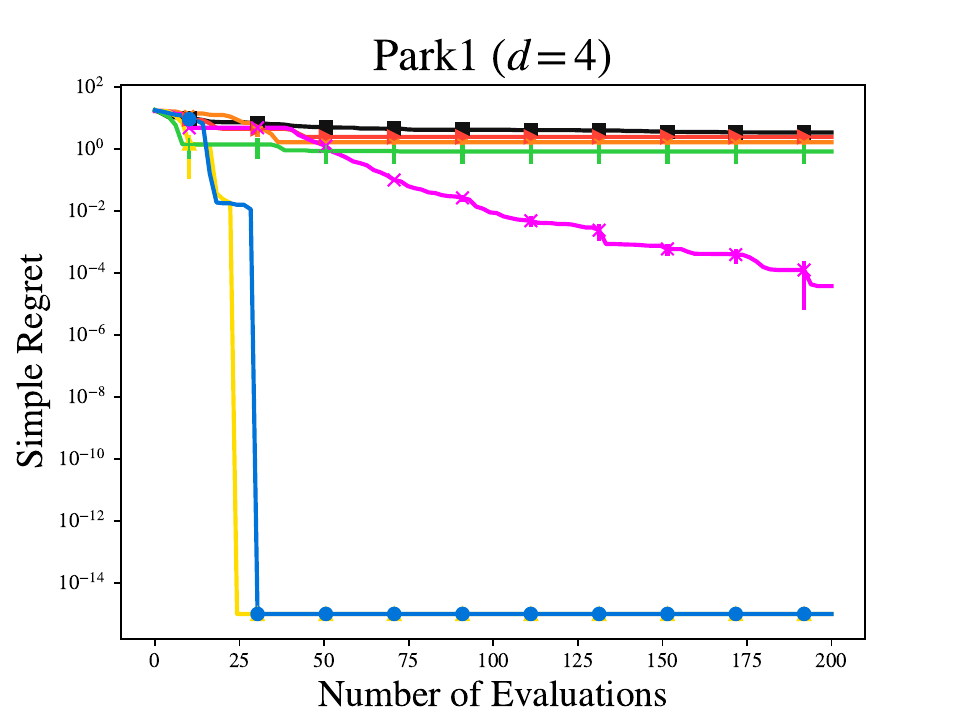}
\hspace{\imhspdfleuc}
\\
\hspace{\imhspdfleuc}
  \includegraphics[width=\imarrwdfleuc]{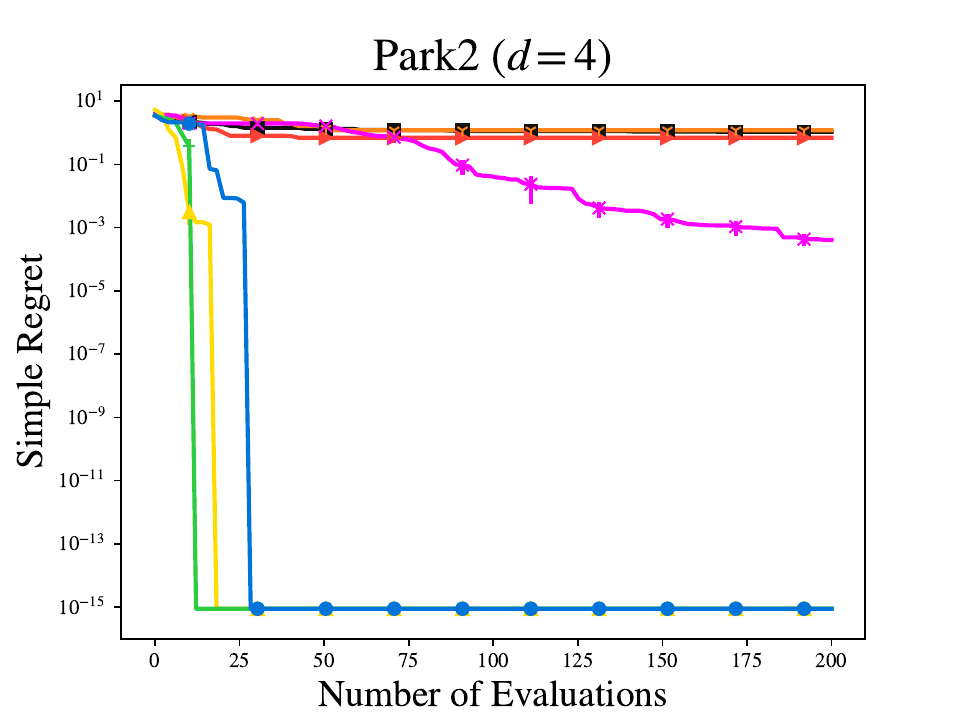}
\hspace{\imhspdfleuc}
  \includegraphics[width=\imarrwdfleuc]{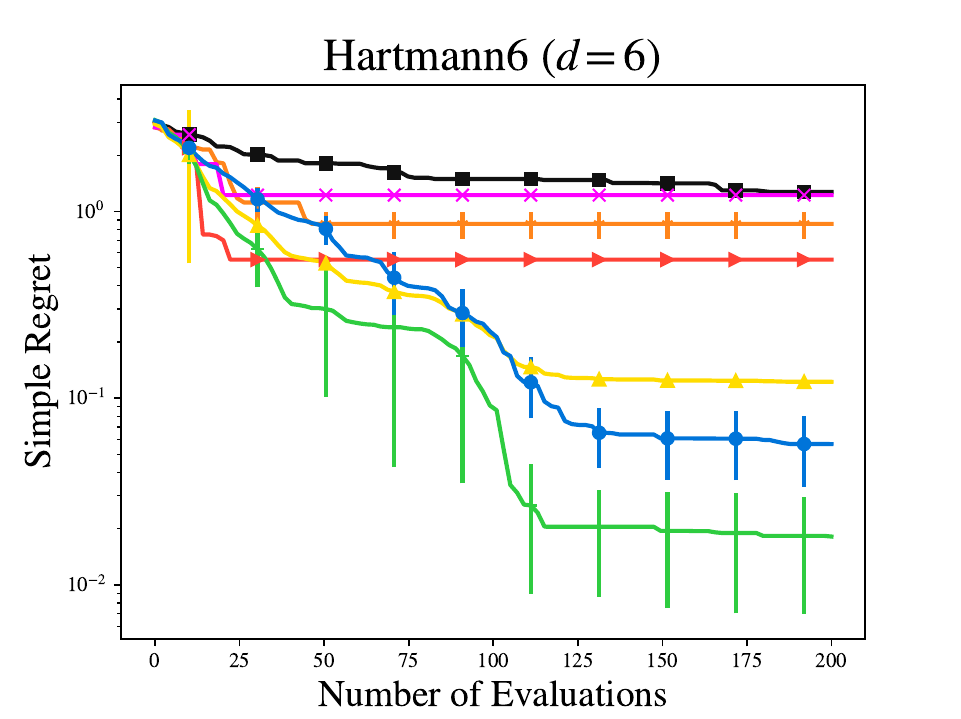}
\hspace{\imhspdfleuc}
  \includegraphics[width=\imarrwdfleuc]{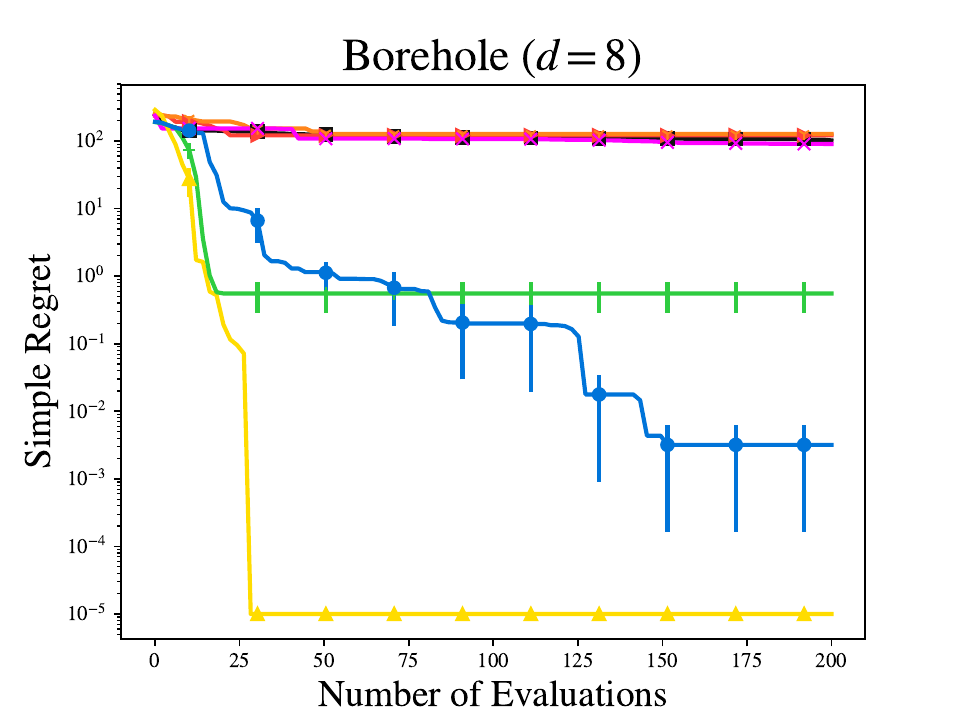}
\hspace{\imhspdfleuc}
\\
\hspace{\imhspdfleuc}
  \includegraphics[width=\imarrwdfleuc]{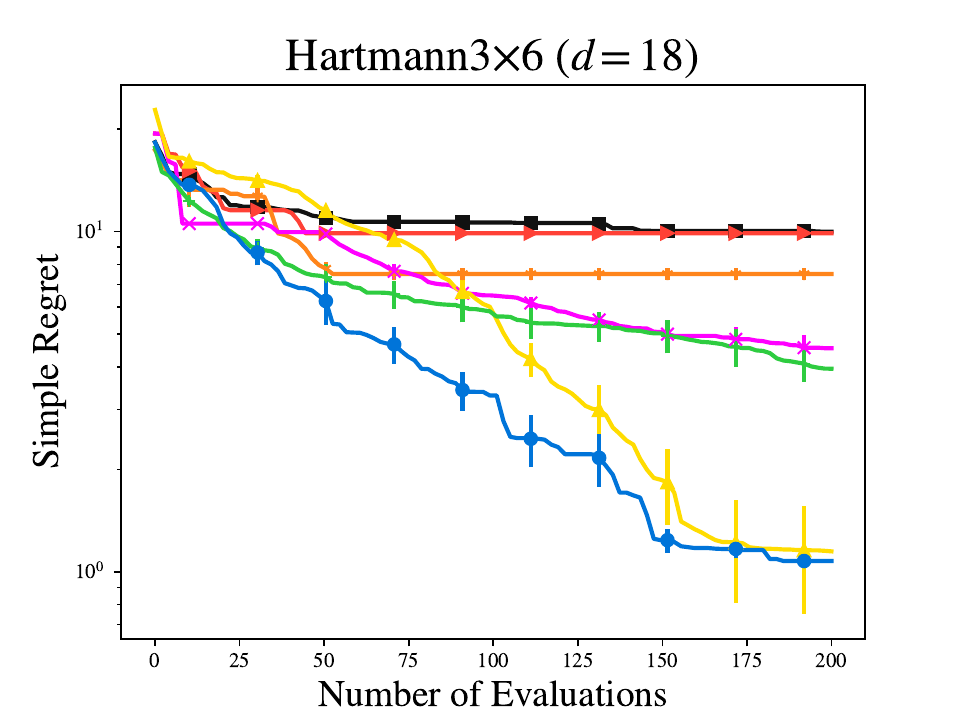}
\hspace{\imhspdfleuc}
  \includegraphics[width=\imarrwdfleuc]{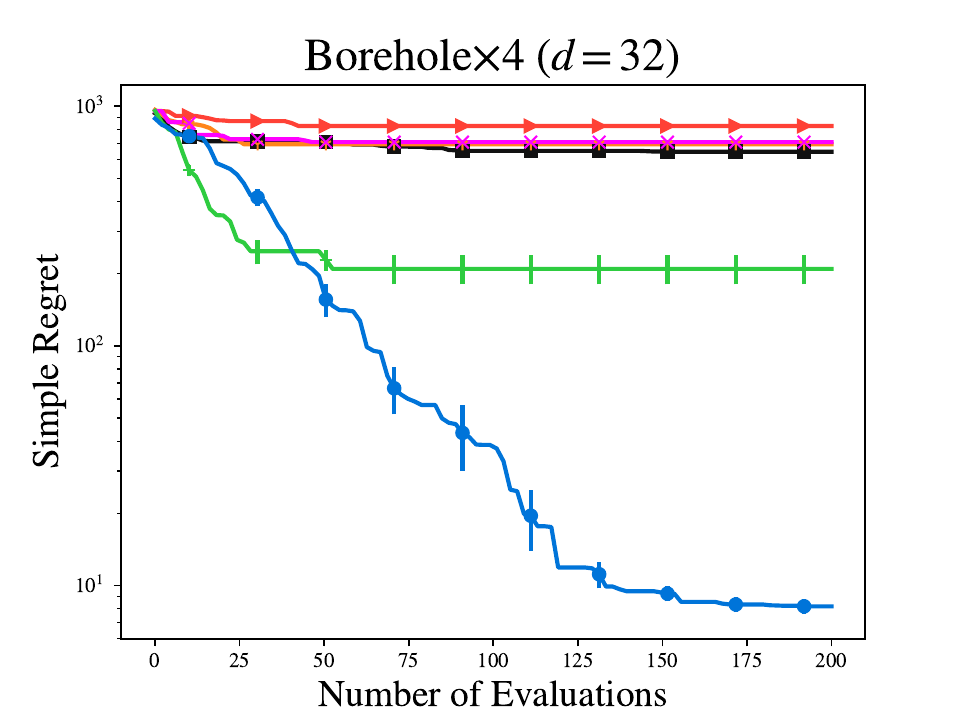}
\hspace{\imhspdfleuc}
  \includegraphics[width=\imarrwdfleuc]{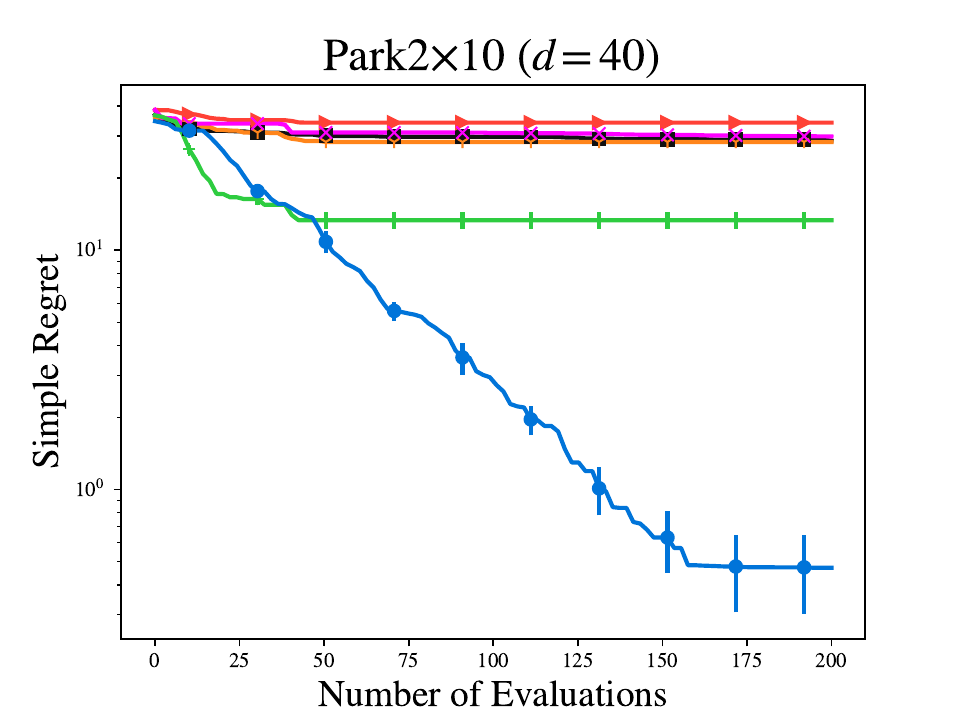}
\hspace{\imhspdfleuc}
\\
\hspace{\imhspdfleuc}
  \includegraphics[width=\imarrwdfleuc]{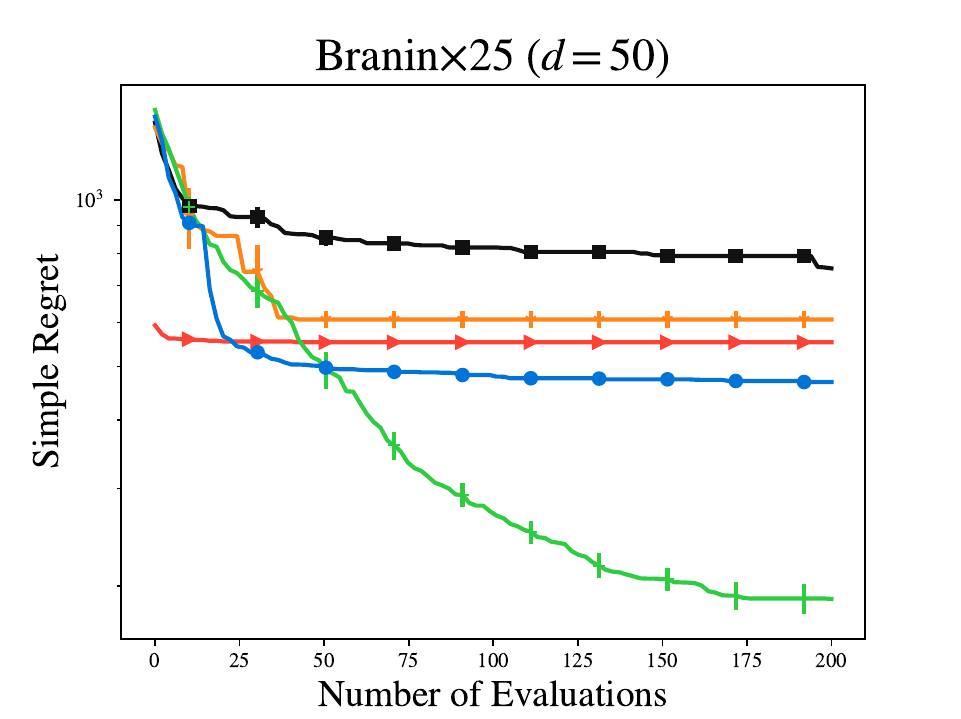}
\hspace{\imhspdfleuc}
  \includegraphics[width=\imarrwdfleuc]{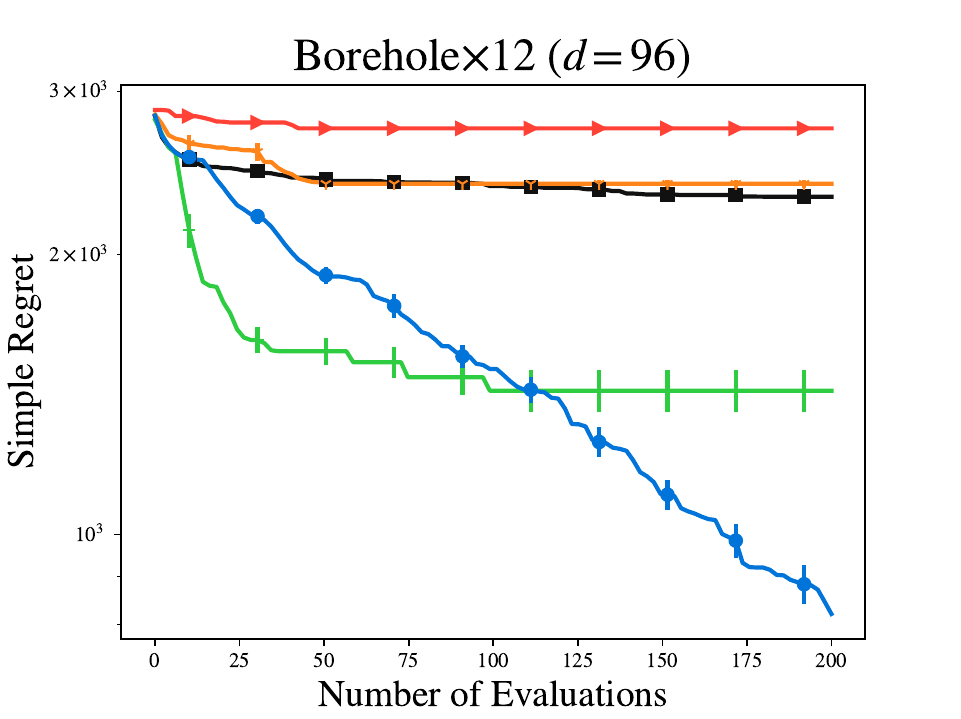}
\hspace{\imhspdfleuc}
  \includegraphics[width=\imarrwdfleuc]{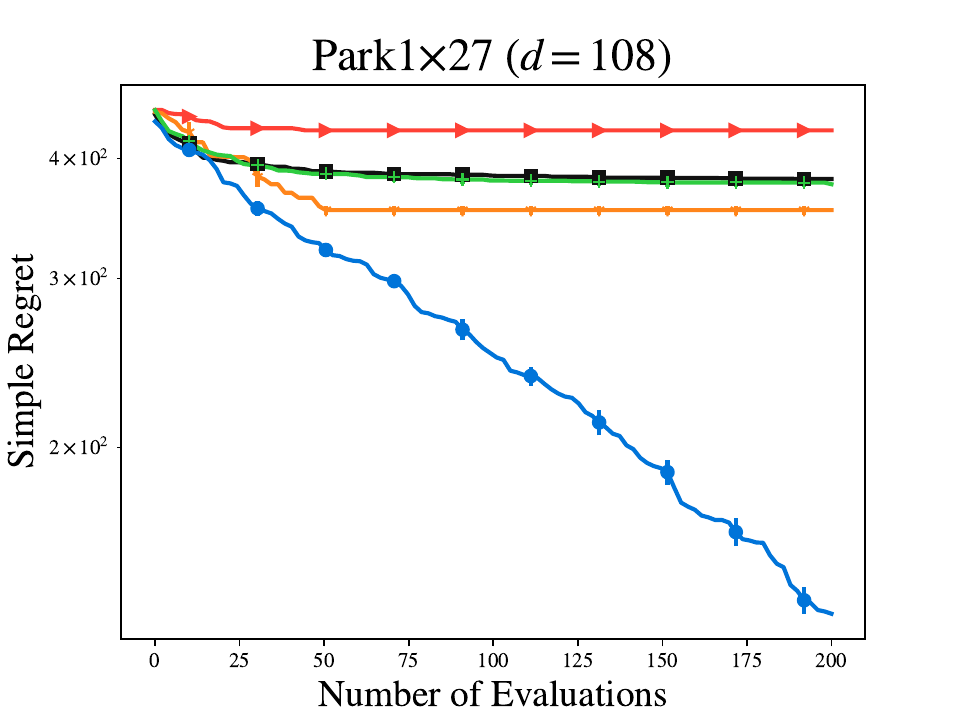}
\hspace{\imhspdfleuc}
\\
% \vspace{\imcaptionspace}
\caption{\small
\label{fig:dfleuc}
Comparison of \dragonflys with other algorithms and BO packages on functions with
\emph{noiseless} evaluations defined on Euclidean domains.
We plot the simple regret~\eqref{eqn:regretDefn} so lower is better.
The title states the name of the function, and its dimensionality.
All curves were produced by averaging over $20$ independent runs.
Error bars indicate one standard error.
The legend for all curves is available in the first figure.
\smac's initialisation procedure did not work in dimensions larger than $40$ so it is not
shown in the respective figures.
\spearmint{} is not shown on all figures since it was too slow to run on
high dimensional problems.
}
\end{figure}
}

\newcommand{\insertFigdflcpresults}{
\newcommand{\imarrwdflcp}{2.15in}
\newcommand{\imhspdflcp}{-0.25in}
\begin{figure}
\centering
\hspace{\imhspdflcp}
  \includegraphics[width=\imarrwdflcp]{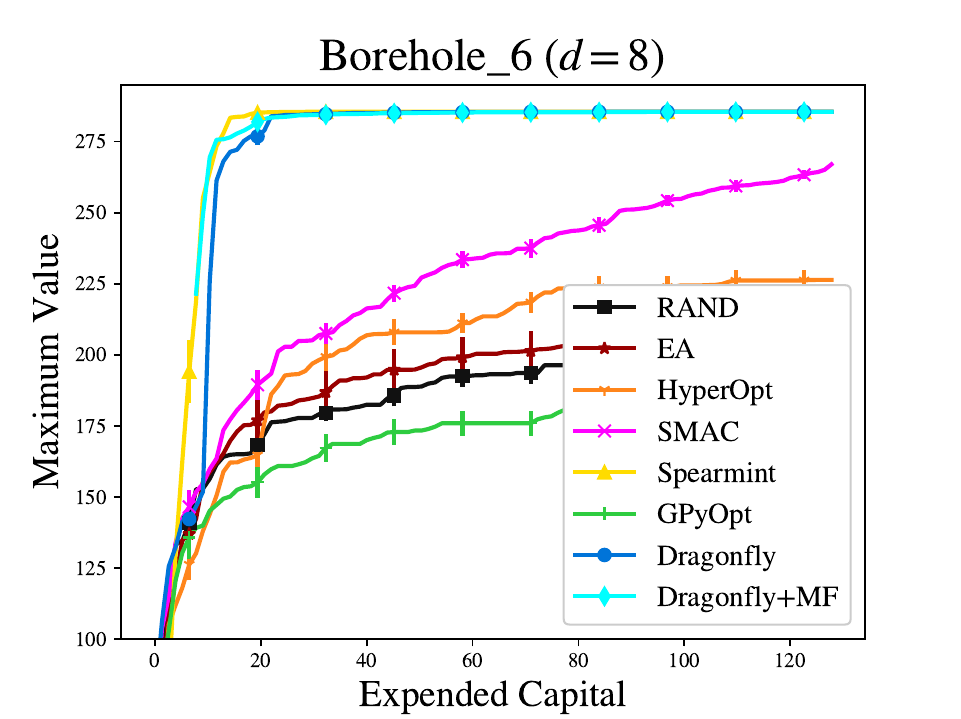}
\hspace{\imhspdflcp}
  \includegraphics[width=\imarrwdflcp]{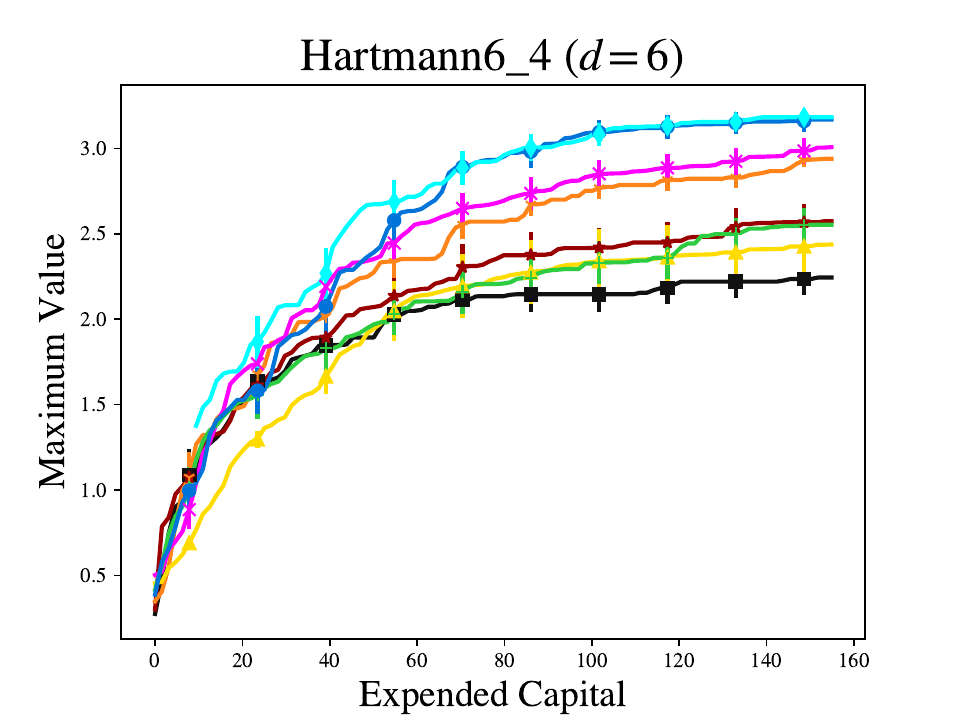}
\hspace{\imhspdflcp}
  \includegraphics[width=\imarrwdflcp]{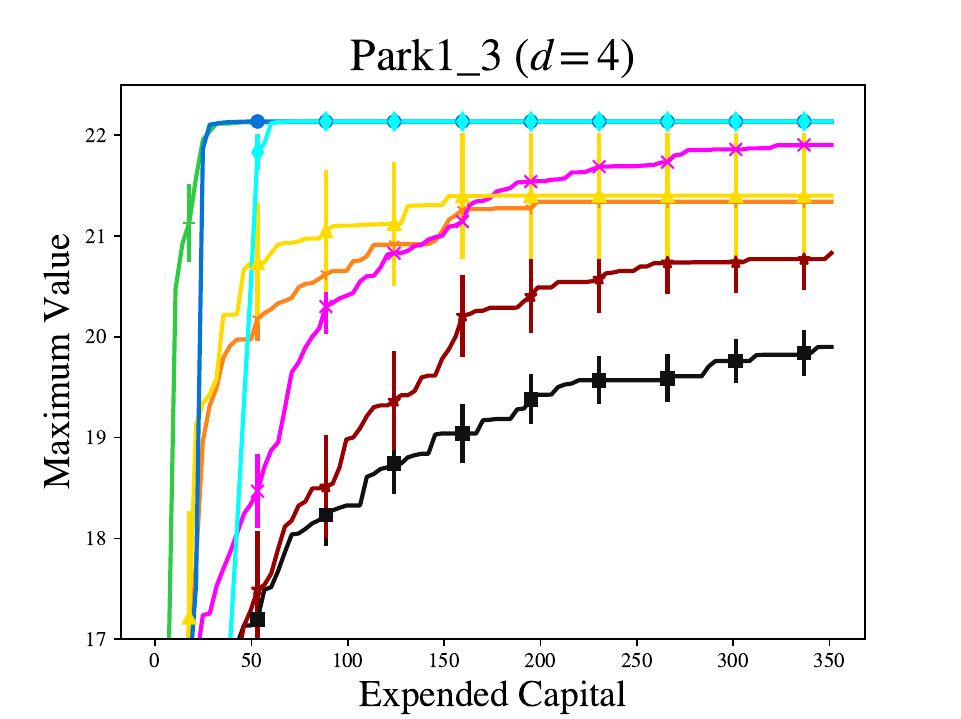}
\hspace{\imhspdflcp}
\\
\hspace{\imhspdflcp}
  \includegraphics[width=\imarrwdflcp]{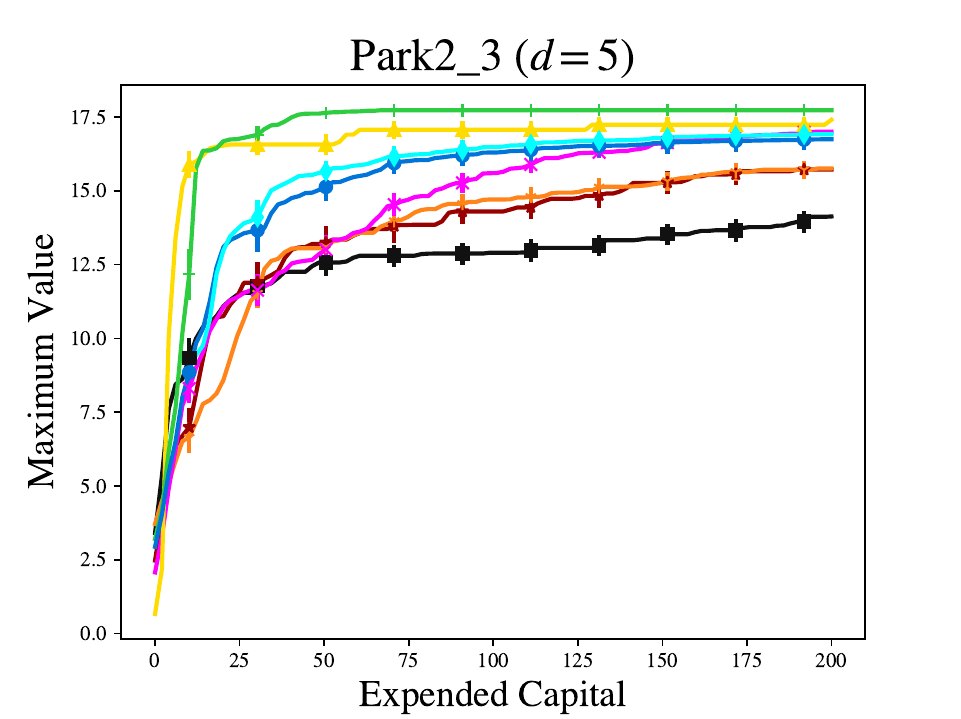}
\hspace{\imhspdflcp}
  \includegraphics[width=\imarrwdflcp]{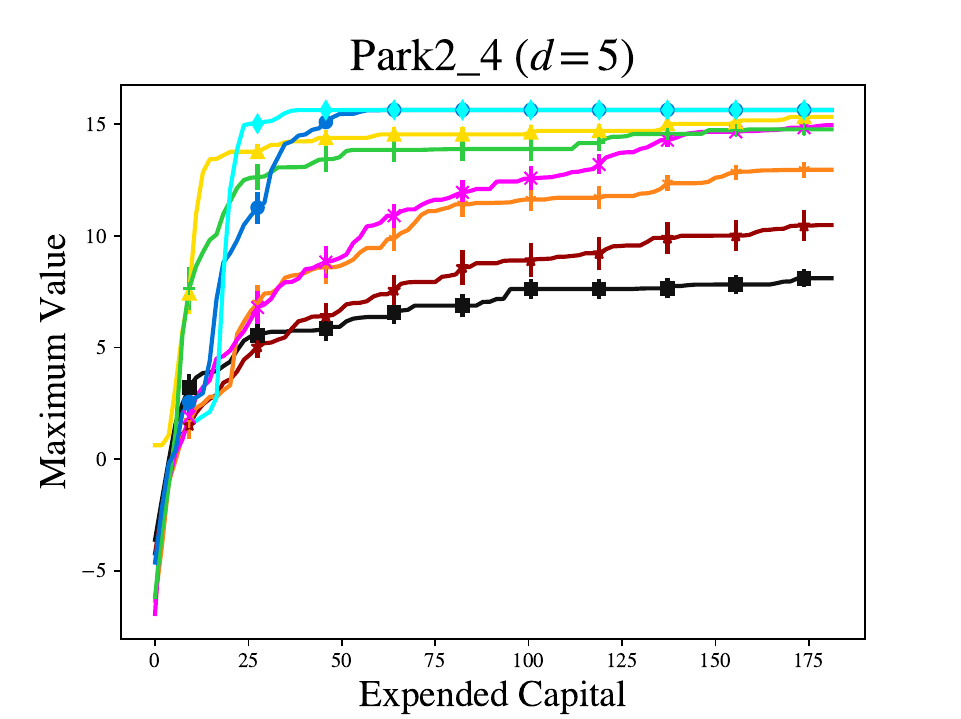}
\hspace{\imhspdflcp}
  \includegraphics[width=\imarrwdflcp]{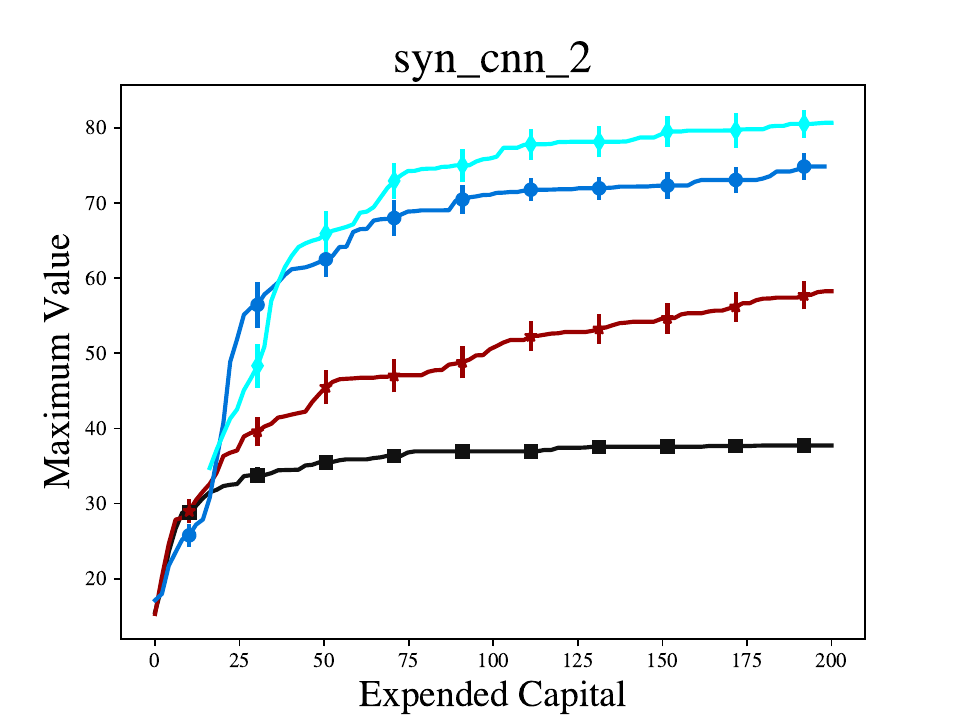}
\hspace{\imhspdflcp}
\\
\vspace{\imcaptionspace}
\caption{\small
\label{fig:dflcpresults}
Comparison of \dragonflys with other algorithms and BO packages on synthetic
functions defined on non-Euclidean domains.
We plot the maximum value, so higher is better.
The $x$-axis shows the expended capital and the $y$ axis is the maximum value
(higher is better).
The title states the name of the function, and its dimensionality (number of variables).
We do not state the dimensionality for the synthetic CNN function since the
dimensionality of a space of CNN architectures is not defined.
All curves were produced by averaging over $20$ independent runs.
Error bars indicate one standard error.
The legend for all curves is available in the first figure.
We do not compare \spearmint, \hyperopt, \smac, and
\gpyopt{} on the synthetic CNN function since
they do not support optimising over neural architectures.
  \vspace{\imtextspace}
}
  \vspace{\imtextspace}
\end{figure}
}

\newcommand{\insertFigdflconstrainedresults}{
\begin{figure}
\centering
\hspace{\imhspdflcp}
  \includegraphics[width=\imarrwdflcp]{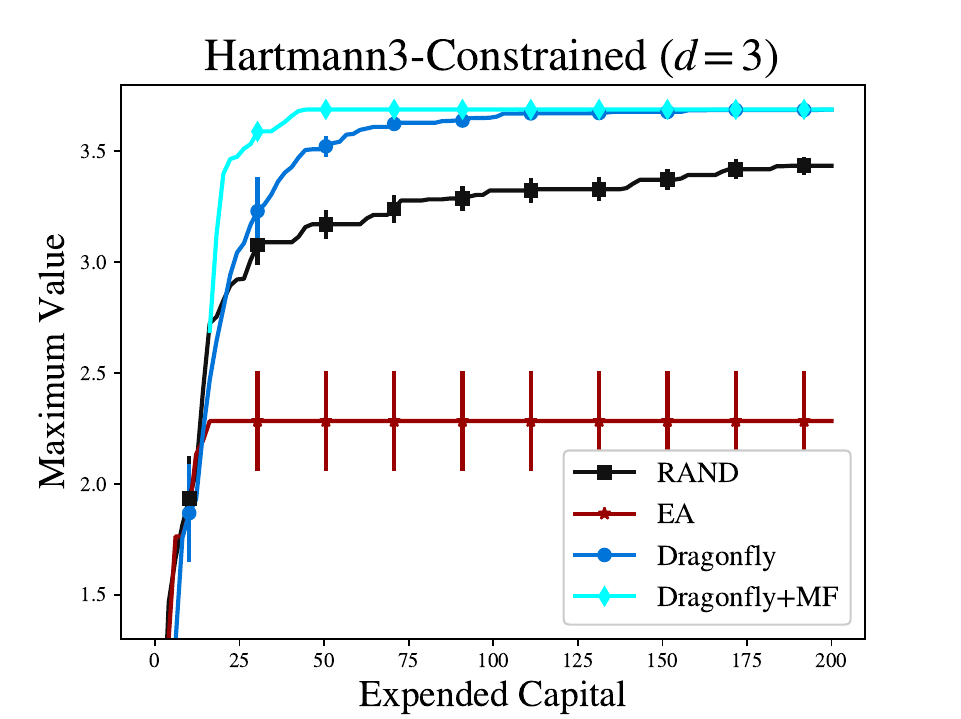}
\hspace{\imhspdflcp}
  \includegraphics[width=\imarrwdflcp]{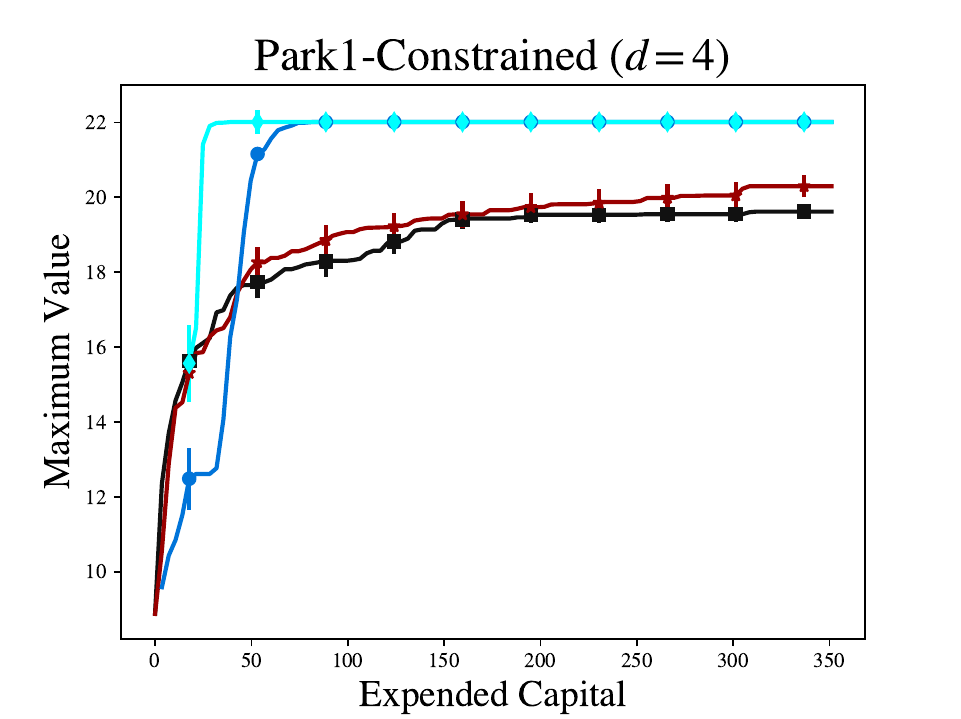}
\hspace{\imhspdflcp}
  \includegraphics[width=\imarrwdflcp]{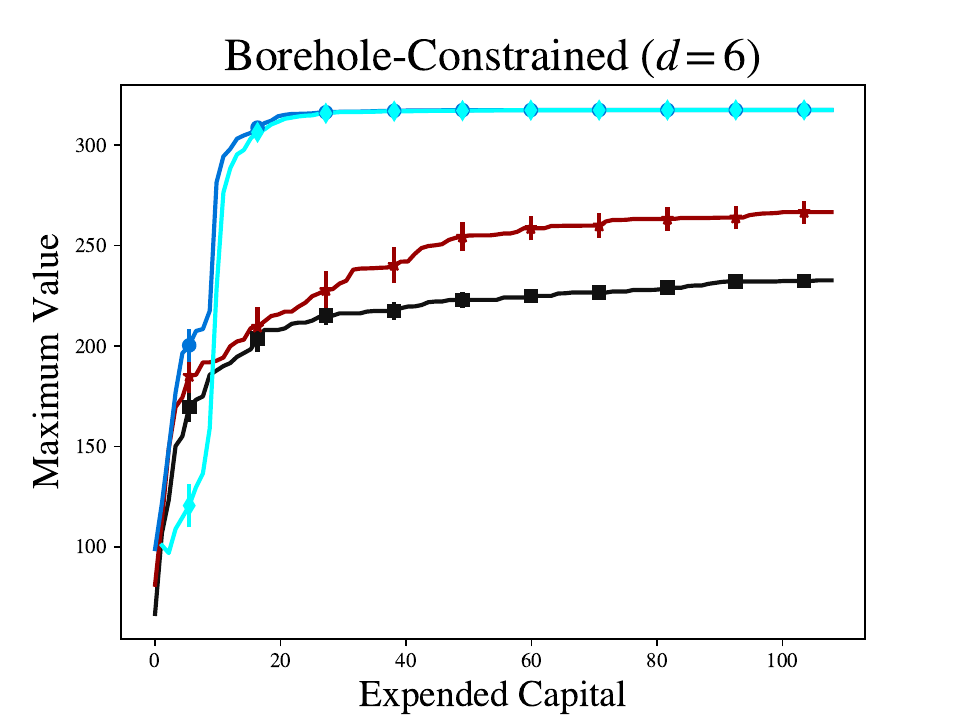}
\hspace{\imhspdflcp}
\\
\vspace{\imcaptionspace}
\caption{\small
\label{fig:dflconstrained}
Comparison of \dragonflys with \rands and \evoalgs on synthetic
functions with constraints on the domain.
See caption under Figure~\ref{fig:dflcpresults} for more details.
  \vspace{\imtextspace}
%   \vspace{\imtextspace}
}
  \vspace{\imtextspace}
\end{figure}
}

\newcommand{\insertFigdflnoisyresults}{
\newcommand{\imarrwdflnoisy}{2.05in}
\newcommand{\imhspdflnoisy}{-0.15in}
\begin{figure}
\centering
% \hspace{\imhspdflnoisy}
  \includegraphics[width=\imarrwdflnoisy]{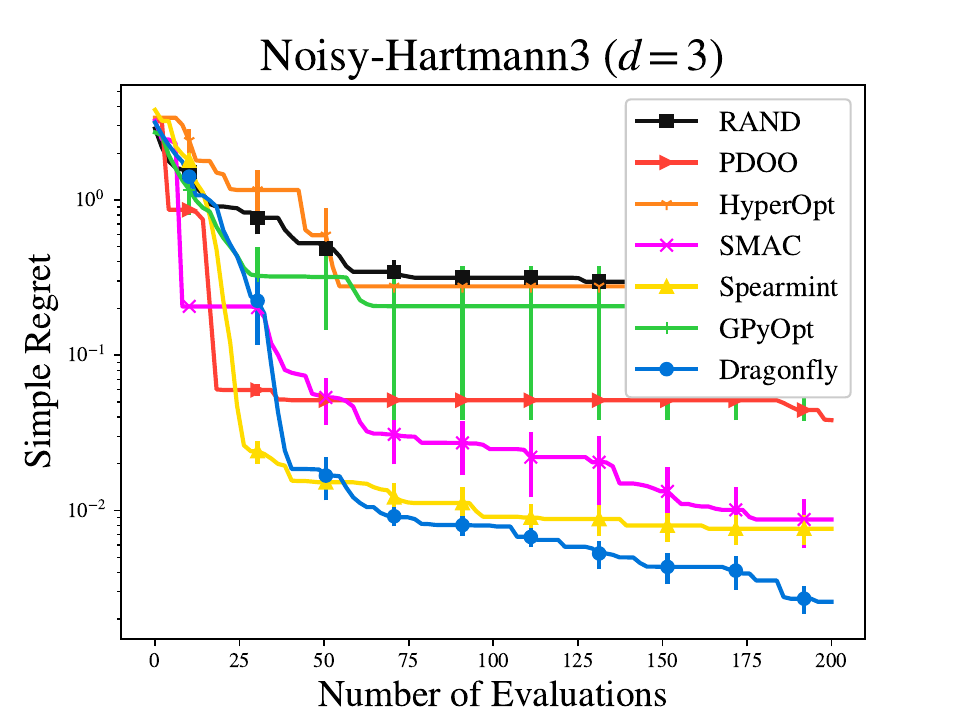}
\hspace{\imhspdflnoisy}
  \includegraphics[width=\imarrwdflnoisy]{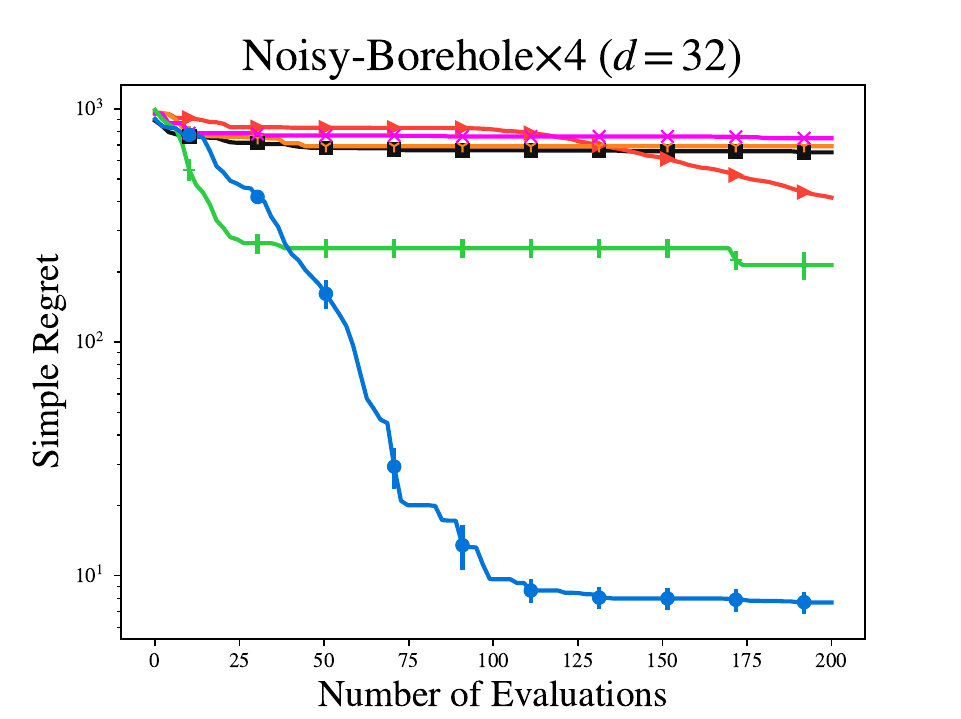}
\hspace{\imhspdflnoisy}
  \includegraphics[width=\imarrwdflnoisy]{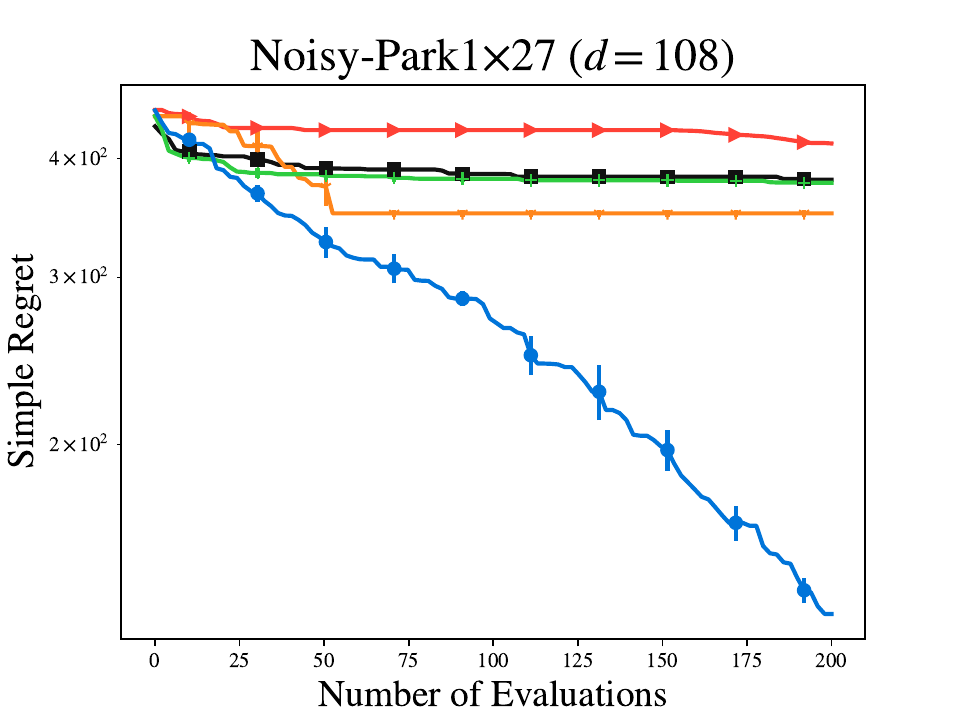}
% \hspace{\imhspdflnoisy}
\\
% \hspace{\imhspdflnoisy}
  \includegraphics[width=\imarrwdflnoisy]{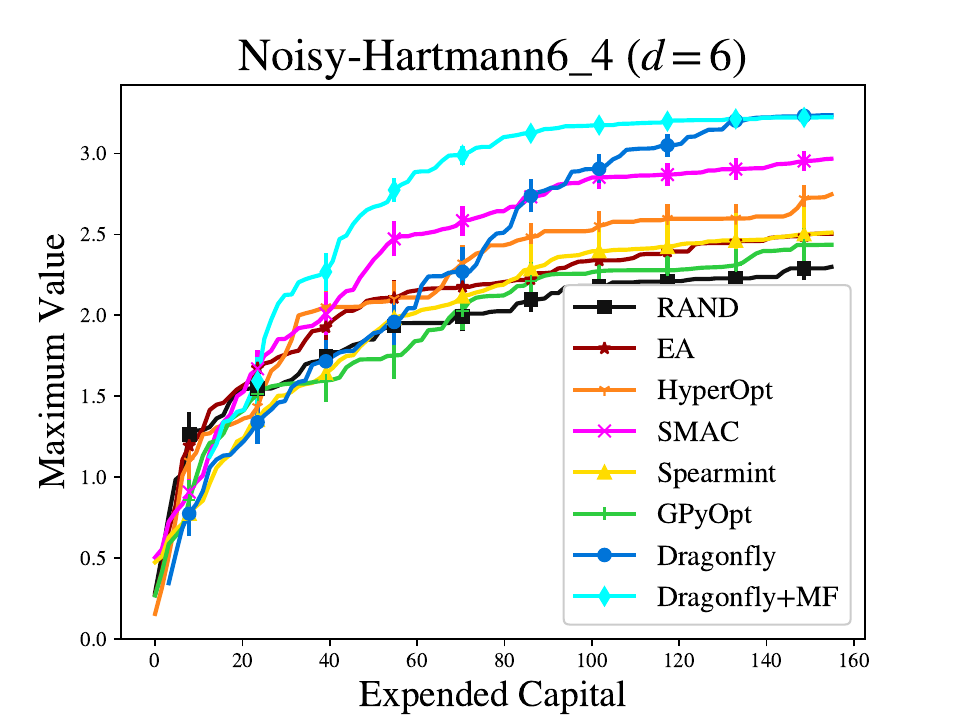}
\hspace{\imhspdflnoisy}
  \includegraphics[width=\imarrwdflnoisy]{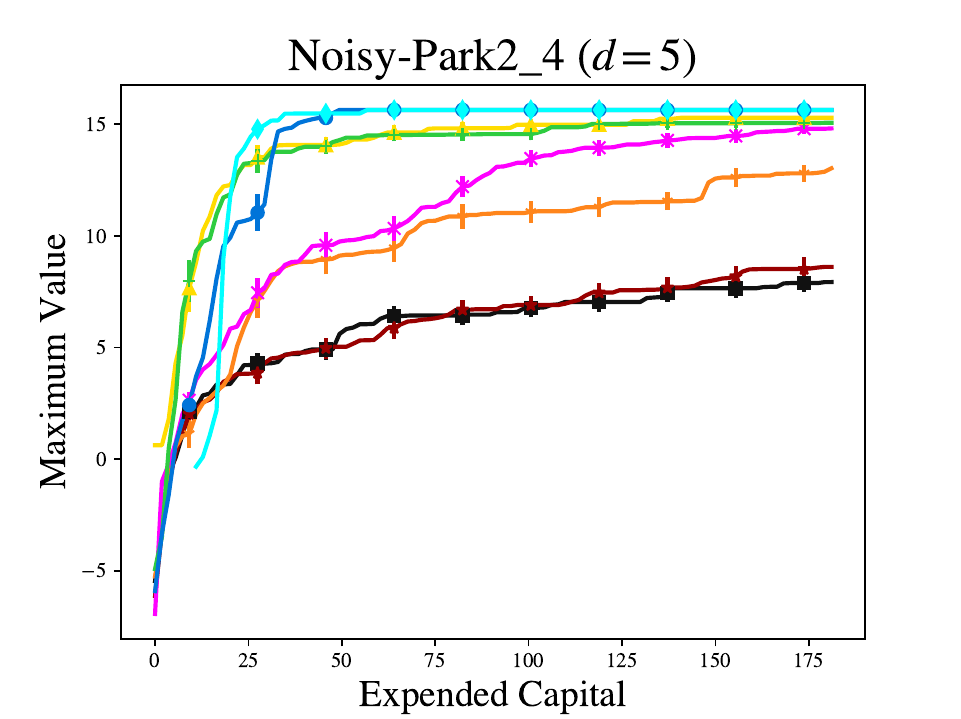}
\hspace{\imhspdflnoisy}
  \includegraphics[width=\imarrwdflnoisy]{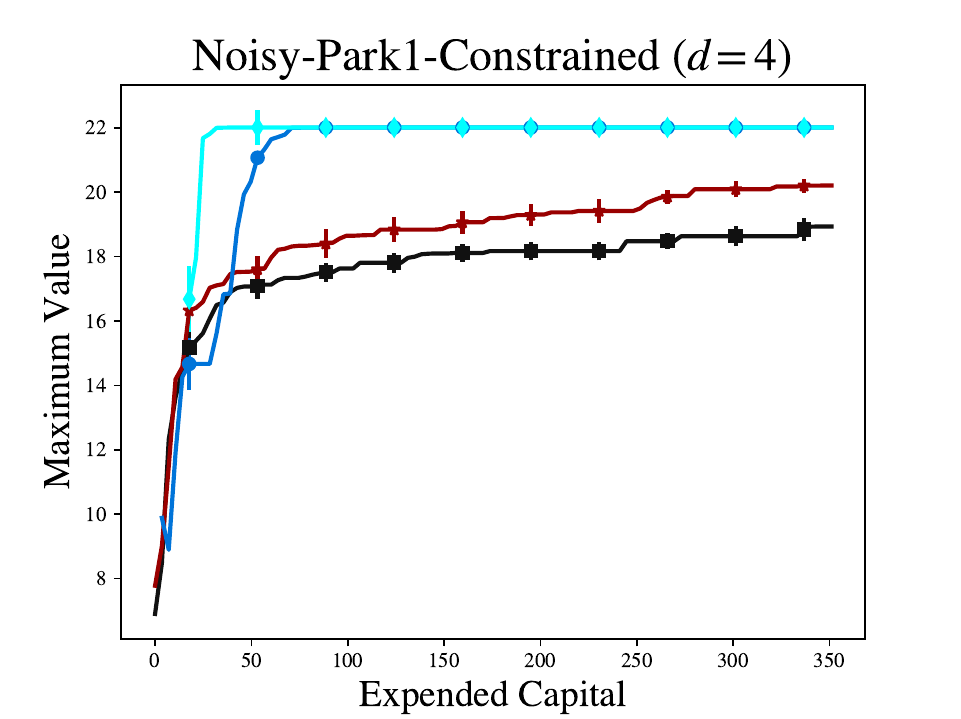}
% \hspace{\imhspdflnoisy}
\\
\vspace{\imcaptionspace}
\caption{\small
\label{fig:dflnoisy}
Experiments on synthetic
functions when evaluations are noisy.
In the top row, the domain is Euclidean and we plot the
number of evaluations versus simple regret (lower is better).
In the bottom row, the domain is non-Euclidean, and we plot the
expended capital versus the maximum true value found (higher is better).
The legend for each row is given in the leftmost figure.
See captions under Figures~\ref{fig:dfleuc},~\ref{fig:dflcpresults},
and~\ref{fig:dflconstrained} for more details. \hspace{-0.1in}
  \vspace{\imtextspace}
%   \vspace{\imtextspace}
}
  \vspace{\imtextspace}
  \vspace{\imtextspace}
\end{figure}
}

\newcommand{\insertDflFigLRG}{
\begin{figure}
\centering
  \begin{minipage}[c]{2.3in}
    \includegraphics[width=2.2in]{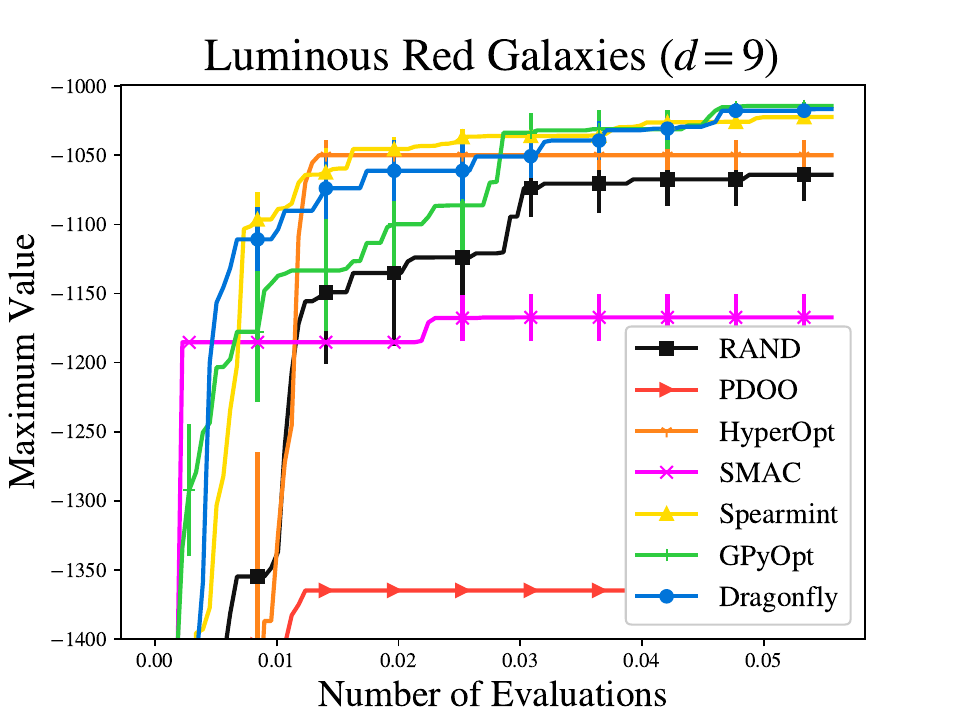}
  \end{minipage} \hspace{-0.1in}
  \begin{minipage}[l]{3.2in}
  \vspace{-0.1in}
    \caption{\small
    Results on the maximum likelihood estimation problem on the luminous red galaxies
    dataset~\citep{tegmark06lrgs}.
    The $x$-axis is the number of evaluations and the $y$-axis is the highest
    likelihood found so far (higher is better).
  All curves were produced by averaging over $10$ independent runs.
  Error bars indicate one standard error.
\label{fig:dfllrg}
    }
  \end{minipage}
%   \vspace{0.10in}
\end{figure}
}

\newcommand{\insertDflFigSNLS}{
\begin{figure}
\centering
  \begin{minipage}[c]{2.3in}
    \includegraphics[width=2.2in]{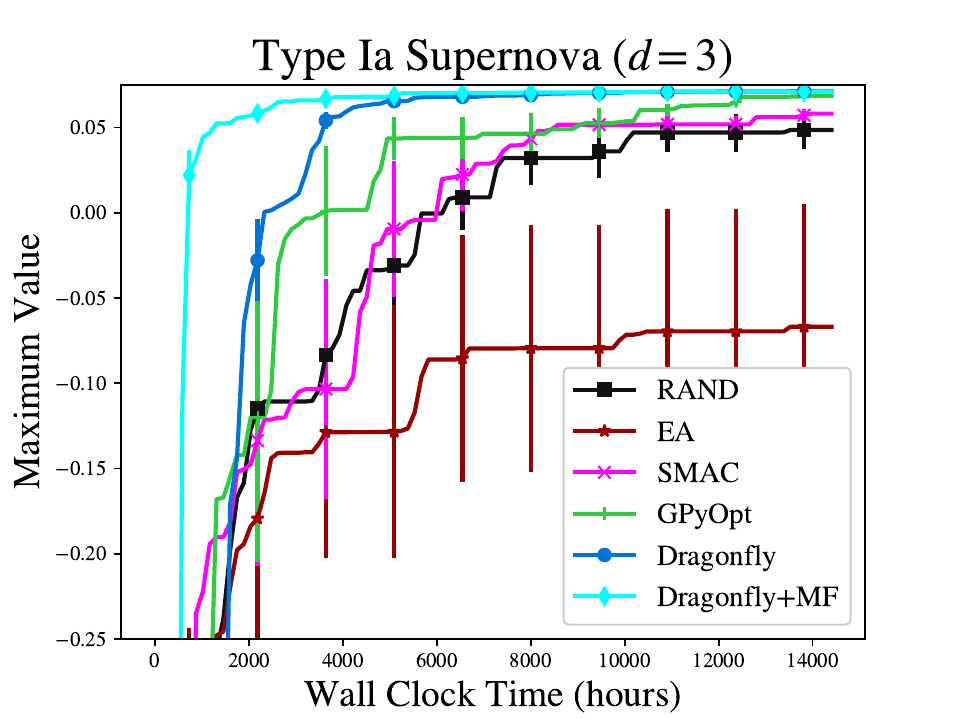}
  \end{minipage} \hspace{-0.1in}
  \begin{minipage}[l]{3.2in}
  \vspace{-0.1in}
    \caption{\small
    Results on the maximum likelihood estimation problem on the Type Ia supernova
    dataset~\citep{davis07supernovae}.
    The $x$-axis is time and the $y$-axis is the highest likelihood found so far (higher
is better).
All curves were produced by averaging over $10$ independent runs.
Error bars indicate one standard error.
\label{fig:dflsnls}
    }
  \end{minipage}
%   \vspace{0.10in}
\end{figure}
}

\newcommand{\insertFigDflAstrophysics}{
\begin{figure}
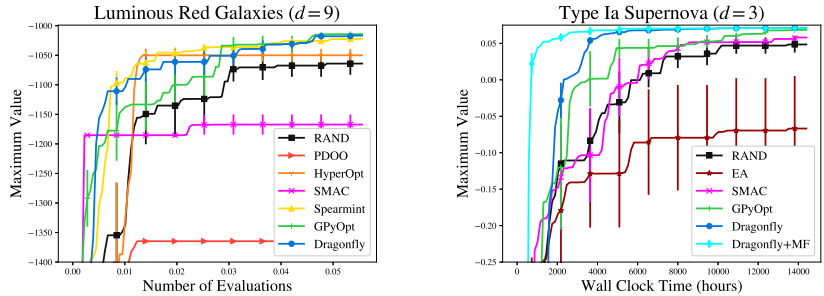

\centering
% \hspace{\imhspdflcp}
  \includegraphics[width=\imarrwdflcp]{figsdfl/real_lrg}
% \hspace{\imhspdflcp}
  \hspace{0.1in}
  \includegraphics[width=\imarrwdflcp]{figsdfl/real_supernova}
% \hspace{\imhspdflcp}
% \hspace{\imhspdflcp}
\\
\vspace{\imcaptionspace}
\caption{\small
\label{fig:dfllrg}
\label{fig:dflsnls}
    Results on the maximum likelihood estimation problem on the luminous red galaxies
    dataset (left) and the supernova dataset (right).
In both figures, the $y$-axis is the highest log likelihood found (higher is better).
On the left figure, the $x$-axis is the number of evaluations and on the right figure
it is wall clock time.
%     likelihood found so far (higher is better).
  All curves were produced by averaging over $10$ independent runs.
  Error bars indicate one standard error.
  \vspace{\imtextspace}
}
  \vspace{\imtextspace}
\end{figure}
}

\newcommand{\insertFigDflModSel}{
\newcommand{\imarrwmodsel}{2.0in}
\newcommand{\imhspmodsel}{-0.15in}
\begin{figure}
\centering
\hspace{\imhspmodsel}
  \includegraphics[width=\imarrwmodsel]{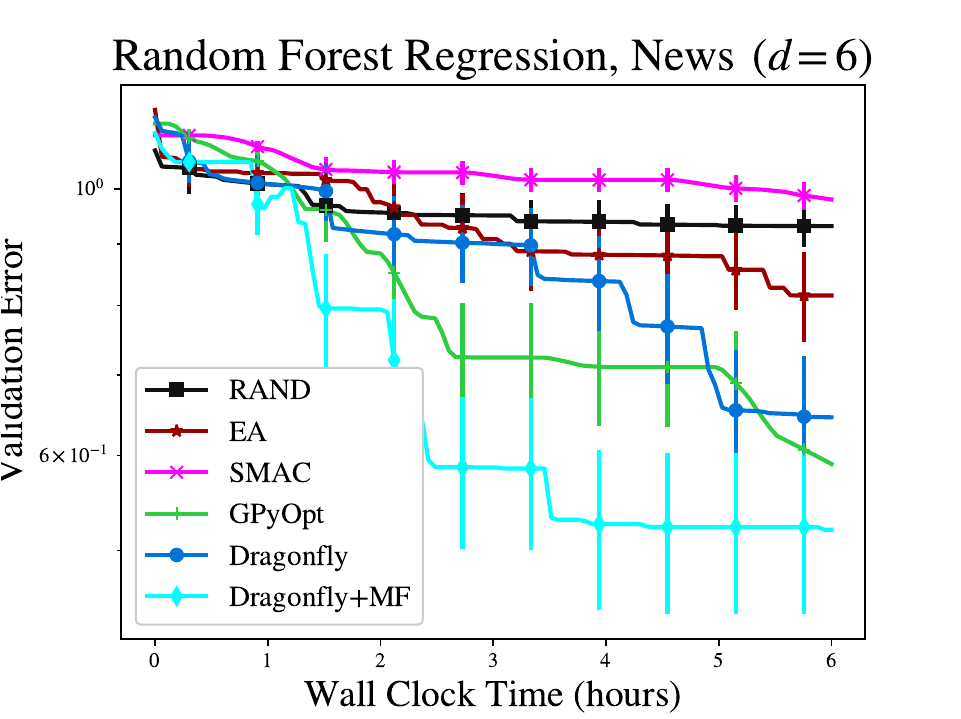}
\hspace{\imhspmodsel}
  \includegraphics[width=\imarrwmodsel]{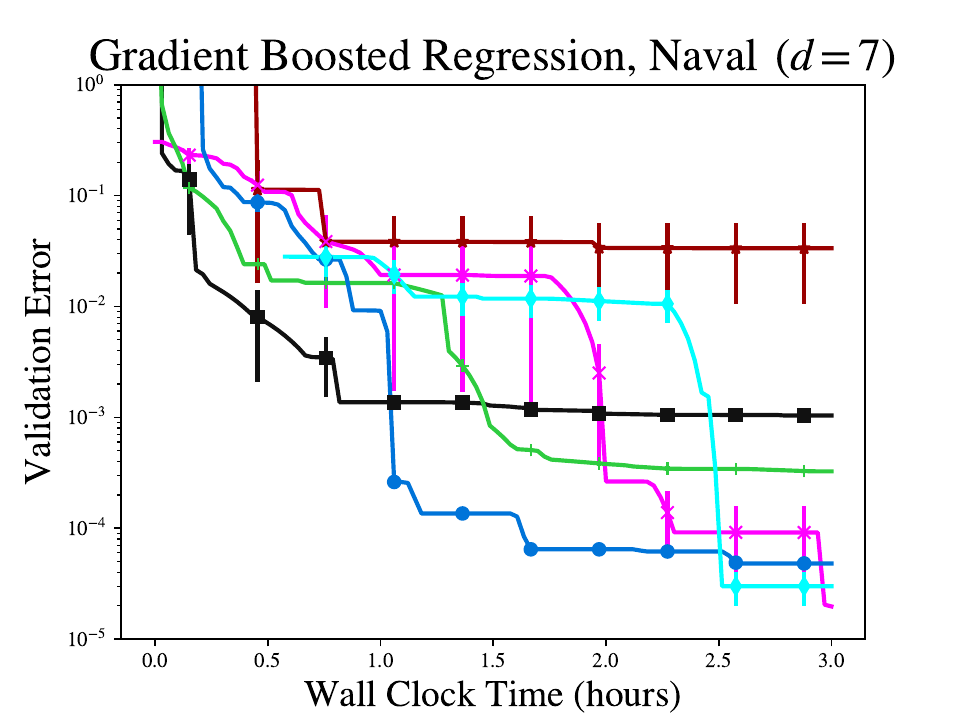}
\hspace{\imhspmodsel}
  \includegraphics[width=\imarrwmodsel]{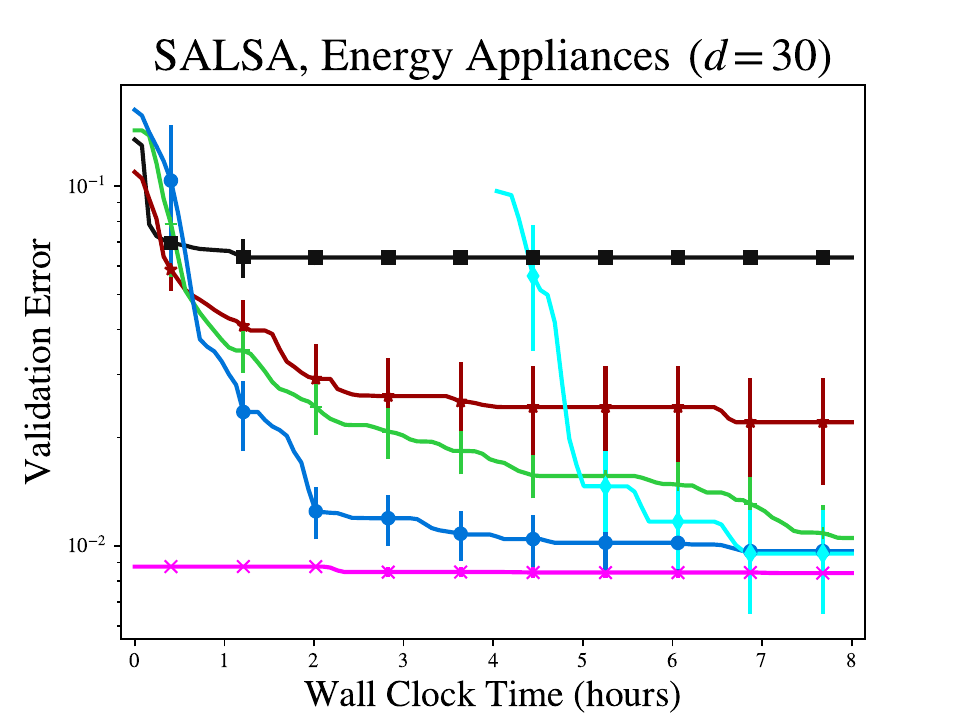}
% \hspace{\imhspmodsel}
\\
\vspace{\imcaptionspace}
\caption{\small
\label{fig:dflmodsel}
    Results on the model selection problems in Section~\ref{sec:dflexpmodsel}.
The title states the method and data set used.
In all figures, the $y$-axis is the validation error (lower is better),
and the $x$-axis is   wall clock time.
%     likelihood found so far (higher is better).
  All curves were produced by averaging over $10$ independent runs.
  Error bars indicate one standard error.
}
  \vspace{\imtextspace}
  \vspace{\imtextspace}
\end{figure}
}

\newcommand{\insertFigdflnnas}{
\newcommand{\imarrwdflnn}{1.95in}
\newcommand{\imhspdflnn}{-0.07in}
\begin{figure}
\centering
%       \hspace{\imhspdflnn}
    \includegraphics[width=\imarrwdflnn]{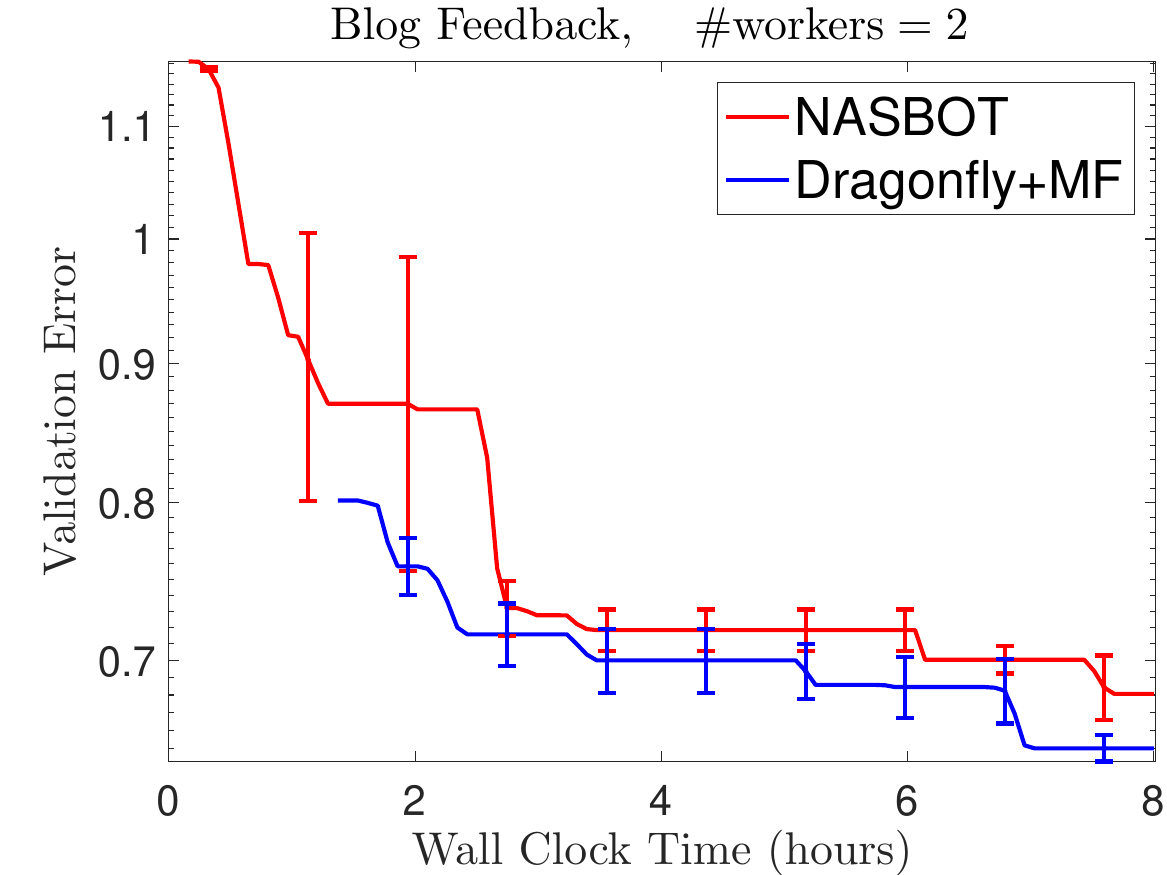}
      \hspace{\imhspdflnn}
    \includegraphics[width=\imarrwdflnn]{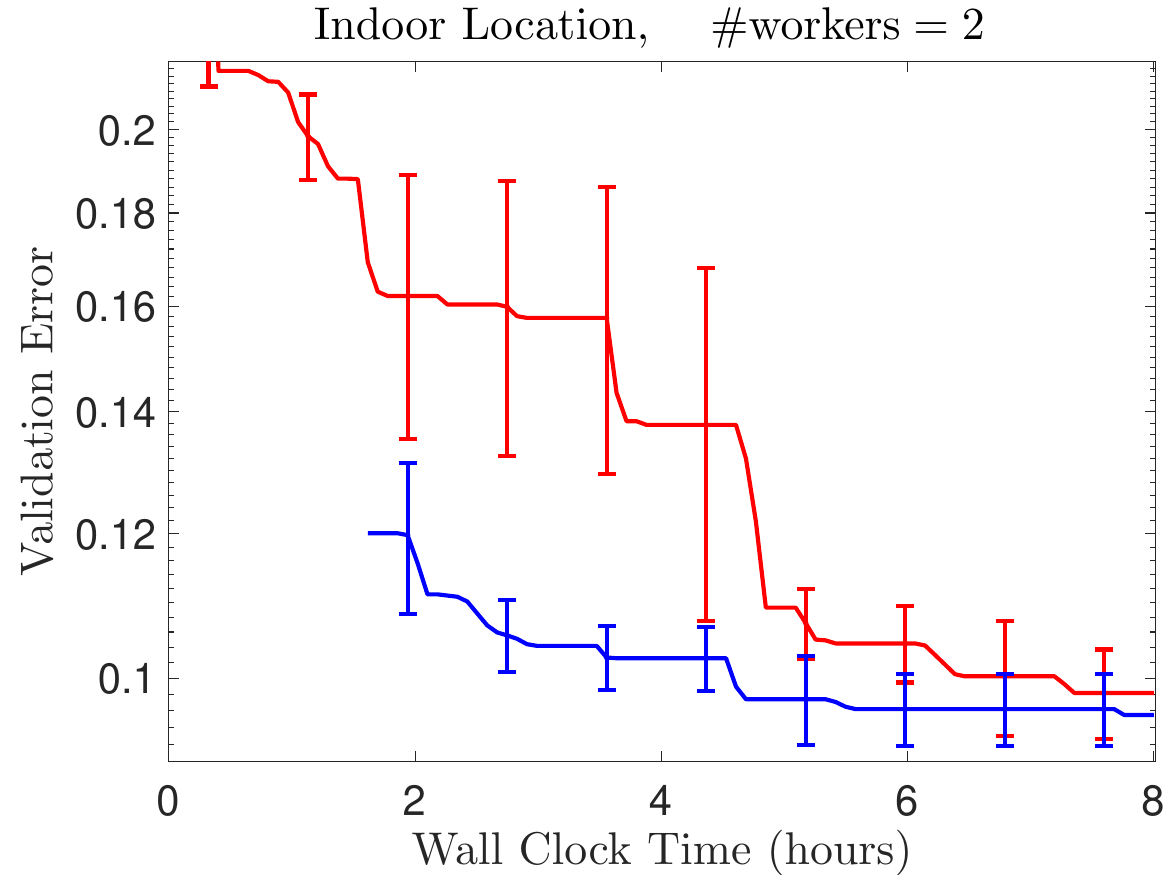}
      \hspace{\imhspdflnn}
    \includegraphics[width=\imarrwdflnn]{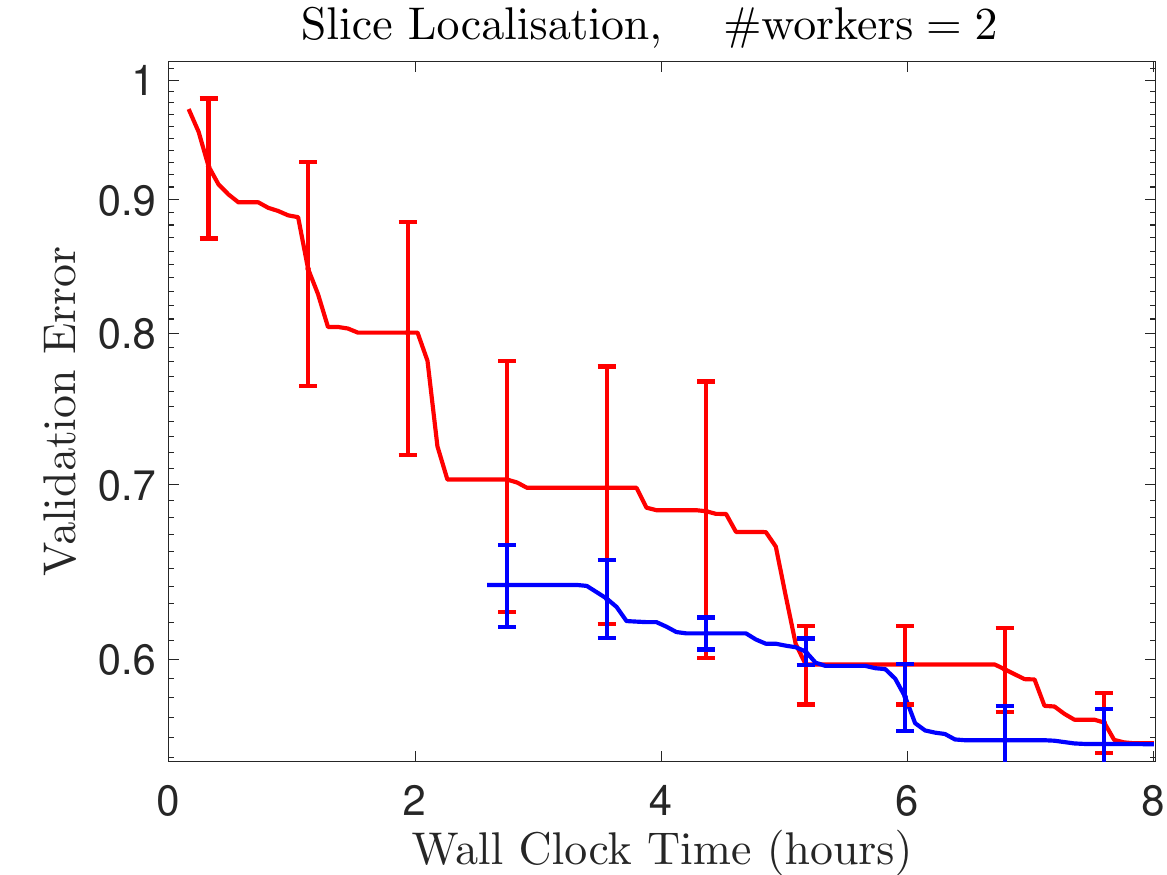}
%       \hspace{\imhspdflnn}
\vspace{\imcaptionspace}
    \caption{\small
Results on the neural architecture search experiments.
In all figures, the $x$-axis is wall clock time.
The $y$ axis is the mean squared validation error (lower is better).
In all cases, we used a parallel set up of two asynchronous workers, where each worker
is a single GPU training a single model.
We used a one dimensional fidelity space where we chose the number of
batch iterations from 4000 to 20,000 ($\zhf=20,000$).
All figures were averaged over 5 independent runs.
Error bars indicate one standard error.
  \label{fig:dflnnas}
    }
  \vspace{\imtextspace}
  \vspace{\imtextspace}
\end{figure}
}

\begin{abstract}
Bayesian Optimisation (BO) refers to a suite of techniques for global optimisation
of expensive black box functions, which
use introspective Bayesian models of the function to
efficiently search for the optimum.
While BO has been applied successfully in many applications,
modern optimisation tasks usher in new challenges where conventional methods
fail spectacularly.
In this work, we 
present \dragonfly, an open source Python library for scalable and robust BO.
\dragonflys
 incorporates multiple recently developed methods that allow BO to be
applied in challenging real world settings;
these include better methods for handling higher dimensional domains,
methods for handling multi-fidelity evaluations when cheap approximations of an
expensive function are available,
methods for optimising over structured combinatorial spaces, such as the space of
neural network architectures,
and methods for handling parallel evaluations.
Additionally, we develop new methodological improvements in BO for selecting the
Bayesian model, selecting the acquisition function, and optimising over
complex domains with different variable types and additional constraints.
% when performing model selection due to larger datasets and increasingly complex models.
We compare \dragonflys to a suite of other packages and algorithms for global
optimisation and demonstrate that when the above methods are integrated, 
they enable significant improvements in the performance of BO.
The 
\dragonflys library is available at \dflrepolink.
% Our python implementation of \dragonflys will be made available open-source.
\end{abstract}

\section{Introduction}
\label{sec:intro}

Many scientific and engineering tasks can be cast as black box optimisation problems,
where
we need to sequentially evaluate a noisy black box function with the goal of finding its
optimum.
A common use case for black box optimisation, pervasive in many industrial and scientific
applications, is \emph{hyperparameter tuning}, where we need to find the optimal
configuration of a black box system by tuning the several knobs which affect the
performance of the system.
For example, 
in scientific simulation studies, parameters in expensive simulations must be
chosen to yield realistic results~\citep{parkinson06wmap3};
in materials design and drug discovery, parameters of a material or drug should be
chosen to optimise the various desired
criteria~\citep{griffiths2017constrained}.
% in industrial design,
% ~\citep{hornby06antenna};
An application for hyperparameter tuning, most relevant to the machine learning
community is \emph{model selection}, where we cannot model
the generalisation performance of a statistical model analytically.
Hence, we need to carry out
expensive train and validation experiments to find the best model for a given task.
Common methods for hyperparameter tuning, in practice, are often
inefficient or
based on heuristics. For example, parameters may be chosen via an exhaustive
(i.e. ``grid'') or random search over the parameter space, or via manual tuning
by domain experts.
In academic circles, the lore is that this work is often done manually
via trial and error by graduate students.

Formally, given a black box
function $\func:\Xcal\rightarrow\RR$ over some domain
$\Xcal$, we wish to find its optimum (maximum) using repeated evaluations to $\func$.
Typically, $\func$
is accessible only via noisy point evaluations, is
non-convex, and has no gradient information.  In
many applications, each evaluation is expensive, incurring a large
computational or economic cost.  Hence, the goal is to maximise $\func$ using
as few evaluations as possible. Methods for this task aim to determine the next
point $x_t$ for evaluation using knowledge of $\func$ acquired via previous
query-observation pairs $\{(\xtt{i},\ytt{i})\}_{i=1}^{t-1}$.

Bayesian Optimisation (\bayo) refers to a suite of  methods for
optimisation, which use a prior belief distribution for $\func$.
To determine future evaluations, \bayos methods use the posterior given the current
evaluations to reason about where to evaluate next.
Precisely, it uses the posterior to construct an acquisition function
$\acqt:\Xcal\rightarrow\RR$ and chooses its maximum
$\xt \in \argmax_{x\in\Xcal} \acqt(x)$ as the next point for evaluation.
 BO usually consumes more computation
to determine future points than alternative methods for global optimisation, but
this pays dividends when evaluating $\func$ is
expensive, as it is usually able to find the optimum in a fewer number of iterations than
such methods.
Bayesian optimisation has shown success in a variety of hyperparameter tuning tasks
including optimal policy search, industrial design, scientific experimentation,
and model selection.

% \bayos methods have a common modus operandi to determine the
% evaluation point $x_t$ at time $t$: first use the posterior for $\func$
% conditioned on $\{(x_i,y_i)\}_{i=1}^{t-1}$ to construct an acquisition function
% $\utilt:\Xcal\rightarrow\RR$; then maximise the acquisition to determine the
% next point, $x_t = \argmax_{x\in\Xcal} \utilt(x)$.  At time $t$, the posterior
% represents our beliefs about $\func$ after $t-1$ observations and $\utilt(x)$
% captures the utility of performing an evaluation at $x$ according to the
% posterior.  In many cases the ancillary optimisation procedure for the
% acquisition $\utilt$ can be nontrivial.  However, since $\utilt$ is
% analytically available, it is usually assumed that the effort for optimising
% $\utilt$ is negligible when compared to an evaluation of $\func$ which requires
% executing an expensive black box experiment.

% \toworkon{list the areas and a few references where BO has had success}.

That said,
optimisation tasks in modern applications face new challenges which cannot be handled by
conventional approaches.
This paper describes \dragonflys (\dflrepolink),
a new open source Python library for BO.
Dragonfly has a primary focus of making BO scalable for modern settings, and
a secondary focus on making BO robust.
\begin{itemize}
\item \textbf{Scalability:}
Over the last few years, we have published a line of work on scaling up BO to address
modern challenges.
These include better methods for handling higher dimensional domains,
methods for handling multi-fidelity evaluations when cheap approximations of an
expensive function are available,
methods for optimising over 
neural network architectures,
and methods for handling parallel evaluations.
These methods have been incorporated into \dragonfly.
In addition, we use evolutionary algorithms to optimise the acquisition,
which enables BO
over complex domains, including those with different variable types and with fairly
general constraints on these variables.

\item \textbf{Robustness:}
Conventional BO methods tend to be sensitive to the choice of the acquisition and
the parameters of the underlying Bayesian model. A common symptom of a bad choice
for these options is that the procedure is unable to improve on a previously
found optimum for a large number of iterations.
Current approaches to handle these settings tend to be very expensive, limiting their
applicability in settings where evaluations to $\func$ are only moderately expensive.
We describe new randomised approaches implemented in \dragonflys which stochastically
sample among an available set of acquisition choices and model parameters instead of
relying on a single value for the entire optimisation routine.
This approach adds much needed robustness to BO, at significantly lower computational
overhead than existing approaches.
% When compared to other tools and algorithms,
% \dragonflys works consistently well across a wider
% range of problems.
\end{itemize}

 We compare \dragonflys to several other packages and algorithms for black box
optimisation and demonstrate that we perform better or competitively in a variety of
synthetic benchmarks and real world tasks in computational astrophysics and model
selection.
Crucially, \dragonflys is able to consistently perform well across a wide array of
problems.

The remainder of this manuscript is organised as follows.
In Section~\ref{sec:boreview}, we review Gaussian processes and \bayo, in
Section~\ref{sec:scaling} we describe our previous efforts for scaling up BO,
and 
in Section~\ref{sec:dflrobust} we describe techniques for improving robustness of BO.
Section~\ref{sec:dflimplementation} describes our implementation, and
Section~\ref{sec:dflexperiments} compares \dragonflys to other popular methods and
packages for global optimisation.

\section{A Brief Review of Gaussian Processes and Bayesian Optimisation}
\label{sec:boreview}

% \subsection{A Review of Gaussian Processes}
\label{sec:gpreview}

\paragraph{A Review of Gaussian Processes:}
A Gaussian Process (GP) over a space $\Xcal$ is a random process from $\Xcal$
to $\RR$.  GPs are typically used as a prior for functions in Bayesian
nonparametrics.  A GP is characterised by a mean function
$\mu:\Xcal\rightarrow\RR$ and a kernel (covariance function)
$\kernel:\Xcal^2\rightarrow\RR$.  If $\func\sim\GP(\mu,\kernel)$, then $f(x)$
is distributed normally $\Ncal(\mu(x), \kernel(x,x))$ for all $x\in\Xcal$.
Some common options for the prior kernel $\kernel$ are the squared exponential
and  \matern{} kernels.
Suppose that we are given $n$ observations $\Dcal_n=\{(x_i, y_i)\}_{i=1}^n$
from this GP, where $x_i\in\Xcal$, $y_i = \func(x_i) + \epsilon_i \in\RR$ and
$\epsilon_i\sim\Ncal(0,\eta^2)$.  Then the posterior $\func|\Dcal_n$ is
also a GP with mean $\mu_n$ and covariance $\kernel_n$ given by,
\begin{align*}
\hspace{-0.05in}
\mu_n(x) = k^\top(K + \eta^2I_n)^{-1}Y, \hspace{0.45in}
\numberthis \label{eqn:gpPost}
\kernel_n(x,x') = \kernel(x,x') - k^\top(K + \eta^2I_n)^{-1}k'.
\hspace{0.05in}
\end{align*}
Here, $Y\in\RR^n$ is a vector with $Y_i=y_i$, and $k,k'\in\RR^n$ are such that
$k_i = \kernel(x,x_i),k'_i=\kernel(x',x_i)$.
% Also, the matrix $\Delta\in \RR^{n\times n}$ is defined as 
% $\Delta = \Kb + \eta^2I$ where
% $\Kb_{i,j} = \kernel(x_i,x_j)$.
$I_n$ is the $n\times n$ identity matrix.  The Gram matrix $K\in \RR^{n\times
n}$ is given by $K_{i,j} = \kernel(x_i,x_j)$.  We have illustrated the prior
and posterior GPs in Figure~\ref{fig:gpucb}.  We refer the reader to Chapter 2
of~\citet{rasmussen06gps} for more on the basics of GPs and their use in
regression.

% \subsection{A Review of Bayesian Optimisation}

\paragraph{A Review of Bayesian Optimisation:}
BO refers to a suite of methods for black box
optimisation in the Bayesian paradigm which use a prior belief distribution for
$\func$.  \bayos methods have a common modus operandi to determine the
next point $\xt$ for evaluation: first use the posterior for $\func$
conditioned on the past evaluations $\{(x_i,y_i)\}_{i=1}^{t-1}$ to construct an
acquisition function
$\utilt:\Xcal\rightarrow\RR$; then maximise the acquisition to determine the
next point, $\xt \in \argmax_{x\in\Xcal} \utilt(x)$.  At time $t$, the posterior
represents our beliefs about $\func$ after $t-1$ observations and $\utilt(x)$
captures the utility of performing an evaluation at $x$ according to this
posterior.
Typically, optimising $\utilt$ can be nontrivial. 
However, since $\utilt$ is
analytically available, it is assumed that the effort for optimising
$\utilt$ is negligible when compared to an evaluation of $\func$.
After $n$ evaluations of $\func$, usually, the goal of an optimisation algorithm is to achieve
small simple regret $\Sn$, defined below.
\begin{align*}
S_n = \func(\xopt) - \max_{t=1,\dots,n} f(x_t),
\numberthis
\label{eqn:regretDefn}
% \label{eqn:Sn}
\end{align*}
While there are several options for the prior for $\func$, such as neural
networks~\citep{snoek2015scalable,springenberg2016bayesian}
and random forests~\citep{hutter11smac}, the most
popular option is to use a GP.
Similarly, while
there are several choices for the acquisition,
for the purpose of this introduction, we focus on
Gaussian process upper confidence bound
(\gpucb)~\citep{auer03ucb,srinivas10gpbandits} and Thompson sampling
(\tsamp) \citep{thompson33sampling}.
The \gpucbs acquisition forms an upper confidence bound for $\func$,
and is defined as,
\begin{align}
\utilt(x) \;= \mutmo(x) + \betath \sigmatmo(x).
\label{eqn:gpucbacqn}
\end{align}
Here $\mutmo$ is the posterior mean of the GP after $t-1$ observations and is
our current estimate of $\func$.  The posterior standard deviation,
$\sigmatmo$, is the uncertainty associated with this estimate.  The $\mutmo$
term encourages an \emph{exploitative} strategy---in that we want to query
regions where we already believe $\func$ is high---and $\sigmatmo$ encourages
an \emph{exploratory} strategy---in that we want to query where we are
uncertain about $\func$ lest we miss high valued regions which have not been
queried yet.  $\betat$ controls the
trade-off between exploration and exploitation.
We have illustrated \gpucbs in Figure~\ref{fig:gpucb}.
Thompson sampling is another popular BO method, where
at each time step, a random sample drawn from the posterior serves as the acquisition
$\acqt$.
Precisely, at time $t$, the next evaluation point is determined by drawing a sample
$h$ from the posterior
and then choosing $\xt \in \argmax_{x\in\Xcal} h(x)$.
\insertFigGPUCB
% 
% In \gpei, the acquisition function measures the expected amount of improvement
% in $\func$, according to the posterior GP. In particular, the acquisition
% function is given by
% \begin{align*}
%   \acqt(x) = \EE\big[\max\{0, \func(x) - \tau_{t-1}\} \big|\{(\xtt{i},
%   \ytt{i})\}_{i=1}^{t-1} \big],
%   \numberthis
% \label{eqn:eiacq}
% \end{align*}
% where $\tau_{t-1} = \argmax_{i\leq t-1} \func(\xtt{i})$ denotes the current best
% value of \func. Note that this expectation can be computed in closed form for GPs.
% where $\tau_{t-1} = \argmax_{i\leq t-1} \mutmo(\xtt{i})$ denotes the current best
% posterior mean among all points that have been evaluated.
% Note that this expectation can be computed in closed form for GPs.
% 
%\begin{align}
%\utilt(x) \;= \mathbb{E}\left[ \text{max}\{\mu'_t(x)-\mutmo(x_\text{best}),0\}
%\right]
%\end{align}
%where the expectation is taken with respect to the posterior of $\func$ given
%the choice of $x$ at time $t$, with mean denoted $\mu'_t(x)$, and where
%$\mutmo(x_\text{best})$ is the posterior mean of the GP evaluated at the point
%$x_\text{best}$ that yields the greatest value, i.e. $x_\text{best} =
%\text{argmax}_{x\in\Xcal}\mutmo(x)$.
% 
% In addition to \gpucbs and \tsamp, 
Other common acquisitions for BO include 
probability of improvement (\probi)~\citep{kushner1964new},
expected improvement (\gpei)~\citep{jones98expensive},
knowledge gradient~\citep{frazier2009kg},
top-two expected improvement \ttei~\citep{qin2017improving},
and entropy based methods~\citep{hennig12entropy}.

Finally, we mention that the mean and kernel function of the prior GP have parameters
which are typically chosen via empirical Bayes methods, such as maximum likelihood or
posterior marginalisation; we will elaborate more on this in Section~\ref{sec:dflgphps}.

\section{Scaling up Bayesian Optimisation}
\label{sec:scaling}

We now describe our prior work in scaling up BO
to modern large scale problems.
% In each subsection, we first review the material in our previous work.
% Then, we propose our extensions and modifications in their implementations in
% \dragonfly.
We provide only a brief overview of each method and refer
the reader to the original publication for more details.
Where necessary, we also provide some details on our implementation in \dragonfly.

\subsection{Additive Models for High Dimensional Bayesian Optimisation}
\label{sec:highdimbo}

In this subsection we consider settings where $\Xcal$ is a compact subset of $\RR^d$.
While BO has been successful in many low dimensional
applications (typically $d<10$),
expensive high dimensional functions occur in several fields such as computer
vision~\citep{bergstra13modelsearch}, antenna design~\citep{hornby06antenna},
computational astrophysics~\citep{parkinson06wmap3} and biology~\citep{gonzalez14gene}.
Existing theoretical and empirical results suggest that BO is exponentially
difficult in high dimensions without further
assumptions~\citep{srinivas10gpbandits,wang13rembo}.

In~\citet{kandasamy15addBO}, we identify two key challenges in scaling BO
to high dimensions.
The first is the \emph{statistical challenge} in estimating the function --
nonparametric regression is inherently difficult in high
dimensions~\citep{gyorfi02distributionfree}.
The second is the \emph{computational challenge} in maximising $\utilt$.
Commonly used methods to maximise $\utilt$ themselves
require computation exponential in dimension.
We showed that we can overcome both challenges by modeling $\func$ as an additive
function.
Prior to our work,
most literature for \bayos in high dimensions
are in the setting where the function varies only along a very low dimensional
subspace \citep{chen12varselbandits,wang13rembo, djolonga13highdimbandits}.
In these works, the authors do not encounter
either the statistical or computational challenge as they perform BO in either a
random or carefully selected lower dimensional subspace.
However, these assumptions can be restrictive in many practical problems.
While our additive assumption is strong in its own right, it is considerably more
expressive.

\insertparaspace

\noindent
\textbf{Key structural Assumption:}
In order to make progress in high dimensions, in~\citet{kandasamy15addBO},
we assumed that $\func$ decomposes into the
following additive form,
\begin{equation}
\func(x) = \funcii{1}(\xii{1}) + \funcii{2}(\xii{2}) + \dots + 
  \funcii{M}(\xii{M}).
\label{eqn:addmodel}
\end{equation}
Here each $\xii{j} \in \Xcalj$ are
lower dimensional groups of dimensionality $p_j$.
% and $\Xcal \supset \bigcup_j \Xcalj$.
% In high dimensional BO,
In this setting,
we are interested in cases where $d$ is very large and
the group dimensionality is bounded: $p_j \leq p \ll d$.
We will refer to the $\Xcalj$'s as \emph{groups} and the grouping of different
dimensions into these groups $\{\Xcalj\}_{j=1}^M$ as the \emph{decomposition}.
The groups are \emph{disjoint} -- i.e.
if we treat the coordinates as a set, $\xii{i}\cap\xii{j} = \emptyset$.
% $\Xcalj = [0,1]^{d_j}$.
% We have $D \geq dM \geq \sum_j d_j \defeq p$.
% We have $D \asymp pM \geq \sum_j p_j$.
% Paranthesised superscripts index the groups and a union over the
% groups denotes the reconstruction of the whole from the groups (e.g.
% $x = \bigcup_j \xii{j}$ and $\Xcal = \bigcup_j \Xcalj$).
% $x = \bigcup_j \xii{j}$).
% 
In keeping with the BO literature,
we assume that each $\funcj$ is sampled from a GP, $\GP(\zero,\kernelj)$
where the $\funcj$'s are independent.
Here, $\kernelii{j}:\Xcalj \times \Xcalj \rightarrow \RR$ is the kernel for
$\funcj$. %W.l.o.g  let $\gpmeanj = \zero$ for all $j$.
This implies that $\func$ itself is sampled from a GP with
an additive kernel $\kernel(x,x') = \sum_j \kernelj(\xii{j},{\xii{j}}')$.
% We will call a kernel such as $\kernelj$ which acts only on $p$ variables a
% $p^{th}$ order kernel.
While other additive GP models have been studied before
(e.g.~\citep{duvenaud11additivegps}),
the above form will pave the way to nice computational properties, as we will see shortly.
% A kernel which acts on all the variables is a $d^{th}$ order kernel.

A natural first inclination given~\eqref{eqn:addmodel} is to try \gpucbs with an additive
kernel.
Since an additive kernel is simpler than a $d$\ssth order kernel,
we can expect statistical gains---in~\citet{kandasamy15addBO} we showed that the regret
improves from being exponential in $d$ to linear in $d$.
However, the main challenge in directly using \gpucbs is that optimising $\utilt$ in high
dimensions can be computationally prohibitive in practice.
For example, using any grid search or branch and bound method,
maximising $\utilt$ to within $\zeta$ accuracy,
requires $\bigO(\zeta^{-d})$ calls to $\utilt$.
To circumvent this,  we proposed  \addgpucbs which exploits the
additive structure in $\func$ to construct an alternative acquisition function.
For this, we first describe inferring the individual $\funcj$'s using observations
 from $\func$.

\insertparaspace

\noindent
\textbf{Inference in additive GPs:}
Suppose we are given observations $Y=\{y_1, \dots, y_n\}$ at
$X = \{x_1, \dots, x_n\}$, where $y_i = \func(x_i) + \epsilon$ and
$\epsilon\sim\Ncal(0,\eta^2)$.
For \addgpucb, we will need the distribution of
$\funcj(\xii{j}_*)$ conditioned on $X, Y$,
which can be shown to be the following Gaussian.
\begingroup
\allowdisplaybreaks
\begin{align*}
\hspace{-0.19in}
\funcj(\xii{j}_*) | x_*, X, Y \,\sim\, \Ncal \big( %\kernelj(\xii{j}_*, \Xii{j}) 
  {k^{(j)}}^\top(K + \eta^2 I_n)^{-1} Y \,,
\kernelj(\xii{j}_*, \xii{j}_*) - {k^{(j)}}^\top (K + \eta^2 I_n)^{-1} {k^{(j)}} \big)
\hspace{-0.08in}
\numberthis \label{eqn:addPosterior}
\end{align*}
\endgroup
where  ${k^{(j)}}\in\RR^n$ are such that ${k_i^{(j)}} = \kernelj(x,x_i)$.
% Also, the matrix $\Delta\in \RR^{n\times n}$ is defined as 
% $\Delta = \Kb + \eta^2I$ where
% $\Kb_{i,j} = \kernel(x_i,x_j)$.
$I_n$ is the $n\times n$ identity matrix.
The Gram matrix $K\in \RR^{n\times n}$ is given by
$K_{i,j} = \kernel(x_i,x_j)=\sum_j\kernelj(x_i,x_j)$.

\textbf{The \addgpucbs acquisition:}
We now define the \addgpucbs acquisition $\utilt$ as,
\begin{equation}
\utilt(x) = \sum_{j=1}^M \gpmeanjtmo(\xii{j}) + \betath \gpstdjtmo(\xii{j}).
\label{eqn:addgpucb}
\end{equation}
$\utilt$ can be maximised by maximising
$\gpmeanjtmo + \betath\gpstdjtmo$ separately on $\Xcalj$.
As we need to solve
$M$ at most $p$ dimensional optimisation problems, it requires only
$\bigO(M^{p+1}\zeta^{-p})$ calls in total to optimise within $\zeta$ accuracy---%
far more favourable than maximising $\utilt$.

\insertparaspace

% \noindent
% \textbf{Implementation in \dragonfly:}
We conclude this section with a couple of remarks.
First,
while our original work used a fixed group dimensionality, in \dragonflys we treat
this and the decomposition as kernel parameters.
We describe how they are chosen at the end of Section~\ref{sec:dflimplementation}.
Second, we note that
several subsequent works have built on our work and studied various additive models
for high dimensional BO.
For example,%
~\citet{wang2017batched,gardner2017discovering} study methods for learning the
additive structure.
Furthermore, the disjointedness in our model~\eqref{eqn:addmodel} can be restrictive
in applications where the function to be optimised has additive structure, but there is
dependence between the variables. Some works have tried to generalise this.%
~\citet{li2016high} use additive models with non-axis-aligned groups,
and~\citet{rolland2018high} study additive models with overlapping groups.

\subsection{Multi-fidelity Bayesian Optimisation}
\label{sec:mfbo}

% \insertKDEHorFigure

Traditional methods for BO are studied in \emph{single
fidelity} settings; i.e. it is assumed that there is just a single expensive
function $\func$.
However, in practice, cheap approximations to $\func$ may be available.
These lower fidelity approximations can be used to discard regions in $\Xcal$ with
low function value. We can then reserve the expensive
evaluations for a small promising region.
For example, in hyperparameter tuning, the cross validation curve of an expensive machine
learning algorithm can be approximated via cheaper training routines using less data
and/or fewer training iterations.
Similarly, scientific experiments can be approximated to varying degrees using cheaper
data collection and computational techniques.

\bayo{} techniques have been used in developing multi-fidelity optimisation methods
in various applications such as hyperparameter tuning and industrial
design~\citep{huang06mfKriging,swersky2013multi,klein2015towards,%
poloczek2016multi}.
However, these methods do not come with theoretical underpinnings.
There has been a line of work with theoretical guarantees developing
multi-fidelity methods for
specific tasks such as active learning~\citep{zhang15weakAndStrong},
and model selection~\citep{li2016hyperband}.
However, these papers focus on the specific applications themselves
and not on general optimisation problems.
In a recent line of
work~\citep{kandasamy2016mfbandit,kandasamy16mfbo,kandasamy2016mfgpbo,kandasamy2017boca},
we studied multi-fidelity optimisation and bandits under various
assumptions on the approximations.
To the best of our knowledge, this is the first line of work that
theoretically formalises and analyses multi-fidelity optimisation.
Of these, while our work in%
~\citet{kandasamy2016mfbandit,kandasamy16mfbo,kandasamy2016mfgpbo}
requires stringent assumptions on the approximations,
our follow up work, \bocas~\citep{kandasamy2017boca}, uses assumptions of a more Bayesian
flavour and can be applied as long as we define a kernel on the approximations.
We first describe the setting and algorithm for \bocas and then discuss the various
modifications in our implementation in \dragonfly.

% \insertparaspace

% \noindent
% \paragraph{Formalism for Multi-fidelity Optimisation:}
\textbf{Formalism for Multi-fidelity Optimisation:}
We will assume the existence of a fidelity space $\Zcal$ and
a function
$\gunc:\Zcal\times\Xcal \rightarrow \RR$ defined on the product space of the fidelity
space and domain.
The function $\func$ which we wish to maximise is related to $\gunc$ via
$\func(\cdot) = \gunc(\zhf, \cdot)$, where $\zhf\in\Zcal$. $\zhf$ is the fidelity at which
we wish to maximise the multi-fidelity function.
% For instance, in the hyperparameter tuning example from Section~\ref{sec:intro},
% $\Zcal = [1,\Nmax]\times[1,\Tmax]$ and $\zhf = [\Nmax, \Tmax]$.
Our goal is to find a maximiser $\xopt \in \argmax_x \func(x) = \argmax_x \gunc(\zhf, x)$.
% We have illustrated this setup in Figure~\ref{fig:fidelSpace}.
In the rest of the manuscript, the term ``fidelities'' will refer to points $z$
in the fidelity space $\Zcal$.
% 
% \insertHorFigFidelSpace
The multi-fidelity framework is attractive when the following two conditions are true.
% \vspace{-0.10in}
% \begin{itemize}[label=-\hspace{-0.02in},leftmargin=0.13in]
\begin{enumerate}[leftmargin=0.19in]
\item
\emph{The cheap $\gunc(z,\cdot)$ evaluation gives us information about
$g(\zhf,\cdot)$.}
% This is true if $\gunc$ is smooth across the fidelity space
% as illustrated in Fig.~\ref{fig:fidelSpace}.
As we will describe shortly, this can be achieved by modelling $\gunc$ as
a GP
with an appropriate kernel for the fidelity space $\Zcal$.
% \end{itemize}
\item
\emph{There exist fidelities $z\in\Zcal$ where evaluating $\gunc$ is cheaper than
evaluating at $\zhf$.}
To this end, we will associate a \emph{known} cost function $\cost:\Zcal\rightarrow
\RR_+$.
% In the hyperparameter tuning example, $\cost(z) = \cost(N,T) = \bigO(N^2T)$.
It is helpful to think of $\zhf$ as being the most expensive fidelity,
i.e. maximiser of $\cost$, and that $\cost(z)$ decreases as we move away from
$\zhf$.
% However, our algorithm
% and results apply to much more general situations.\hspace{-0.1in}
However, this notion is strictly not necessary for our algorithm or
results.\hspace{-0.1in}
\end{enumerate}

We will assume
$\gunc\sim\GP(\zero,\kernel)$, and upon querying at $(z,x)$ we observe
$y = g(z,x) + \epsilon$ where $\epsilon\sim\Ncal(0,\eta^2)$.
$\kernel:(\Zcal\times\Xcal)^2\rightarrow \RR$ is
the prior covariance defined on the product space.
We will exclusively study product kernels $\kernel$ of the following form,
\begin{align*}
\hspace{-0.1in}
\kernel([z,x],[z',x']) \,=\,
\kernelscale\,\kernelz(z,z')\, \kernelx(x,x').
\label{eqn:mfkernel}
\numberthis
\end{align*}
Here, $\kernelscale\in\RR_+$ is the scale of the kernel and $\kernelz,\kernelx$ are
kernels defined on $\Zcal,\Xcal$ such that
$\|\kernelz\|_\infty = \|\kernelx\|_\infty = 1$.
This assumption implies that for any sample $g$ drawn from this GP,
and
for all $z\in\Zcal$, $g(z,\cdot)$ is a GP with kernel $\kernelx$, and vice versa.
This assumption is fairly expressive -- for instance, if we use an SE kernel on the
joint space, it naturally partitions into a product kernel of the above form.

At time $t$,
a multi-fidelity algorithm would choose a fidelity $\zt\in\Zcal$ and a domain point
$\xt\in\Xcal$ to evaluate based on its previous fidelity, domain point,
observation triples $\{(z_i, x_i, y_i)\}_{i=1}^{t-1}$.
Here $\ytt{i}$ was observed when evaluating $g(\ztt{i},\xtt{i})$.
% Before we present our algorithm for the above set up, we will introduce notation for the
% posterior GPs for $\gunc$ and $\func$.
Let $\Dcal_n = \{(\ztt{i},\xtt{i},\ytt{i})\}_{t=1}^n$ be $n$ such
triples from the GP $\gunc$.
We will denote the posterior mean and standard
deviation of $\gunc$ conditioned on $\Dcal_n$ by
$\nutt{n}$ and $\tautt{n}$ respectively
($\nutt{n},\tautt{n}$ can be computed from~\eqref{eqn:gpPost} by replacing
$x\leftarrow[z,x]$).
% Therefore $\gunc(z,x)|\Dcal_n\sim\Ncal(\nutt{n}(z,x), \tausqtt{n}(z,x))$ for all
% $(z,x)\in\Zcal\times\Xcal$.
Denoting,
$
\mutt{n}(\cdot) = \nutt{n}(\zhf,\cdot)$, and
$
\sigmatt{n}(\cdot) = \tautt{n}(\zhf,\cdot)$,
% \begin{align*}
% \mutt{n}(\cdot) = \nutt{n}(\zhf,\cdot), \hspace{0.2in}
% \sigmatt{n}(\cdot) = \tautt{n}(\zhf,\cdot),
% \label{eqn:munsigman}
% \numberthis
% \end{align*}
to be the posterior mean and standard
deviation of $\gunc(\zhf,\cdot) = \func(\cdot)$,
we have that $\func|\Dcal_n$ is also a GP and satisfies
$\func(x)|\Dcal_n \sim \Ncal(\mutt{n}(x), \sigmasqtt{n}(x))$ for all $x\in\Xcal$.

In~\citet{kandasamy2017boca}, we defined both $\kernelz$ and $\kernelx$ to be
radial kernels, and proposed the following two step procedure to
determine the next evaluation.
At time $t$, we will first
construct an upper confidence bound $\utilt$ for the function $\func$ we wish to optimise.
It takes the form,
$\utilt(x) = \mutmo(x) + \betath \sigmatmo(x)$,
% \begin{align*}
% \utilt(x) = \mutmo(x) + \betath \sigmatmo(x).
% \numberthis
% \label{eqn:mfucb}
% \end{align*} 
% Recall from~\eqref{eqn:munsigman} that $\mutmo$ and $\sigmatmo$ are the posterior mean and
% standard deviation of $\func$
where $\mutmo$ and $\sigmatmo$ are the posterior mean and
standard deviation of $\func$
% i.e. $\gunc$ restricted to $\zhf$, 
conditioned on the observations from the previous $t-1$ time steps
at all fidelities, i.e. the entire $\Zcal\times\Xcal$ space.
% in the entire $\Zcal\times\Xcal$ space.
% We will specify $\betat$ in theorems~\ref{thm:regret},~\ref{thm:main}.
Our next point $\xt$ in the domain $\Xcal$ for evaluating
$\gunc$ is a maximiser of $\utilt$, i.e. 
\begin{align*}
\xt \,\in\, \argmax_{x\in\Xcal} \mutmo(x) + \betath \sigmatmo(x) \,
= \, \argmax_{x\in\Xcal} \nutmo(\zhf, x) + \betath \tautmo(\zhf, x).
\label{eqn:mfpointselection}
\numberthis
\end{align*}
We then choose $\zt = \argmin_{z\in\candfidelst(\xt)} \cost(z)$ where,
\begin{align*}
\candfidelst(\xt) = \{\zhf\}\cup
\left\{z: \cost(z) < \cost(\zhf),\; \tautmo(z,\xt) >
\sqrt{\kernelscale} \xi(z) \sqrt{\cost(z)/\cost(\zhf)}\right\}.
\label{eqn:fidelselection}
\numberthis
\end{align*}
Here $\xi(z)$ is an information gap function.
We refer the reader to Section 2 in~\citet{kandasamy2017boca} for the definition and
more details on $\xi(z)$, but intuitively, it measures
the price we have to pay, in information, for querying away from $\zhf$.
$\xi(z)$ is a well defined quantity for radial kernels; for e.g. for
kernels of the form $\kernel(z,z') = \exp(\|z-z'\|^{2\gamma})$, one can show that
$\xi(z)$ is approximately proportional to $\|z-\zhf\|^{\gamma}$.
The second step in \bocas says that we will only consider fidelities where the posterior
variance is larger than a threshold.
This threshold captures the trade-off between cost and
information in the approximations available to us; cheaper fidelities cost less,
but provide less accurate information about the function $\func$ we wish to optimise.
% Next, we take a look at 

% % \vspace{-0.05in}
% \vspace{-0.10in}
% \subsection*{Implementation in \dragonfly}
% \vspace{-0.10in}
% % \vspace{-0.05in}

% \insertparaspace
% \noindent
% \paragraph{Other acquisitions:}
\textbf{Other acquisitions:}
A key property about the criterion~\eqref{eqn:fidelselection}
shown in~\citet{kandasamy2017boca}, is that is chooses a fidelity $\zt$
with good cost to information trade-off, \emph{given} that we are going to
evaluate $g$ at $\xt$.
In particular, it applies to $\xt$ chosen in an arbitrary fashion, and
not necessarily via an
upper confidence bound criterion~\eqref{eqn:mfpointselection}.
Therefore in \dragonfly, we adopt the two step procedure described above, but
allow $\xt$ to be chosen also via other acquisitions as well.

% \paragraph{Exponential decay kernels for monotonic approximations:}
\textbf{Exponential decay kernels for monotonic approximations:}
In~\citet{kandasamy2017boca}, we choose the fidelity kernel $\kernelz$ to be a radial
kernel.
This typically induces smoothness in $\gunc$ across $\Zcal$, which can
be useful in many applications.
However,
in model selection, the approximations are obtained by using less data
and/or less iterations in an iterative training procedure.
In such cases, as we move to the expensive fidelities, the validation performance
tends
to be monotonic---for example, when the size of the training
set increases, one expects the validation accuracy to keep improving.
\citet{swersky2014freeze} demonstrated that an exponential decay kernel
$
\kerneled(u, u') = 1/(u + u' + 1)^\alpha
$,
can strongly support such sample paths.
% $\kerneled:\RR_+\times\RR_+\rightarrow\RR$
% of the following form can strongly support such sample paths,
% \[
% \kerneled(u, u') = \frac{1}{(u + u' + 1)^\alpha}.
% \]
We have illustrated such sample paths in Figure~\ref{fig:expdecay}.
In  a $p$ dimensional fidelity space, one can use
$\kernelz(z,z') = \prod_{i=1}^p\kerneled(z_i,z_i')$ as the kernel for the fidelity space
if all fidelity dimensions exhibit such behaviour.
Unfortunately,  the information gain $\xi(z)$ is not defined for non-radial kernels.
In \dragonfly, we use $\xi(z) = \|z-\zhf\|$ which is similar to the approximation
of the information gain for SE kernels.
Intuitively, as $z$ moves away from $\zhf$, the information gap increases as
$g(z, \cdot)$ provides less information about $g(\zhf, \cdot)$.
% The parameters of $\kernelz$ are treated as hyperparameters of the GP
% and sampled from as part of $\theta$ in lines~\ref{line:hpsample},~\ref{line:hppop}.
% The parameter 
% The parameter $c$ is tuned adaptively as follows: we start with $c=0.001$ but
% if it is not observed for $5$ iterations, we

\insertHorFigExpDecayKernel

% \insertparaspace
% \noindent
% % \paragraph{Combining exponential decay kernels with \addgpucb:}
% \textbf{Combining multi-fidelity with \addgpucb:}
% % In the hyperparameter tuning experiments we consider, the fidelity space is typically
% % low dimensional whereas the domains we optimise over could be high dimensional.
% When using an additive kernel $\kernelx = \sum_j \kernelj$ for the domain $\Xcal$
% in multi-fidelity settings,
% % When we use an exponential decay kernel for $\Zcal$,
% % and an additive kernel $\kernelx = \sum_j \kernelj$ for $\Xcal$,
% the resulting product kernel also takes an additive form,
% $
% \kernel([z, x], [z', x']) =  \sum_j
% \kernelz(z, z') \kernel(\xii{j}, {\xii{j}}')
% $.
% % \[
% % \kernel([z, x], [z', x']) =  \sum_j
% % \kernelz(z, z') \kernel(\xii{j}, {\xii{j}}').
% % \]
% When using \addgpucb, in the first step we choose
% $\xii{j}_t = \argmax_{\xii{j}\in\Xcalj}$
% $\nutmo^{(j)}(\zhf, \xii{j}) + \betath \tautmo^{(j)}(\zhf, \xii{j})$ for all $j$
% to obtain the next evaluation $\xt$.
% Here $\nutmo^{(j)}, \tautmo^{(j)}$ are the posterior GP mean and standard deviation
% of the $j$\ssth function in the above decomposition.
% Then we choose the fidelity $\zt$ as described in~\eqref{eqn:fidelselection}.

This concludes our description of multi-fidelity optimisation in \dragonfly.
Following our work, there have been a few papers on multi-fidelity optimisation with
theoretical guarantees.%
~\citet{sen2018multi,sen2018noisy} develop an algorithm in
frequentist settings which builds on the key
intuitions here, %namely query at lower fidelities until the uncertainty has shrunk.
i.e. query at low fidelities and proceed higher only when the uncertainty has shrunk.
In addition,~\citet{song2018general} develop a Bayesian algorithm  which chooses
fidelities based on the mutual information.

\subsection{Bayesian Optimisation for Neural Architecture Search}
\label{sec:nasbot}

In this section, we study using \bayos for neural
architecture search (NAS), i.e. for finding the optimal neural network architecture
for a given prediction problem.
The majority of the BO literature has focused on settings where the domain
$\Xcal$ is either Euclidean and/or categorical. However,
with the recent successes of deep learning, neural networks are increasingly
becoming the method of choice for many machine learning applications.
Recent work has demonstrated that architectures which deviate
from traditional feed forward structures perform well%
e.g.~\citet{he2016deep,huang2017densely}.
This motivates studying model selection methods which search the space of
neural architectures and optimise for generalisation performance.
While there has been some work on BO for architecture search~\citep{
swersky2014raiders,mendoza2016towards},
they only optimise among feed forward architectures.
% e.g. Figure~\ref{fig:mainnneg1}, but not~\ref{fig:mainnneg2} or~\ref{fig:mainnneg3}.
\citet{jenatton2017bayesian} study methods for BO in tree structured spaces,
and demonstrate an application in optimising feed forward architectures.
Besides BO, other techniques for NAS include
reinforcement learning~\citep{zoph2016neural},
evolutionary algorithms~\citep{liu2017hierarchical},
gradient based methods~\citep{liu2018darts},
and random search~\citep{li2019random}.

There are two main challenges for realising GP based \bayos for architecture search
where each element $x\in\Xcal$ in our domain is now a neural network architecture.
First, we must
quantify the similarity between two architectures $x, x'$ in the form of
a kernel $\kernel(x, x')$. 
% The kernel allows
% us to reason about an unevaluated value $\func(x')$ when we have already
% evaluated $\func(x)$. 
Secondly, we must maximise $\acqt$, the acquisition
function, in order to determine which point $x_t$ to test at time $t$.
% 
% Hence, our challenges in this work are two-fold.  First, we need to
% \emph{quantify (dis)similarity between two networks}.
% %Intuitively, in Fig.~\ref{fig:mainnnegs}, network~\ref{fig:mainnneg1} is more
% %similar to network~\ref{fig:mainnneg2}, than it is to~\ref{fig:mainnneg3}.
% Secondly, we need to be able to traverse the space of such networks to
% \emph{optimise the acquisition function}. 
To tackle these issues, in~\citet{kandasamy2018nasbot} we develop a
(pseudo-) distance for neural network architectures called \nndists (Optimal
Transport Metrics for Architectures of Neural Networks) that can be computed
efficiently via an optimal transport program.
Using
this distance, we develop a \bayos framework for optimising functions
defined on neural architectures called \nnbos (Neural Architecture
Search with Bayesian Optimisation and Optimal Transport), which we describe next.
% This includes an evolutionary algorithm search to
% optimise the acquisition function over the space of architectures.
We will only provide high level details and refer the reader to~\citet{kandasamy2018nasbot}
for a more detailed exposition.

\textbf{The \nndists Distance and Kernel:}
Our first and primary contribution in~\citet{kandasamy2018nasbot} was to develop a
distance metric $d$ among neural networks;
given a distance $d$ of this form, we may use $e^{-\beta d}$ as the kernel.
This  distance was designed taking into consideration
the fact that the performance of an architecture is
determined by the amount of computation at each layer, the types of these
operations, and how the layers are connected.
A meaningful distance  should account for these factors.
To that end, \nndists is defined as the minimum of
a matching scheme which attempts to match a notion of \emph{mass} at the layers from one
network to the layers of the other.
The mass is proportional to the amount of computation happening at each layer.
We incur penalties for matching layers
with different types of operations or those at structurally different
positions.
The goal is to find a matching that minimises these penalties, and the
penalty at the minimum is a measure of dissimilarity between two networks
$\Gcal_1,\Gcal_2$.
In~\citet{kandasamy2018nasbot}, we show that this matching scheme can be formulated as
an optimal transport program~\citep{villani2003topics},
and moreover the solution induces a pseudo-distance
$d$ in the space of neural architectures.
% The caption of Figure~\ref{fig:mainnnegs} gives the \otmanns distance $d$ for the
% three networks illustrated.

% \insertparaspace
% \noindent

\textbf{The \nnbos Algorithm and its implementation in \dragonfly:}
Equipped with such a distance, we use a sum of exponentiated
distance terms as the kernel, as explained previously.
These distances are obtained via different parameters
of \otmanns and/or via normalised versions of the original distance.
To optimise the acquisition, we use an evolutionary algorithm.
For this, we define a library of modifiers which modify a given network by
changing the number of units in a layer, or make structural changes such as adding
skip connections, removing/adding layers, and adding branching (see Table 6
in~\citet{kandasamy2018nasbot}).
The modifiers allow us to navigate the search space, and the evolutionary strategy allows
us to choose good candidates to modify.
In \dragonfly, we also tune for the learning rate, and moreover,
allow for multi-fidelity optimisation, allowing
an algorithm to train a model partially and observe its performance.
We use the number of training batch iterations as a fidelity parameter and use
an exponential decay kernel across the fidelity space.

\subsection{Parallelisation}
\label{sec:parallel}

BO, as described in Section~\ref{sec:boreview},
is a sequential algorithm which determines
the next query after completing previous queries.
However, in many applications, we may have access to multiple workers and hence carry
out several evaluations simultaneously.
As we demonstrated theoretically in~\citet{kandasamy2018parallelised},
when there is high variability in evaluation times, it is prudent for BO
to operate in an \emph{asynchronous} setting,
where a worker is re-deployed immediately
with a new evaluation once it completes an evaluation.
In contrast, if all evaluations take roughly the same amount of time,
it is meaningful to wait for all workers to finish, and incorporate all their feedback
before issuing the next set of queries in batches.

In our implementations of all acquisitions except \ts,
we handle parallelisation using the hallucination technique
of~\citet{desautels2014parallelizing,ginsbourger2011dealing}.
Here, we pick the next point
exactly as in the sequential setting,
 but the posterior is based on
$\filtrt \cup \{(x, \mutmo(x))\}_{x\in\hallucfiltrt}$,
 where $\hallucfiltrt$ are the points in evaluation by other workers at
step $t$ and $\mutmo$ is the posterior mean conditioned on just
$\filtrt$. 
This preserves the mean of the GP, but shrinks
the variance around the points in $\hallucfiltrt$, thus
discouraging the
algorithm from picking points close to those that are in evaluation.
However, for \ts, as we demonstrated
in~\citet{kandasamy2018parallelised},
a naive application would suffice, as the
inherent randomness of \ts{} ensures that the
points chosen for parallel evaluation are sufficiently diverse.
Therefore, \dragonflys does not use hallucinations for \ts.
While there are other techniques for parallelising BO, they either require choosing
additional parameters and/or are computationally
expensive~\citep{gonzalez2015batch,shah2015parallel,wang2018batched}.

\section{Robust Bayesian Optimisation in \dragonfly}
\label{sec:dflrobust}

We now describe our design principles for robust GP based
BO in \dragonfly.
We favour randomised approaches which uses multiple acquisitions and GP hyperparameter
values at different iterations since we found them to be quite robust in our experiments.

\vspace{-0.05in}
\subsection{Choice of Acquisition}
\label{sec:dflacqs}
\vspace{-0.05in}

\insertFigdflacqs

\dragonflys implements several common acquisitions for BO such as \gpucb, \gpei,
\ttei, \tsamp, \addgpucb, and \probi.
The general practice in the BO literature has been for a practitioner to pick their
favourite acquisition, and use it for the entire optimisation process.
However, the performance of each acquisition can be very problem dependent,
as demonstrated in Figure~\ref{fig:dflacqs}.
Therefore, instead of trying to pick the single best acquisition,
we adopt an adaptive sampling strategy which chooses different acquisitions at
different iterations instead of attempting to pick a single best one.
% Initially, \dragonfly chooses all acquisistions with equal probability, but as the
% optimisation routine progresses, tends to favour acquisitions that do well over the
% others.

Our sampling approach maintains a list of $m$ acquisitions $\acqlist$
along with a weight vector
$\wacqt = $ $ \{\wacqtalpha\}_{\alpha\in\acqlist}$ $\in \RR^m$.
We set $\wacqttjj{0}{\alpha} = \initacqwt$ for all $\alpha\in\acqlist$.
Suppose at time step $t$, we chose acquisition $\theta$ and found a higher
$\func$ value than the current best value.
We then update $\wacqttjj{t+1}{\alpha} \leftarrow \wacqtalpha + \indfone(\theta=\alpha)$;
otherwise, $\wacqttjj{t+1}{\alpha} \leftarrow \wacqtalpha$.
At time $t$, we choose acquisition $\theta\in\acqlist$ with probability
$\wacqttjj{t}{\theta}/\sum_\alpha \wacqtalpha$.
This strategy initially samples all acquisitions with equal probability, but
progressively favours those that perform better on the problem.

By default, we set $\acqlist =
\{\text{\gpucb, \gpei, \ts, \ttei}\}$;
for entirely Euclidean domains, we also include \addgpucb.
We do not incorporate \probis since it consistently underperformed other
acquisitions in our experiments.
As Figure~\ref{fig:dflacqs} indicates, the combined approach is robust across different
problems, and is competitive with the best acquisition on the given
problem.
A similar sampling approach to ours is used in~\citet{hoffman2011portfoli}.
\citet{shahriari2014entropy}
use an entropy based approach to select among multiple acquisitions;
however, this requires optimising all of them which can be 
expensive. We found that our approach, while heuristic in nature, performed well
in our experiments.
Finally, we note that we do not implement entropy based
acquisitions,
since their computation can, in general, be quite expensive.

% \insertparaspace
\vspace{-0.05in}
\subsection{GP Hyperparameters}
\vspace{-0.05in}
\label{sec:dflgphps}

One of the main challenges in GP based BO is that the selection of the GP
hyperparameters\footnote{%
Here ``hyperparameters'' refer to those of the GP, such as kernel parameters,
and should not be conflated with the
title of this paper, where ``hyperparameter tuning''
refers to the general practice of optimising a system's performance.%
}
themselves could be notoriously difficult.
While a
common approach is to choose them by maximising the marginal likelihood,
% which
% typically works well once we have collected a sufficient amount of data if the
% landscape of the function is fairly smooth.
% However,
in some cases, this could also cause overfitting in the GP, especially
in the early iterations~\citep{snoek12practicalBO}.
The most common strategy to overcome this issue is to maintain a prior on the
hyperparameters and integrate over the
posterior~\citep{malkomes2016bayesian,snoek12practicalBO,hoffman2014modular}.
%  or
% sample the hyperparameters from the posterior~\citep{hoffman2014modular}.
However, this can be very computationally burdensome,
and hence prohibitive in applications where function evaluations are only moderately
expensive.
Instead, in this work, we focus on a different approach that uses posterior
sampling.
Precisely, at each iteration, one may sample a set of GP hyperparameters from
the posterior conditioned on the data, and use them for the GP at that iteration.
Intuitively, this is similar to a Thompson sampling procedure where the prior on
the hyperparameters specifies a prior on a meta-model, and once we sample the
hyperparameters, we use an acquisition of our choice.
When this acquisition is \ts, this procedure is exactly Thompson sampling
using the meta-prior.

Our experience suggested that maximising the marginal likelihood (\mml)
generally worked well in settings where the function was smooth;
for less smooth functions, sampling from the posterior (\sfp) tended to work better.
We speculate that this is because, with smooth functions,
a few points are sufficient to
estimate the landscape of the function,
and hence maximum likelihood does not overfit;
since it has already estimated the GP hyperparameters well,
it does better than \sfp.
On the other hand,
while \mmls is prone to overfit for non-smooth functions,
the randomness in \sfps prevents us from getting stuck at
bad GP hyperparameters.
As we demonstrate in Figure~\ref{fig:dflgphps}, either of these approaches
may perform better than the other depending on the problem.

% In \dragonfly, we allow a user to specify either approach.
% However, the dafult strategy is to adopt a randomised ensemble
% approach, where we choose either maximum
% likelihood or sampling from the posterior at every iteration,
% similar to how we handled the acquisitions.

\insertFigdflhps

Therefore, similar to how we handled the acquisitions, we adopt a sampling
approach where we choose either maximum likelihood or sampling from the posterior
at every iteration.
Our GP hyperparameter tuning strategy proceeds as follows.
After every $\ncyc$ evaluations of $\func$, we fit a single GP to it via
maximum likelihood, and also sample $\ncyc$ hyperparameter values from the posterior.
At every iteration, the algorithm chooses either \mmls or
\sfps in a randomised fashion.
If it chooses the former, it uses
the single best GP, and if it chooses the latter, it uses one of the sampled values.
% Initially, we choose them with uniform probability, but progressively
% prefer the technique that performs better on the problem -- for this, we adopt a
% weighting strategy similar to the one used for the acquisitions.
For the sampling strategy,
we let $\whpt = \{\whpth\}_{h\in\hplist} \in\RR^2$ where
$\hplist = \{\text{\mml, \sfp}\}$ and choose strategy $h\in\hplist$ with 
probability $\whpttjj{t}{h}/(\whpttjj{t}{\text{\mml}} + \whpttjj{t}{\text{\sfp}})$.
We update $\whpt$ in a manner similar to $\wacqt$.
Figure~\ref{fig:dflgphps} demonstrates that this strategy performs as well as,
if not better than the best of \mmls and \sfp.
% We have demonstrated how this strategy compares against only using maximum
% likelihood or sampling from the posterior in Figure~\ref{fig:dflgphps}.
% We also note that,
% we fit GP hyperparameters only once every $\ncyc$ evaluations due to computational
% reasons; regardless of the chosen hyperparameters, the GP
% posterior~\eqref{eqn:gpPost}, is computed using all the data collected up until the
% current iteration.

By default, \dragonflys uses $\ncyc=17$.
For maximum likelihood of continuous GP hyperparameters, we use either
\direct~\citep{jones93direct} or
PDOO~\citep{grill2015black}.
If discrete hyperparameters are also present, we optimise the
continuous parameters for all choices of discrete values; this is feasible,
since, in most cases,
there are only a handful of discrete GP hyperparameter values.
For posterior sampling, we impose a uniform prior
and use Gibbs sampling as follows.
% and use the following Gibbs sampling procedure.
% At every iteration, we visit each hyperparameter in a randomised
% order; on each visit, we fix the values of the rest of the
% hyperparameters, and sample a value of the current one conditioned on the
% fixed values.
At every iteration, we visit each hyperparameter in a randomised
order; on each visit, we sample a new value for the current hyperparameter,
conditioned on the values of the rest of the hyperparameters.
For continuous hyperparameters, we do so via slice sampling~\citep{neal2003slice}
and, for discrete hyperparameters, we use Metropolis-Hastings.
We use a burn-in of $1000$ samples and collect a sample every $100$ samples from thereon
to avoid correlation.
% Our implementations of slice sampling and Metropolis-Hastings are taken 
% from the PyMC3 library~\citep{salvatier2016probabilistic}, and adapted to suit our
% setting.
% We also experimented with NUTS~\citep{hoffman2014no} for continuous
% hyperparameters, and found it to be significantly
% more computationally expensive but with no significant gains in performance over
% slice sampling.
Next, we describe our BO implementation in \dragonfly.

\section{BO Implementation in \dragonfly}
\label{sec:dflother}
\label{sec:dflimplementation}
\vspace{-0.05in}

% We now describe our BO implementation in \dragonfly.
% These include some modest methodological contributions in optimising the acquisition
% and defining kernels, in order to be able to handle
% domains with different variable types and arbitrary constraints.

% \subsection{Domains \& Fidelity Spaces}
% \label{sec:dfldomains}

% \paragraph{Variable Types:}
% \paragraph{Domains:}
\textbf{Domains:}
\dragonflys allows optimising over domains with Euclidean, integral, and
discrete variables.
% We also define a discrete numeric variable type, where a variable
% can assume one of a finite discrete set of real numbers, and a discrete
% Euclidean type
We also define discrete numeric and discrete Euclidean variable types, where a variable
can assume one of a finite discrete set of real numbers and Euclidean vectors
respectively.
We also allow neural network variable types which permits us to
optimise functions over neural network architectures.
In addition to specifying variables, one might wish to impose constraints on the
allowable values of these variables.
These constraints can be specified via a Boolean function
which takes a point and returns True if and only if the point exists in the domain.
% the point exists in the domain if and only if it returns True.
For example, if we have a Euclidean variable $x\in[-1,1]^2$,
the function $\indfone\{\|x\|_2\leq 1\}$ constrains the domain to the unit ball.
In \dragonfly, these constraints can be specified via
a Python expression or function.

% \subsection{Kernels}
\label{sec:dflkernels}

% \paragraph{Kernels:}
\textbf{Kernels:}
For Euclidean, integral, and discrete numeric variables,
\dragonflys implements the squared exponential and \matern{} kernels.
We also implement additive variants of above kernels for
the \addgpucbs acquisition.
Following recommendations in~\citet{snoek12practicalBO}, we use the \matern-2.5 
kernel by default for above variable types.
We found that it generally performed better across our experiments as well.
For discrete variables, we use the Hamming kernel~\citep{hutter11smac};
precisely, given $x,x'\in\times_{i=1}^k A_i$, where each $A_i$ is a discrete
domain, we use
$\kernel_{\sigma, \alpha}(x, x') = \sigma \sum_{i=1}^k \alpha_i \indfone(x_i = x'_i)$
as the kernel.
Here $\sigma>0$ and $\alpha\in\RR_+^k$, $\sum_i\alpha_i = 1$ are kernel hyperparameters
determining the scale and the relative importance of each discrete variable respectively.
\dragonflys also implements the \otmanns and exponential decay kernels discussed
in Sections~\ref{sec:nasbot} and~\ref{sec:mfbo} respectively.
% For neural network variables, we use the OTMANN kernel~\citep{kandasamy2018nasbot}.
% Finally, we also implement the exponential decay kernel~\citep{swersky2014freeze}
% for multi-fidelity optimisation.

% \subsection{Optimising the Acquisition}
% \label{sec:dflkernels}

% \paragraph{Optimising the Acquisition:}
\textbf{Optimising the Acquisition:}
To maximise the acquisition $\acqt$ in purely Euclidean spaces with no constraints,
we use \direct~\citep{jones93direct} or
PDOO~\citep{grill2015black}, depending on the dimensionality $d$.
Our Fortran implementation of \direct, can only handle up to $d=64$; for larger dimensions
we use PDOO.
In all other cases, we use an evolutionary algorithm.
For this,
we begin with an initial pool of randomly chosen points in the domain and evaluate
the acquisition at those points.
We then generate a set of $\Nmut$ mutations of this pool as follows;
first, stochastically select $\Nmut$ candidates from this set such that those
with higher $\acqt$ values are more likely to be selected;
then apply a mutation operator to each candidate.
Then, we evaluate the acquisition on this $\Nmut$ mutations, add it to the initial
pool, and repeat for the prescribed number of steps.
We choose an evolutionary algorithm since it is simple to implement
and works well for cheap functions, such as the acquisition $\acqt$.
However, as we demonstrate in Section~\ref{sec:dflexperiments}, it is not ideally
suited for expensive-to-evaluate functions.
% 
% We apply the following mutation operators to each individual variable of a given
% candidate point to obtain the final mutation.
% For Euclidean variables, we sample from a Gaussian centred at the Gaussian.
% For integral and discrete numeric, we assign probabilities to all points in the domain
% based on how far they are from the candidate -- we then sample using these probabilities.
% For discrete variables, we choose the same candidate with probability 0.5 or randomly
% choose another element in the domain with probability 0.5.
Each time we generate a new candidate we test if they satisfy the
constraints specified. If they do not, we reject that sample and keep sampling until all
constraints are satisfied.
One disadvantage to this rejection sampling procedure is that if the constrains only
permit a small subset of the entire domain, it could significantly slow down the
optimisation of the acquisition.

% \subsection{Initialisation}
% \label{sec:dflinitialisation}

% \paragraph{Initialisation:}
\textbf{Initialisation:}
% Following recommendations in a line of BO work~\citep{brochu12bo,martinez2015locally},
We bootstrap our BO routine with $\ninit$ evaluations.
For Euclidean and integral variables, these points are chosen via latin hypercube
sampling, while for discrete and discrete numeric variables they are chosen uniformly at
random.
For neural network variables, we choose $\ninit$ feed forward architectures.
Once sampled,
as we did when optimising the acquisition, we use rejection sampling
to test if the constraints on the domain are satisfied.
By default,
$\ninit$ is set to $5d$ where $d$ is the dimensionality of the domain
but is capped off at $7.5\%$ of the optimisation budget.

% We have summarised the resulting procedure in Algorithm~\ref{alg:dragonfly}.
We have summarised the resulting procedure for BO in
\dragonflys in Algorithm~\ref{alg:dragonfly}.
$q,q'$ denote a query.
In usual BO settings, they simply refer to
the next point $x\in\Xcal$, i.e. $q=(x)$;
in multi-fidelity settings they also include the fidelity $z\in\Zcal$,
i.e. $q=(z, x)$.
$\acqlabel(q), \gphplabel(q)$ refer to the acquisition and the choice of \{\mml, \sfp\}
for GP parameter selection.
$\textrm{multinomial-sample}(\ell, w)$ samples an element from a set $\ell$ from
the multinomial distribution $\{w_i/\sum_j w_j\}_{i=1}^{|\ell|}$.
Before concluding this section, we take a look at two implementation details
for \addgpucb.

\insertAlgoDragonfly

% \paragraph{Choosing the decomposition for \addgpucb:}
\textbf{Choosing the decomposition for \addgpucb:}
% We conclude this section by describing the procedure for choosing the additive
Recall from Section~\ref{sec:highdimbo}, the additive
decomposition and maximum group dimensionality for \addgpucbs
are treated as kernel parameters.
% Note that,
% this can pose some computational concerns since the number of possible decompositions
% grows
% combinatorially with dimension.
% We address these challenges, both for maximising the
We conclude this section by describing the procedure for choosing these parameters,
both for \mmls and \sfp.
For the former, since a complete maximisation can be computationally challenging,
we perform a partial maximisation by first choosing
an upper bound $\pmax$ for the maximum group dimensionality. For each group
dimensionality $p\in\{1,\dots,\pmax\}$, we select $k$ different decompositions chosen
at random. For each such decomposition, we optimise the marginal likelihood over the
remaining hyperparameters, and choose the decomposition with the highest likelihood.
For \sfp, we use the following prior for Gibbs sampling.
First, we pick the maximum group dimensionality $p$ uniformly from
 $\{1,\dots,\pmax\}$.
We then randomly shuffle the coordinates $\{1,\dots,d\}$ to produce an ordering.
Given a maximum group dimensionality $p$ and an ordering,
one can uniquely identify a decomposition by iterating through the
ordering and grouping them in groups of size at most $p$.
For example, in a $d=7$ dimensional problem, $p=3$ and the ordering
$4, 7, 3, 6, 1, 5, 2$ yields the decomposition
$\{(3, 4, 7), (1, 5, 6), (2)\}$ of three groups
having group dimensionalities  3, 3, and 1 respectively.
% It is easy to see that sampling from this prior is simple.

% Since the set of possible decompositions is a large combinatorial space,
% a partial maximisation for \mml{} may not recover the true maximiser of the marginal
% likelihood.
% Similarly, for \sfp, we might not be able to search this large
% combinatorial space well with a few samples.
Since the set of possible decompositions is a large combinatorial space,
we might not be able to find the true maximiser of the marginal likelihood in \mml{}
or cover the entire space via sampling in \sfp.
However, we adopt a pragmatic view of the additive
model~\eqref{eqn:addmodel} which views it as a sensible
approximation to $\func$ in the small sample regime,
as opposed to truly believing that $\func$ is additive.
Under this view, we can hope to recover any existing marginal structure in $\func$ via
a partial maximisation or a few posterior samples.
In contrast, an exhaustive search may not do much better when there is no additive
structure.
% Under this view, a partial maximisation may
% capture some existing marginal structure in $\func$, whereas an exhaustive
% maximisation will not do much better than a partial maximisation when there is no
% additive structure.
% Likewise, a few samples would serve the purpose of finding a suitable
% additive model in \sfp.
By default, \dragonflys uses $\pmax = 6$ and $k=25$ for \addgpucb.

% \paragraph{Combining exponential decay kernels with \addgpucb:}
\textbf{Combining multi-fidelity with \addgpucb:}
% In the hyperparameter tuning experiments we consider, the fidelity space is typically
% low dimensional whereas the domains we optimise over could be high dimensional.
When using an additive kernel $\kernelx = \sum_j \kernelj$ for the domain $\Xcal$
in multi-fidelity settings,
% When we use an exponential decay kernel for $\Zcal$,
% and an additive kernel $\kernelx = \sum_j \kernelj$ for $\Xcal$,
the resulting product kernel also takes an additive form,
$
\kernel([z, x], [z', x']) =  \sum_j
\kernelz(z, z') \kernel(\xii{j}, {\xii{j}}')
$.
% \[
% \kernel([z, x], [z', x']) =  \sum_j
% \kernelz(z, z') \kernel(\xii{j}, {\xii{j}}').
% \]
When using \addgpucb, in the first step we choose
$\xii{j}_t \in \argmax_{\xii{j}\in\Xcalj}$
$\nutmo^{(j)}(\zhf, \xii{j}) + \betath \tautmo^{(j)}(\zhf, \xii{j})$ for all $j$
to obtain the next evaluation $\xt$.
Here $\nutmo^{(j)}, \tautmo^{(j)}$ are the posterior GP mean and standard deviation
of the $j$\ssth function in the above decomposition.
Then we choose the fidelity $\zt$ as described in~\eqref{eqn:fidelselection}.

\section{Experiments}
\label{sec:dflexperiments}
% \vspace{-0.05in}

% We now compare \dragonflys to the following algorithms and packages for zeroth order
% optimisation.
We now compare \dragonflys to the following algorithms and packages.
\rand: uniform random search;
\evoalg: evolutionary algorithm;
\pdoo: parallel deterministic optimistic optimisation~\citep{grill2015black};
\hyperopt{} (v0.1.1)~\citep{bergstra2013hyperopt};
\smac{} (v0.9.0)~\citep{hutter11smac};
\spearmint~\citep{snoek12practicalBO};
\gpyopt{} (v1.2.5)~\citep{authors2016gpyopt}.
Of these \pdoos is a deterministic non-Bayesian algorithm for
Euclidean domains.
\smac, \spearmint, and \gpyopts are model based BO procedures, where
\smacs uses random forests, while \spearmints and \gpyopts use GPs.
For \evoalg, we use the same procedure used to optimise the acquisition in
Section~\ref{sec:dflimplementation}.
We begin with experiments on some standard synthetic benchmarks
for zeroth order optimisation.

\vspace{-0.05in}
\subsection{Experiments on Synthetic Benchmarks}
\vspace{-0.05in}

\insertFigdfleucresults

% \paragraph{Euclidean Domains:}
\textbf{Euclidean Domains:}
Our first set of experiments are on a series of synthetic benchmarks in Euclidean domains.
We use the 
Branin $(d=2)$,
Hartmann3 $(d=3)$,
Park1 $(d=4)$,
Park2 $(d=4)$,
Hartmann6 $(d=6)$,
and
Borehole $(d=8)$
benchmarks,
and additionally, construct high dimensional versions of the above benchmarks.
The high dimensional forms were obtained via an additive model
$f(x) = f'(\xii{1}) + f'(\xii{2}) \dots$ where $f'$ is a lower dimensional function
and the $\xii{i}$'s are coordinates forming a low dimensional subspace.
For example, in the Hartmann3x6 problem, we have an $18$ dimensional function obtained
by considering six Hartmann3 functions along coordinate groups
$\{1, 2, 3\}, \{4, 5, 6\}, \dots, \{16, 17, 18\}$.
We compare all methods on their performance over $200$ evaluations.
The results are given in Figure~\ref{fig:dfleuc}, where we plot the
simple regret~\eqref{eqn:regretDefn} against the number of evaluations
(lower is better).
As its performance was typically worse than all other BO methods,
we do not compare to \evoalgs to avoid clutter in the figures.
\spearmints is not shown on the higher dimensional problems since it was too slow
beyond 25-30 dimensions.
\smac's initialisation procedure failed in dimensions larger than $40$ and is
not shown in the corresponding experiments.

% \insertFigdfleucresultstwo

% \insertFigdfleucresultsNoisy
% \insertFigdfleucresultstwoNoisy

\insertFigdflcpresults

% \paragraph{Non-Euclidean Domains:}
\textbf{Non-Euclidean Domains:}
Next, we compare \dragonflys to the above baselines on non-Euclidean domains.
For this, we modify the above benchmarks, originally defined on Euclidean domains,
so that they can take non-Euclidean arguments.
Specifically, we use modified versions of the Borehole, Hartmann6, Park1 and
Park2 functions.
We also construct a synthetic function defined on CNN architectures, on a
synthetic NAS problem.
The results are given in Figure~\ref{fig:dflcpresults}. 
Since the true maxima of these functions are not known, we simply plot the
maximum value found against the number of evaluations (higher is better).
In this set of experiments, in addition to vanilla BO,
we also construct variations of these function which can take a fidelity
argument.
Hence, a strategy may use these approximations to speed up the optimisation process.
The $x$-axis in all cases refers to the expended capital, which was chosen so that
a single fidelity algorithm would perform exactly 200 evaluations.
We compare a multi-fidelity version of \dragonfly, which uses the \bocas
strategy~\citep{kandasamy2017boca}, to choose the fidelities and points for evaluation.
See \href{https://github.com/dragonfly/dragonfly/tree/master/examples/synthetic}{%
\incmtt{\footnotesize github.com/dragonfly/dragonfly/tree/master/examples/synthetic}}
for a description of these functions and the approximations for the multi-fidelity
cases.

% \insertFigNoisydflcpresults

% \paragraph{Domains with Constraints:}
\textbf{Domains with Constraints:}
Next, we consider three optimisation tasks where we impose
additional constraints on the domain variables.
Specifically, we consider versions of the Hartmann3, Park1 and Borehole functions.
As an example, the Hartmann3 function is usually defined on the domain $[0,1]^3$;
however, we consider a constrained domain $\Xcal = \big\{x\in[0,1]^3 \,;\,
x_1^2 + x_2^2 \leq 1/2\big\}$.
Descriptions of the other functions and domains are available in the
\dragonflys repository.
In Figure~\ref{fig:dflconstrained}, 
we compare \dragonflys to \rands and \evoalg.
We do not compare to other methods and packages, since, to our knowledge, they cannot
handle  arbitrary constraints of the above form.

\insertFigdflconstrainedresults

% \paragraph{Noisy Evaluations:}
\textbf{Noisy Evaluations:}
Finally, we compare all methods when evaluations
are noisy.
We use 6 test functions from above, but add Gaussian noise to each
evaluation; the width of the Gaussian was chosen based on the range of the
function.
We evaluate all methods on the maximum true function value queried
(as opposed to the observed maximum value).
The results are given in Figure~\ref{fig:dflnoisy}.

\insertFigdflnoisyresults

\textbf{Take-aways:}
On Euclidean domains,
\spearmints and \dragonflys perform consistently well across the lower dimensional
tasks, but
\spearmints is prohibitively expensive in high dimensions.
On the higher dimensional tasks, \dragonflys is the most competitive.
It is worth pointing out that on the Park1 and Park2 benchmarks, \dragonflys and \spearmints do
significantly better than other methods.
We believe this is because these packages have discovered a good but hard to find optimum
on these problems,
whereas other packages have only discovered worse local optima.
On non-Euclidean domains, we see that
\dragonflys is able to do consistently well.
\gpyopts and \dragonflys perform very well on some problems, but also perform poorly
on others.
It is interesting to note that the improvements due to multi-fidelity
optimisation are modest in some cases.
We believe this is due to two factors.
First, the multi-fidelity method spends an initial fraction of its capital at the lower
fidelities, and the simple regret is $\infty$ until it queries the highest fidelity.
Second, there is an additional statistical difficulty in estimating, what is now a more
complicated GP model across the domain and fidelity space.
We have summarised the results on the synthetic experiments in Table~\ref{tb:ranks} separately
for the Euclidean and non-Euclidean synthetic benchmarks.
These include the results in Figures~\ref{fig:dfleuc},~\ref{fig:dflcpresults},
and~\ref{fig:dflnoisy}.
We exclude the constrained optimisation results from Figure~\ref{fig:dflconstrained} since there are
too few baselines in this setting.
From these results, we see that
\dragonflys performs consistently well across a wide range of benchmarks.

% \subsection*{Experiments on Astrophysical Maximum Likelihood Problems}
\vspace{-0.05in}
\subsection{Experiments on Astrophysical Maximum Likelihood Problems}
\vspace{-0.05in}

In this section, we consider two maximum likelihood problems in computational
Astrophysics.

\insertTableRankSummary

\textbf{Luminous Red Galaxies:}
Here we used data on Luminous Red Galaxies (LRGs)
for maximum likelihood inference on 9 Euclidean cosmological parameters.
The likelihood is computed via the galaxy power spectrum.
Software and data were taken from~\cite{kandasamy15activePostEst,tegmark06lrgs}.
% See~\ref{sec:lrgapp} for more details.
Each evaluation here is relatively cheap,
and hence we compare all methods on the number of evaluations in Figure~\ref{fig:dfllrg}.
We do not compare a multi-fidelity version since cheap approximations were not
available for this problem.
\spearmint, \gpyopt, and \dragonflys do well on this task.

% \insertDflFigLRG
% \insertDflFigSNLS
\insertFigDflAstrophysics

\textbf{Type Ia Supernova:}
We use data on Type Ia supernova for maximum likelihood
inference on $3$ cosmological parameters, the Hubble constant, the
dark matter fraction, and the dark energy fraction.
We use data from~\citet{davis07supernovae}, and
the likelihood is computed using the method described
in~\citet{shchigolev2017calculating}.
This requires a one dimensional numerical integration for each point in the dataset.
We construct a $p=2$ dimensional multi-fidelity problem where we can choose
data set size $N\in[50,192]$ and perform the integration on grids of size
$G\in[10^2, 10^6]$ via the trapezoidal rule.
As the cost function for fidelity selection, we used $\cost(N,G) = NG$.
% the computation is linear in both parameters.
Our goal is to maximise the average log likelihood at $\zhf=[192, 10^6]$.
Each method was given a budget of 4 hours on a 3.3 GHz Intel Xeon processor with
512GB memory.
% The results are given in Figure~\ref{fig:dflsnls} where we plot the maximum average log
% likelihood (higher is better) against wall clock time.
The results are given in Figure~\ref{fig:dflsnls} where we plot the maximum average log
likelihood (higher is better) against wall clock time.
The plot includes the time taken by each method to determine the next point for
evaluation.
We do not compare \spearmints and \hyperopts as they do not provide an API for
optimisation on a time budget.

% \subsection*{Experiments on Model Selection Problems}
\vspace{-0.05in}
\subsection{Experiments on Model Selection Problems}
\label{sec:dflexpmodsel}
\vspace{-0.05in}

We begin with three experiments on tuning hyperparameters of regression methods,
where we wish to find the hyperparameters with the smallest validation error.
We set up a one dimensional fidelity space where a multi-fidelity algorithm
may choose to use a subset of the dataset to approximate the performance when training
with the entire training set.
For multi-fidelity optimisation with \dragonfly, we use an exponential decay kernel
for $\kernelz$~\eqref{eqn:mfkernel}.
The results are presented in Figure~\ref{fig:dflmodsel}, where we plot wall clock time
against the validation error (lower is better).
We do not compare \spearmints and \hyperopts since they do not provide an API for
optimisation on a time budget.

% \insertDflFigSalsa

% \paragraph{Random forest regression, News popularity:}
\textbf{Random forest regression, News popularity:}
In this experiment, we tune random forest regression (RFR) on the
news popularity dataset~\citep{fernandes2015proactive}.
We tune 6 integral, discrete and Euclidean parameters available in the Scikit-Learn
implementation of RFR.
The training set had 20000 points, but could be approximated via a subset of
size $z\in(5000, 20000)$ by a multi-fidelity method.
As the cost function, we use $\cost(z) = z$, since training time is linear in
the training set size.
Each method was given a budget of 6 hours on a 3.3 GHz Intel Xeon processor with
512GB memory.

\insertFigDflModSel
\insertModSelTable

% \paragraph{Gradient Boosted Regression, Naval Propulsion:}
\textbf{Gradient Boosted Regression, Naval Propulsion:}
In this experiment, we tune gradient boosted regression (GBR) on the
naval propulsion dataset~\citep{coraddu2016machine}.
We tune 7 integral, discrete and Euclidean parameters available in the Scikit-Learn
implementation of GBR.
The training set had 9000 points, but could be approximated via a subset of
size $z\in(2000, 9000)$ by a multi-fidelity method.
As the cost function, we use $\cost(z) = z$, since training time is linear in
the training set size.
Each method was given a budget of 3 hours on a 2.6 GHz Intel Xeon processor with
384GB memory.

% \paragraph{SALSA, Energy Appliances:}
\textbf{SALSA, Energy Appliances:}
We use the SALSA regression method~\citep{kandasamy2016salsa} on the
energy appliances dataset~\citep{candanedo2017data} to tune $30$ integral,
discrete, and Euclidean parameters of the model.
The training set had 8000 points, but could be approximated via a subset of
size $z\in(2000, 8000)$ by a multi-fidelity method.
As the cost function, we use $\cost(z) = z^3$, since training time is cubic in
the training set size.
Each method was given a budget of 8 hours on a 2.6 GHz Intel Xeon processor with
384GB memory.
In this example, \smacs does very well because its (deterministically chosen)
initial value luckily landed at a good configuration.

Table~\ref{tb:modselresults} compares the final error achieved by all methods
on the above three datasets at the end of the respective optimisation budgets.
In addition to the methods in Figure~\ref{fig:dflmodsel}, we also show the results
for BO (with \dragonfly) at the lowest fidelity,
random search at the lowest fidelity and
\hyperband~\citep{li2016hyperband}, which is a multi-fidelity method which uses
random search and successive halving.
For example, for BO and random search at the lowest fidelity on the RFR problem, we
performed the same procedure as \rands and \dragonfly, but only using $5000$ points
at each evaluation.
Interestingly, we see that for the GBR experiment, BO using only a fraction of the
training data, outperforms BO using the entire training set.
This is because, in this problem, one can get good predictions even with $2000$ points,
and the lower fidelity versions are able to do better since they are able to
perform more evaluations within the specified time budget than the versions which
query the higher fidelity.

% \subsection*{Experiments on Neural Architecture Search}

% \vspace{-0.1in}
% \subsection*{Experiments on Neural Architecture Search}
% \vspace{-0.1in}

\textbf{Experiments on Neural Architecture Search:}
Our next set of model selection experiments demonstrate the NAS features
in \dragonfly.
Here, we tune for the architecture of the network using the \otmann{}
kernel~\citep{kandasamy2018nasbot} and the learning rate.
Each function evaluation,
trains an architecture with stochastic gradient descent (SGD)
with a fixed batch size of 256.
We used the number of batch iterations in a one dimensional fidelity space, i.e.
$\Zcal  = [4000, 20000]$ for \dragonflys while \nnbos always queried with
$\zhf = 20,000$ iterations.
% As the cost function, we use $\cost(z) = z$, since training time is linear in
% the number of iterations.
We test both methods in an asynchronously parallel set up of two
GeForce GTX 970 (4GB) GPU workers with a computational budget of $8$ hours.
Additionally, we also impose the following constraints on the space of architectures:
maximum number of layers: $60$,
maximum mass: $10^8$,
maximum in/out degree: $5$,
maximum number of edges: $200$,
maximum number of units per layer: $1024$,
minimum number of units per layer: $8$.
We present the results of our experiments on the
blog feedback~\citep{buza2014feedback},
indoor location~\citep{torres2014ujiindoorloc},
and 
slice localisation~\citep{graf20112d},
datasets in Figure~\ref{fig:dflnnas}.
For reference, we also show the corresponding results for a vanilla
implementation of \nasbot~\citep{kandasamy2018nasbot}.
\dragonflys  outperforms  \nasbots primarily because it
 is able to use cheaper evaluations to approximate fully trained models.
Additionally, it benefits from tuning the learning rate and
using robust techniques for selecting the acquisition and GP hyperparameters.

\insertFigdflnnas

% \input{conclusion}
% \vspace{-0.05in}
\section{Conclusion}
\label{sec:conclusion}
% \vspace{-0.05in}
% \vspace{-0.05in}

Bayesian optimisation is a powerful framework for optimising expensive blackbox
functions.
% It  uses a Bayesian model of the function to reason
% about where to perform the next evaluation. 
% It finds applications in many hyperparameter tuning applic
However, with increasingly
expensive function evaluations and demands to optimise over complex input
spaces, BO methods face new challenges today.  In this work, we describe our
multiple efforts to scale up BO to address the demands of modern large scale
applications and techniques for improving robustness of BO methods.
We implement them in an integrated fashion in an open source platform, \dragonfly,
and demonstrate that
they outperform existing platforms for BO in a variety of applications.
Going forward, we wish to integrate techniques for
multi-objective optimisation~\citep{paria2018flexible} and for using
customised models via probabilistic programming~\citep{neiswanger2019probo}.
% \vspace{-0.05in}

% \vspace{-0.05in}

% \newpage
% \vspace{0.2in}
% \appendix
% \section*{\Large Appendix}
% \input{jappExperiments}
% \input{jappAncillary}

% \vspace{-0.05in}
\paragraph{Acknowledgements:}
This work was funded by DOE grant DESC0011114, NSF grant IIS1563887,
and the DARPA D3M program.
This work was done when KK was a Ph.D. student at Carnegie Mellon University,
where he was supported by a Facebook fellowship and a Siebel scholarship.
We thank Anthony Yu, Shuli Jiang, and Shalom Yiblet for assisting with the
development of \dragonfly.
% \vspace{-0.05in}

% % % Acknowledgements should go at the end, before appendices and references
% \acks{%
% This work was partly funded by DOE grant DESC0011114, NSF grant IIS1563887,
% the DARPA D3M program, and AFRL.
% KK is supported by a Facebook fellowship and a Siebel scholarship.
% We wish to thank Shuli Jiang and Shalom Yiblet for assisting with the initial
% development of the code base.
% }

% {\small
{\footnotesize
\renewcommand{\bibsection}{\section*{References\vspace{-0.1em}} }
\setlength{\bibsep}{1.1pt}
\bibliography{kky,bib_thesis,bib_boss,koopmans}
}

\end{document}